\documentclass[10pt,twocolumn,letterpaper]{article}

\usepackage[pagenumbers]{arxiv}

\usepackage{graphicx}
\usepackage{amsmath}
\usepackage{amssymb}
\usepackage{booktabs}
\usepackage{enumitem}
\usepackage{comment}
\usepackage{adjustbox}
\usepackage{multirow}
\usepackage{array}
\usepackage{float}
\usepackage{placeins}
\usepackage{pifont}
\newcommand{\cmark}{\ding{51}}%

\usepackage{array}
\newcolumntype{H}{>{\setbox0=\hbox\bgroup}c<{\egroup}@{}}

\usepackage{graphicx}
\usepackage{colortbl}
\usepackage{bbold}
\usepackage[dvipsnames]{xcolor}
\usepackage{amsmath}
\usepackage{amssymb}
\usepackage{booktabs}
\usepackage{enumitem}
\usepackage{comment}

\newcommand\OURS{{POCO}}
\newcommand{\titletext}{\OURS: Point Convolution for Surface Reconstruction}

\newcommand{\footnoteref}[1]{$^{\text{\ref{#1}}}$}

\usepackage{amsmath}

\DeclareMathOperator*{\argmin}{arg\,min}
\DeclareMathOperator*{\closest}{closest}

\newcommand{\decoder}{D}
\newcommand{\encoder}{E}
\newcommand{\latentv}{{\mathbf{z}}}
\newcommand{\neighbors}{{\mathcal{N}}}
\newcommand{\occupancy}{\mathbf{o}}
\newcommand{\occupancyp}{{o}}
\newcommand{\occupancyf}{{\omega}}
\newcommand{\pointcloud}{{\mathcal{P}}}
\newcommand{\point}{{\mathbf{p}}}
\newcommand{\qpoint}{{\mathbf{q}}}

\newcommand{\Real}{{\mathbb{R}}}

\newcommand{\relativizer}{R}
\newcommand{\significance}{s}

\newcommand{\weight}{w}
\newcommand{\weightv}{{\mathbf{w}}}

\newcommand{\ConvONet}{ConvONet}

\newcommand{\balex}{\bgroup\color{BrickRed}}
\newcommand{\ealex}{\egroup}
\newcommand{\brenaud}{\bgroup\color{blue}}
\newcommand{\erenaud}{\egroup}

\definecolor{lightgreen}{rgb}{0.95,1,0.95}

\newcommand{\tabfirst}{\cellcolor{blue!25}}
\newcommand{\tabsecond}{\cellcolor{blue!10}}

\newcommand{\Ntrain}{N_{\text{train}}}
\newcommand{\Ntest}{N_{\text{test}}}
\newcommand{\Nview}{N_{\text{view}}}
\usepackage[accsupp]{axessibility}  
\usepackage{tocloft}
\usepackage{scrwfile}
\TOCclone[\contentsname~(\appendixname)]{toc}{atoc}
\newcommand\StartAppendixEntries{}
\AfterTOCHead[toc]{%
  \renewcommand\StartAppendixEntries{\value{tocdepth}=-10000\relax}%
}
\AfterTOCHead[atoc]{%
  \edef\maintocdepth{\the\value{tocdepth}}%
  \value{tocdepth}=-10000\relax%
  \renewcommand\StartAppendixEntries{\value{tocdepth}=\maintocdepth\relax}%
}
\newcommand*\appendixwithtoc{%
  \cleardoublepage
  \appendix
  \addtocontents{toc}{\protect\StartAppendixEntries}
  \listofatoc
}

\usepackage[pagebackref,breaklinks,colorlinks]{hyperref}

\usepackage[capitalize]{cleveref}
\crefname{section}{Sec.}{Secs.}
\Crefname{section}{Section}{Sections}
\Crefname{table}{Table}{Tables}
\crefname{table}{Tab.}{Tabs.}

\begin{document}

\title{\titletext}

\author{
Alexandre Boulch$^1$
\and 
Renaud Marlet$^{1,2}$
\and
\large
\hspace{-3mm}\textsuperscript{1}Valeo.ai, Paris, France  \hspace{1mm} \textsuperscript{2}LIGM, Ecole des Ponts, Univ Gustave Eiffel, CNRS, Marne-la-Vall\'ee, France
}

\maketitle

\begin{abstract}
Implicit neural networks have been successfully used for surface reconstruction from point clouds. However, many of them face scalability issues as they encode the isosurface function of a whole object or scene into a single latent vector. To overcome this limitation, a few approaches infer latent vectors on a coarse regular 3D grid or on 3D patches, and interpolate them to answer occupancy queries. In doing so, they lose the direct connection with the input points sampled on the surface of objects, and they attach information uniformly in space rather than where it matters the most, i.e., near the surface. Besides, relying on fixed patch sizes may require discretization tuning. To address these issues, we propose to use point cloud convolutions and compute latent vectors at each input point. We then perform a learning-based interpolation on nearest neighbors using inferred weights. Experiments on both object and scene datasets show that our approach significantly outperforms other methods on most classical metrics, producing finer details and better reconstructing thinner volumes.
The code is available at \url{https://github.com/valeoai/POCO}.
\end{abstract}

\section{Introduction}

Constructing a surface or volume representation from 3D points sampled at the surface of an object or scene has numerous applications, from digital twins processing to augmented and virtual reality. Cheaper sensors directly producing 3D points (depth cameras, low-cost lidars) and
mature multi-view stereo techniques \cite{Schoenberger2016CVPR,Schoenberger2016ECCV} operating on images offer increasing opportunities for such reconstructions.

\begin{figure}[t]
    \centering\setlength{\tabcolsep}{2pt}
    \begin{tabular}{ccc}
        Input (65536 pts) & SA-ConvONet & \OURS~(ours) \\
        \midrule
        \includegraphics[width=0.31\linewidth, trim=0 0 0 20, clip]{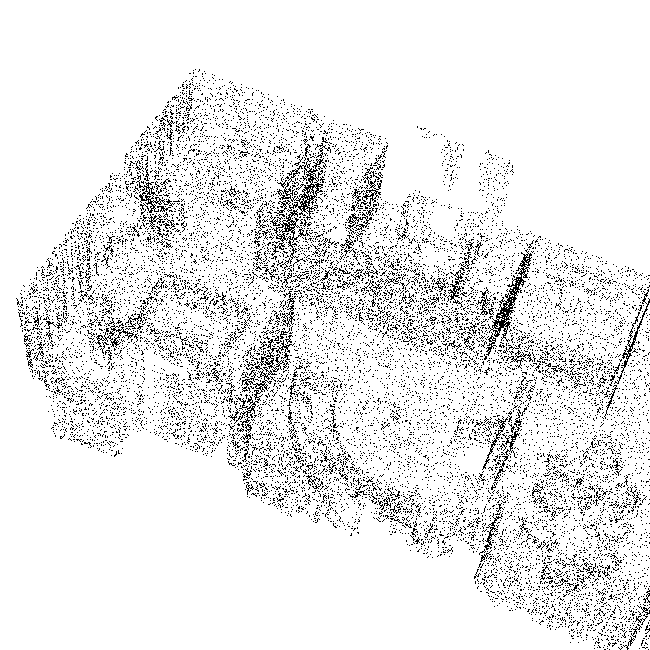}&
        \includegraphics[width=0.31\linewidth, trim=0 0 0 20, clip]{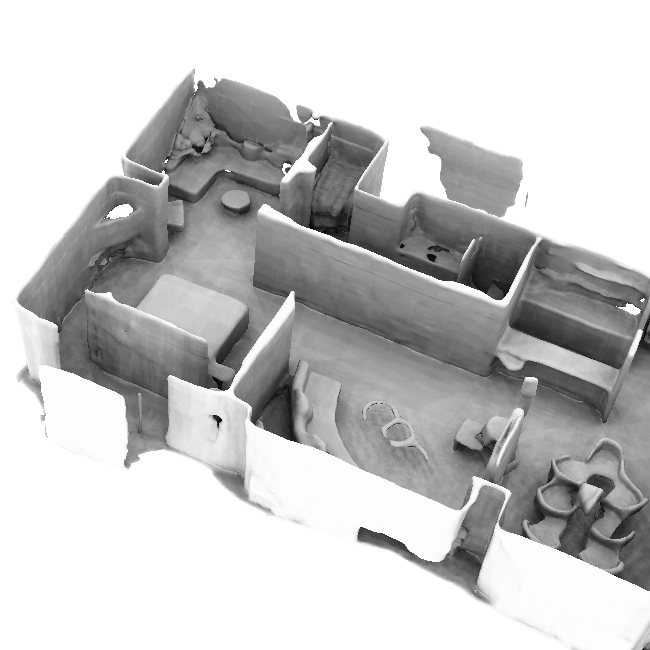}&
        \includegraphics[width=0.31\linewidth, trim=0 0 0 20, clip]{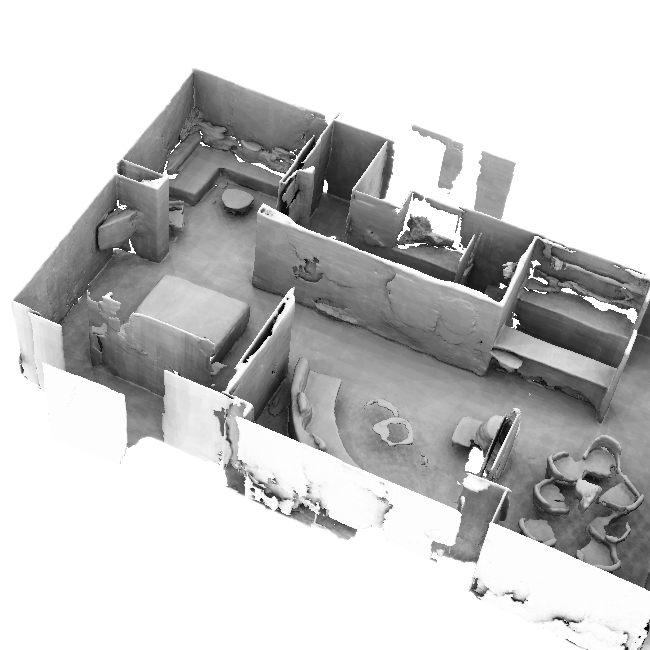}\\
        \includegraphics[width=0.31\linewidth]{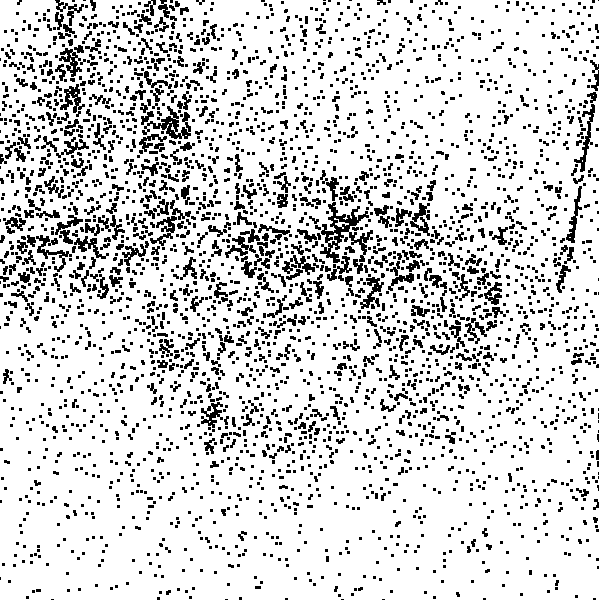} & 
        \includegraphics[width=0.31\linewidth]{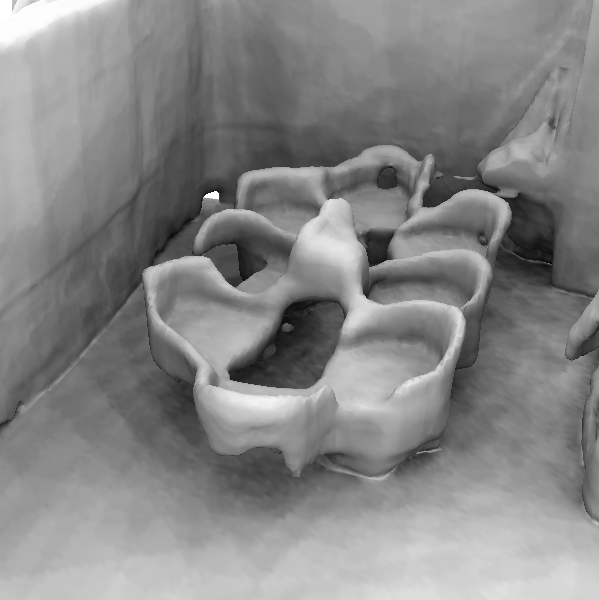} &
        \includegraphics[width=0.31\linewidth]{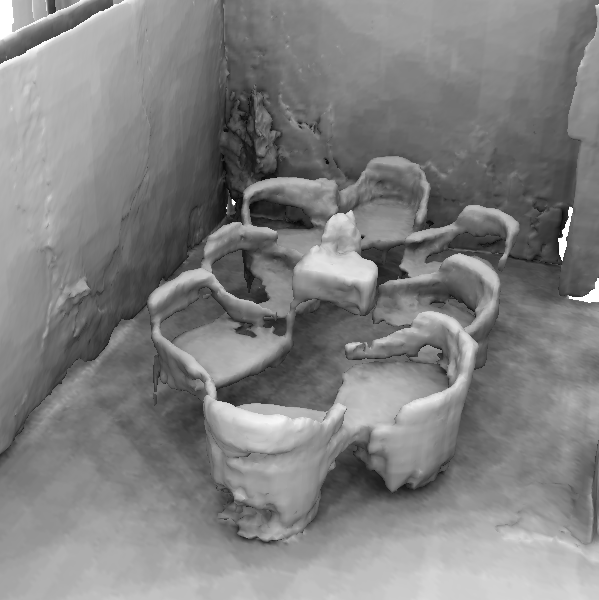} \\
        \multicolumn{1}{l}{\small (a) Scene 1} & \small 56\,min 40\,s & \small 10\,min 19\,s \\
        \midrule
        \includegraphics[width=0.31\linewidth, trim=0 0 0 20, clip]{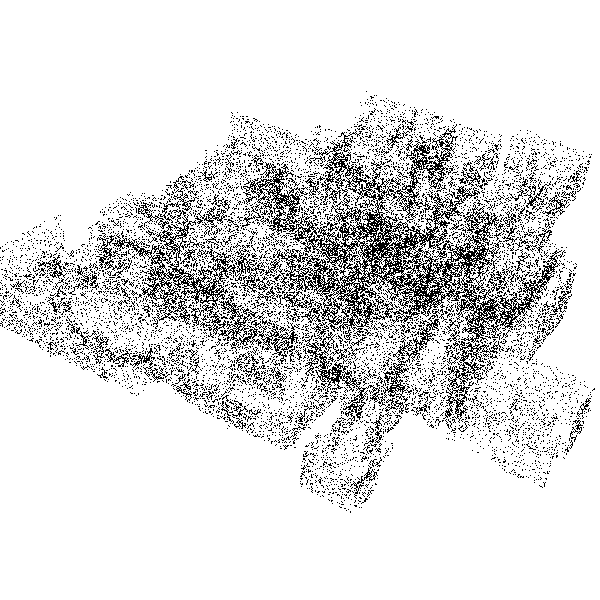}&
        \includegraphics[width=0.31\linewidth, trim=0 0 0 20, clip]{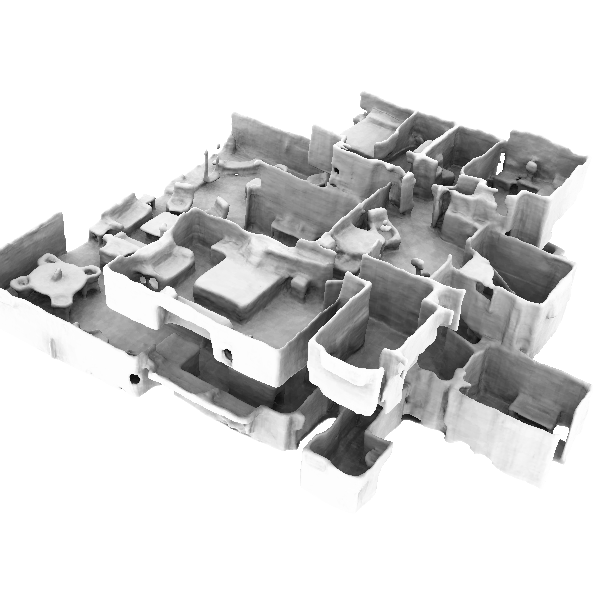}&
        \includegraphics[width=0.31\linewidth, trim=0 0 0 20, clip]{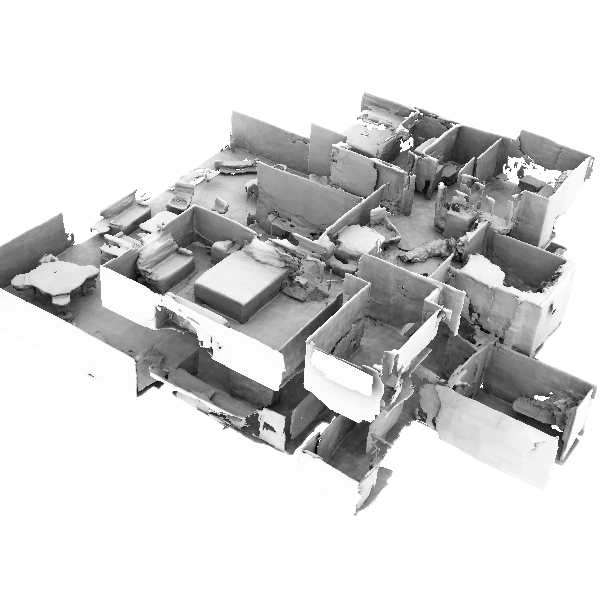}\\[-5mm]
        \includegraphics[width=0.31\linewidth, trim=0 0 0 50, clip]{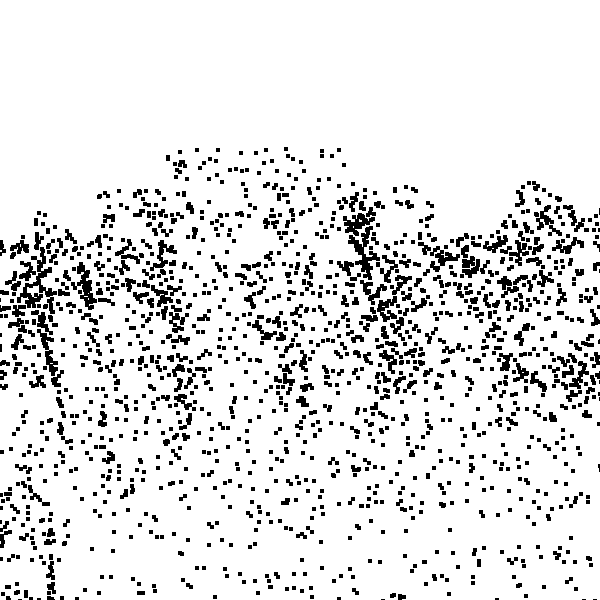}&
        \includegraphics[width=0.31\linewidth, trim=0 0 0 50, clip]{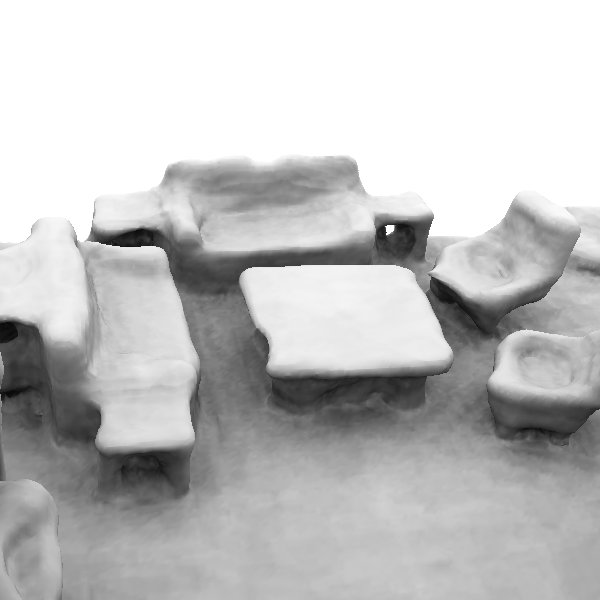}&
        \includegraphics[width=0.31\linewidth, trim=0 0 0 50, clip]{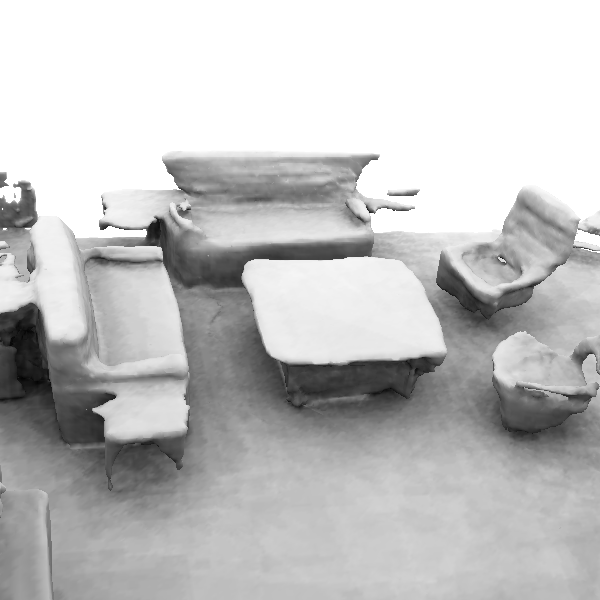}\\
        \includegraphics[width=0.31\linewidth, trim=0 0 0 20, clip]{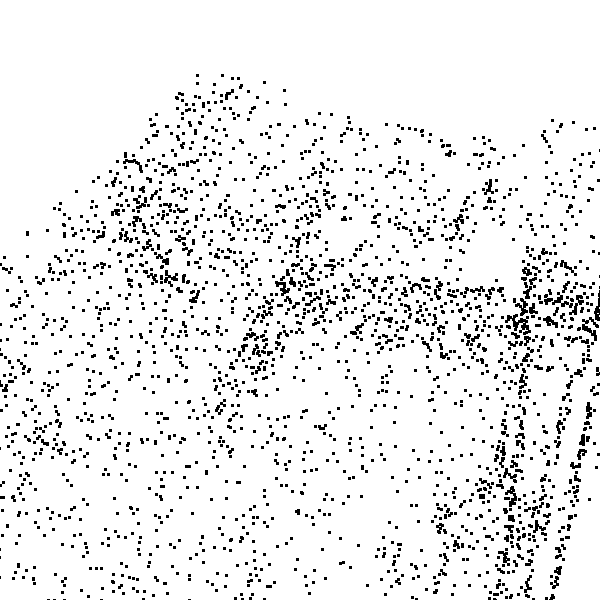}&
        \includegraphics[width=0.31\linewidth, trim=0 0 0 20, clip]{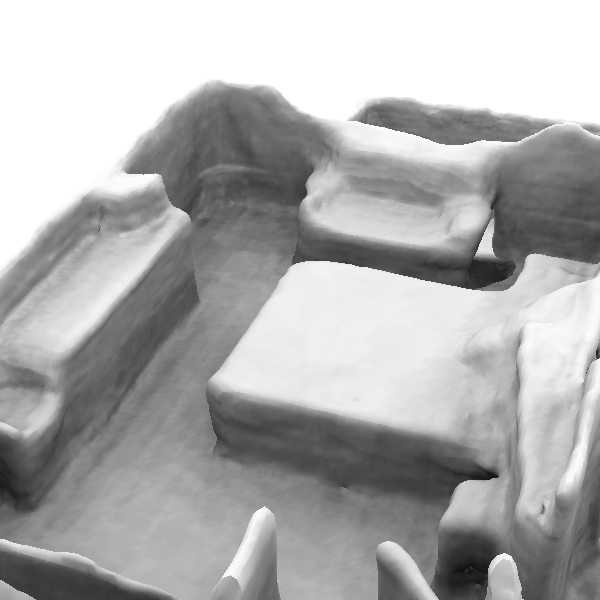}&
        \includegraphics[width=0.31\linewidth, trim=0 0 0 20, clip]{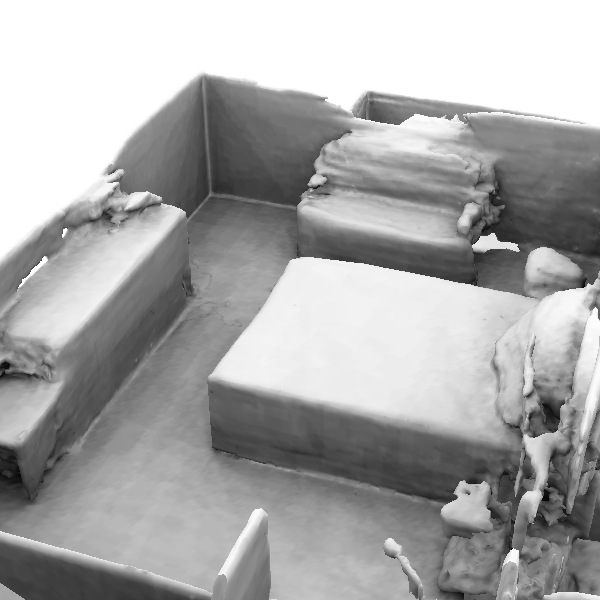}\\
        \multicolumn{1}{l}{\small (b) Scene 2} & \small 1\,h 38\,min & \small 17\,min 22\,s \\
        \midrule
    \end{tabular}
    \vspace{-8pt}
    \caption{\textbf{MatterPort3D.} \OURS\ trains on Synthetic Rooms 10k.}
    \label{fig:matterport}
    \vspace{-4mm}
\end{figure}

\begin{figure*}[t]
    \centering
    \includegraphics[width=\linewidth]{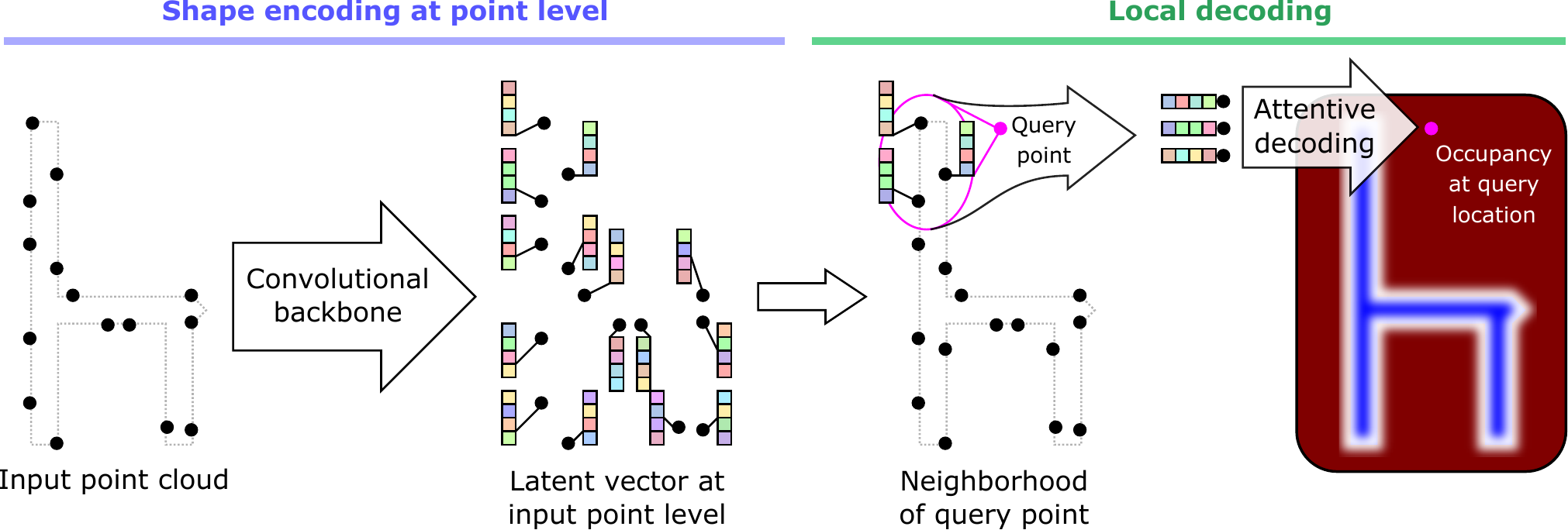}
    \caption{\textbf{Overview of our method (inference).} Given 3D points sampled on a surface, we construct latent vectors at each input point. Then, to estimate the occupancy of a given query point in space, we interpolate with inferred weights the relative occupancy scores in a neighborhood. Last, a mesh is reconstructed based on occupancy queries (white blur indicates uncertainty) using a form of Marching cubes\rlap.} \label{fig:pipeline}
\end{figure*}

Traditional 3D reconstruction approaches \cite{Berger2014EG} generally express the target surface as the solution to an optimization problem under some prior constraints. Possibly leveraging visibility or normal information, they are generally scalable to large scenes and offer a substantial robustness to noise and outliers \cite{Labatut2009CGF,Mullen2010CGF,Vu2012PAMI,Kazhdan2013SIGGRAPH,Zhou2019Sensors,Sulzer2021Scalable,Williams2021NeuralSplines,Poliarnyi2021Outofcore}. 
Although some try to cope with density variation \cite{Jancosek2011CVPR,Jancosek2014ISRN, BodisSzomoru2016ICPR}, a common limitation of these approaches is their inability to properly complete parts of the scene that are less densely sampled or that are missing (typically due to occlusions). A variety of hand-crafted priors try to address this completeness issue: local or global smoothness \cite{Lipman2007SGP}, decomposition into geometric primitives~\cite{Schnabel2009CGF} (in particular for piecewise-planar man-made environments \cite{Chauve2010CVPR,Lafarge2013EG,Boulch2014SGP,Nan2017ICCV, Bauchet2020TOG}) and structural regularities \cite{Pauly2008TOG,Li2011TOG}. Data-driven priors have also been explored, based on shape retrieval \cite{Gal2007CGF}, possibly with deformations \cite{Nan2012TOG}. But it remains limited in applicability.

To use richer priors, learning-based methods have been proposed, using explicit shape representations. Voxel-based approaches leverage a regular grid structure, extending 2D image-based techniques to 3D, but suffer from resolution limitations due to large memory consumption \cite{Maturana2015IROS,Choy2016ECCV,Wu2016NIPS}.  Directly generating a mesh with a neural network remains difficult \cite{Gkioxari2019ICCV} and is limited in practice to template deformation \cite{Groueix2018CVPR}.
Some forms of implicit representations have been used for point cloud generation, but providing much weaker geometrical and topological information \cite{Fan2017CVPR,Lin2018AAAI,Yang2019ICCV}. 

More success has been achieved with explicitly-designed implicit representations, where the network encodes a function $\mathbb{R}^3 \,{\rightarrow}\, \mathbb{R}$ expressing a volume occupancy \cite{Chen2019CVPR,Mescheder2019CVPR} or a distance to the  surface \cite{Michalkiewicz2019ICCV,Park2019CVPR}. 
Such models require no discretization and can address arbitrary topologies. More precisely, discretization only occurs at mesh generation stage, using an algorithm such as the Marching cubes \cite{Lorensen1987CG}. Yet, due to fully-connected architectures that lack translational equivariance, most existing approaches only operate on a single object and cannot apply to arbitrary scenes.

A few recent methods \cite{Jiang2020CVPR, Chibane2020CVPR, dai2020sgnn, Peng2020ECCV, Chibane2020Neural, Ummenhofer2021Adaptive}, however, obtain a form of translational equivariance via Convolutional Neural Networks (CNNs). At least in theory, they can thus scale to larger scenes, possibly benefiting both from local and non-local information. But they operate on a voxelized discretization whose vertices may be far from the input point cloud. They thus lose the direct connection with points sampled on the surface of objects. They are also suboptimal in that the features or latent vectors holding the occupancy or distance information are more or less uniformly distributed in space rather than focused where difficult decisions have to be made, i.e., near the surface. 

Our approach, based on point convolution, overcomes these issues. It is illustrated on Fig.\,\ref{fig:pipeline}. Our contributions \rlap{are:}
\begin{itemize}[topsep=-2pt, itemsep=0pt]
    \item We attach features representing the implicit function to input points. Not only does it preserve point positions until later processing stages, rather than abstract them away too soon, but it concentrates the information to learn where it matters the most: close to the surface.
    
    \item We compute features using point convolution, which yields a natural coverage and scalability to scenes of arbitrary size. (Rather than tailor yet another specific network architecture, we rely on a general point convolution back\-bone, which offers prospects for improvement when better point convolutions are designed.)
    
    \item Rather than relying on hand-designed forms of averaging, we extend prior learning to interpolation, which we apply to query-relative features rather than global features, as others do, as it leads to better results.
    
    \item We propose an efficient test-time augmentation to treat inputs of high density or large size.
    
    \item While simple, our approach outperforms other methods both on object and scene datasets, yielding finer details.  It is robust to domain shift (training on objects, testing on scenes) and faster than methods that overfit to a scene or infer from scratch for each query.
\end{itemize}

\section{Related work}
\label{sec:relwork}

\begin{figure*}[t]
    \centering
    \setlength{\tabcolsep}{7pt}
    \begin{tabular}{c|cccc|c}
    
        \includegraphics[width=0.13\linewidth]{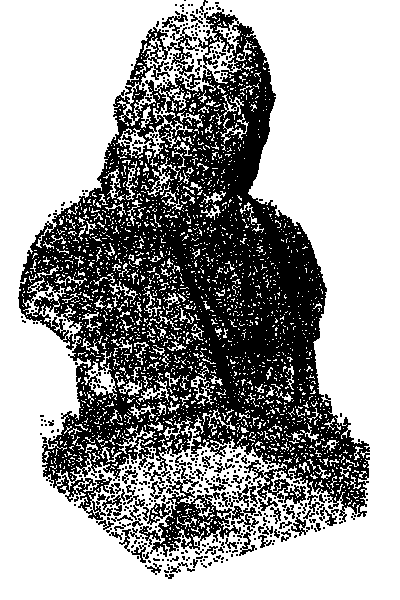}&
        \includegraphics[width=0.13\linewidth]{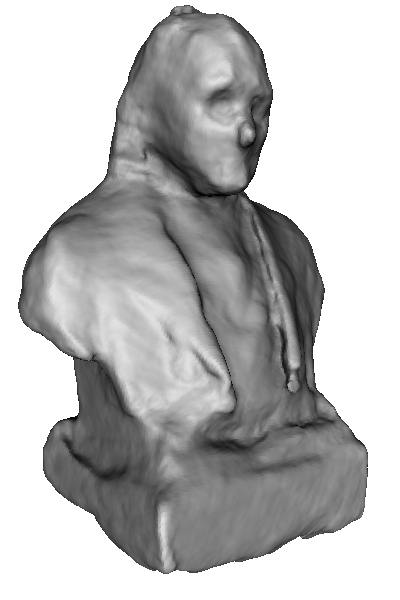} &
        \includegraphics[width=0.13\linewidth]{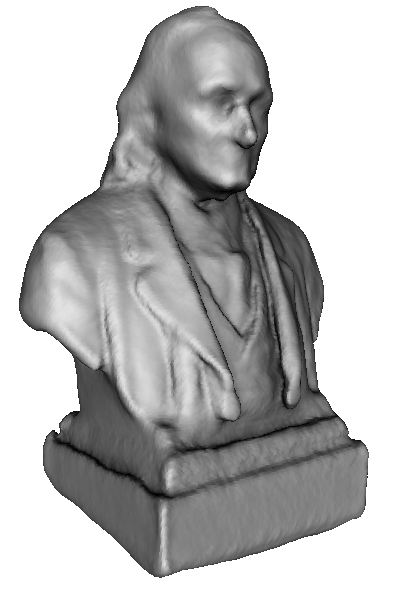} &
        \includegraphics[width=0.13\linewidth]{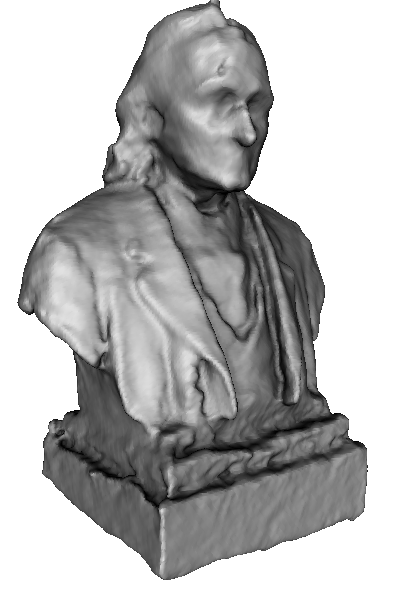} &
        \includegraphics[width=0.13\linewidth]{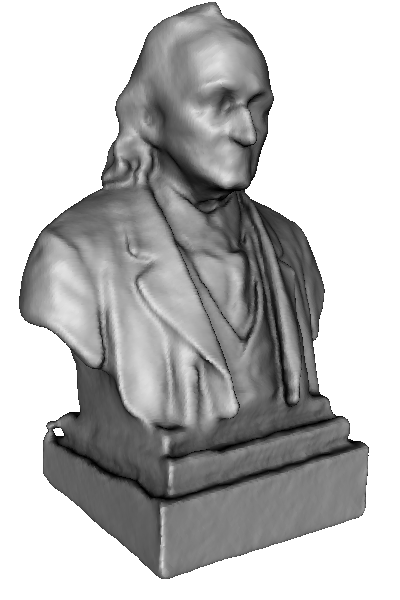}&
        \includegraphics[width=0.13\linewidth]{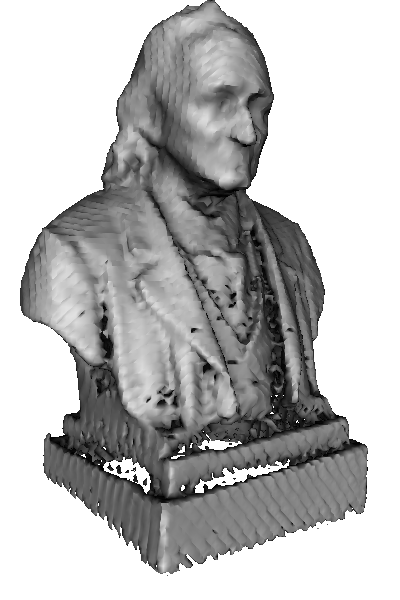}\\
        Input = 50k pts   & $\Ntrain{=}\Ntest{=}$3k & $\Ntrain{=}\Ntest{=}$3k & $\Ntrain{=}\Ntest{=}$10k & $\Ntrain{=}\Ntest{=}$10k & Points2Surf \\
        \multicolumn{1}{c}{}        &               & $\Nview{=}$10 &               & \multicolumn{1}{c}{$\Nview{=}$10} & 
    \end{tabular}
    \vspace{-6pt}
\caption{\textbf{Real World.} Model from Real World reconstructed by {\OURS} in different settings and by Points2Surf}
\label{fig:real_world}
\vspace{-3mm}
\end{figure*}

\subsection{3D representations}
\label{sec:3drepres}

\textbf{Voxels} have been a natural choice for learning to represent 3D volumes \cite{Wu2015CVPR, Maturana2015IROS, Choy2016ECCV, Wu2016NIPS, XieICCV2019, XieIJCV2020}. However, they come with a cubic complexity in space, leading to coarse discretizations due to memory constraints. Multi-scale refinement \cite{Dai2017CVPRb, Hane2017THREEDV} and sparsity-based octrees \cite{Tatarchenko2017ICCV, Riegler2017CVPR, Riegler2017THREEDV} only partly reduce the impact of conforming to a 3D grid.

\textbf{Points clouds} are also produced as a sparse 3D representation, with various density and sampling distribution \cite{Fan2017CVPR, Achlioptas2018ICML, Mandikal2018BMVC, Yuan2018THREEDV, Lin2018AAAI, Yang2019ICCV, Wen2020CVPR}. Point processing and generation do not suffer from the complexity and discretization induced by 3D grids; yet, the range of applications is limited regarding representing actual surfaces and volumes.

\textbf{Meshes} are a preferred representation for many uses, such as visualizations and simulations, but they are harder to directly produce from a neural network (vertex regression and face construction) \cite{Nash2020polygen}. Most existing approaches thus prefer to operate by deforming geometric primitives \cite{Groueix2018CVPR, Wang2018ECCV, Lin2019CVPR, Wen2019ICCV}, voxelized approximations \cite{Liao2018CVPR, Gkioxari2019ICCV} or learned templates \cite{Groueix2018ECCV, Kanazawa2018ECCV}. Rather than actually inferring vertices, a mesh can also be extracted from labels inferred on a Delaunay tetrahedralization \cite{Luo2021AAAI}.

\textbf{Implicit representations} rely on a neural network to model a function expressing the occupancy of a given 3D point \cite{Chen2019CVPR, Mescheder2019CVPR} or its distance to the surface, either signed \cite{Michalkiewicz2019ICCV, Park2019CVPR, Gropp2020ICML}, unsigned \cite{Chibane2020Neural} or sign-agnostic \cite{Atzmon2020CVPR,boulch2021needrop}. The signed or unsigned distance field (SDF, UDF) is often truncated (TSDF, TUDF) and estimated via a multi-layer perceptron (MLP). The isosurface can then be extracted from this occupancy or distance field with various methods such as Marching cubes \cite{Lorensen1987CG}. Whereas voxels, points and mesh vertices are intrinsically discrete representations, implicit representations offer a virtually infinite resolution. Moreover, while mesh-based approaches struggle to enforce watertightness, to limit self-intersections and to address complex topologies (non genus-0), meshes reconstructed from implicit representations are guaranteed to be watertight and have no self-intersections. Besides, they can easily model arbitrary complex topologies. These advantages may explain the recent success of this representation, including to model 3D shapes from images without 3D supervision \cite{Liu2019NIPS, Sitzmann2019NIPS, Niemeyer2020CVPR}, with texturing \cite{Oechsle2019ICCV} or specific rendering \cite{Liu2020CVPR}.
Departing from occupancy or distance fields, ShapeGF \cite{Cai2020ShapeGF} models a shape by learning the gradient field of its log-density, then samples points on high likelihood regions of the shape and meshes them.
Other work also study the decomposition of shapes and implicit surfaces into parts \cite{Genova2019ICCV, Genova2020CVPR, Paschalidou2020CVPR, Deng2020CvxNet, Jeruzalski2020ECCV, Tretschk2020ECCV}, possibly overfitting networks to generate or render a single object or scene \cite{Williams2019DeepGeometric, Sitzmann2019siren, liu2020neural, Zhao2021SignAgnostic, Takikawa2021NGLOD, Martel2021ACORN, Yan2021Continual}.

\textbf{Scalability}, however, is an issue for all these methods. While they can encode reasonably well one object or a class of objects, they cannot cope with the variability and size of an arbitrary scene involving several objects. Even considering a single object and assuming a powerful decoder, the encoding of a single or a few latent vectors hardly can develop into detailed shape information. Using periodic activation functions \cite{Sitzmann2019siren, Tancik2020Fourier} or adding a 2D convolutional component on input images \cite{Saito2019ICCV, Xu2019NIPS} helps, but is not enough.

A solution is to split the input points on a regular 3D grid and to optimize one latent vector per voxel \cite{Chabra2020DeepLS} (DeepLS), possibly from overlapping input patches. Patch splitting can also be irregular and optimization-driven to favor self-similarities, with a global post-optimization to flip inconsistent local signs \cite{Zhao2021SignAgnostic} (SAIL-S3). But whether these methods optimize only the latent vectors or a whole network as well, for patch decoding, they make surface reconstruction significantly slower, leading to reduced test sets.

Besides, these methods rely on fully-connected architectures whereas, we believe, convolutions, and in particular point convolutions \cite{Hua2018CVPR, Wang2018CVPR, Xu2018ECCV, li2018NeurIPS, Liu2019CVPR, Wu2019CVPR, Mao2019ICCV, Thomas2019ICCV, Boulch2020CG, Boulch2020ACCV}, are the key to scalability and increased details.

\begin{figure*}
    \centering
    \begin{tabular}{ccccc|c}
    & Input & SPR & Neural Splines & LIG & Ours\\
    \rotatebox{90}{\quad\quad 20 pts/m$^2$}&
    \includegraphics[width=0.15\linewidth]{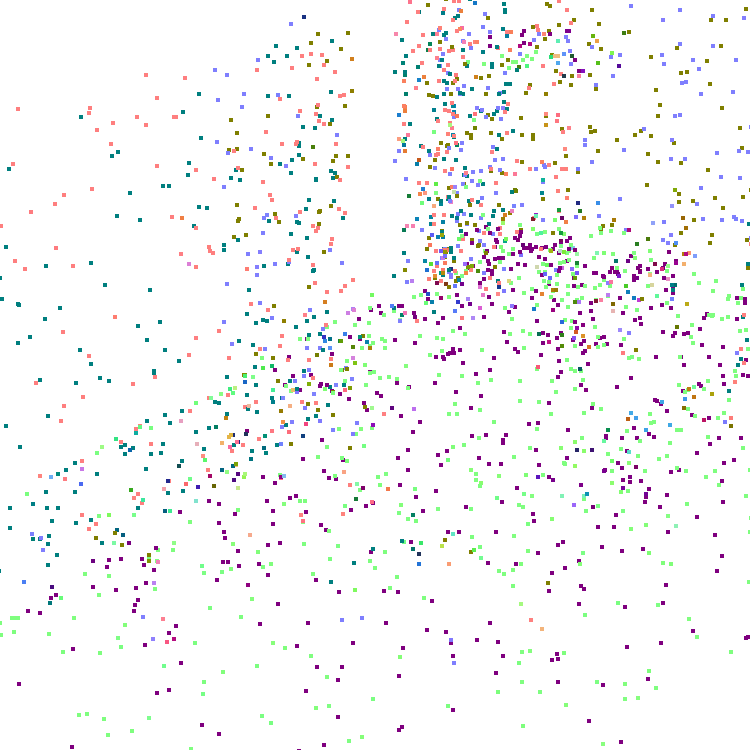}&
    \includegraphics[width=0.15\linewidth]{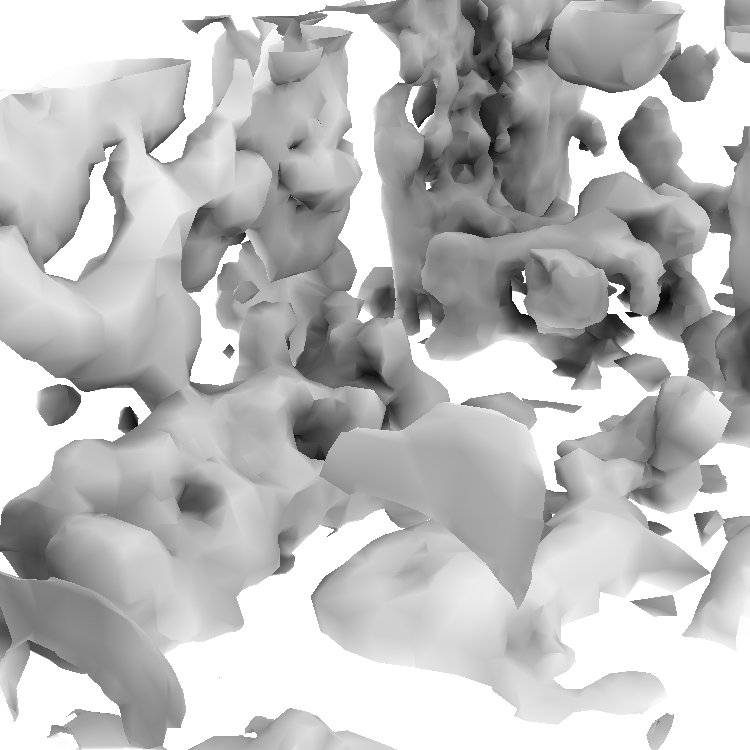}&
    \includegraphics[width=0.15\linewidth]{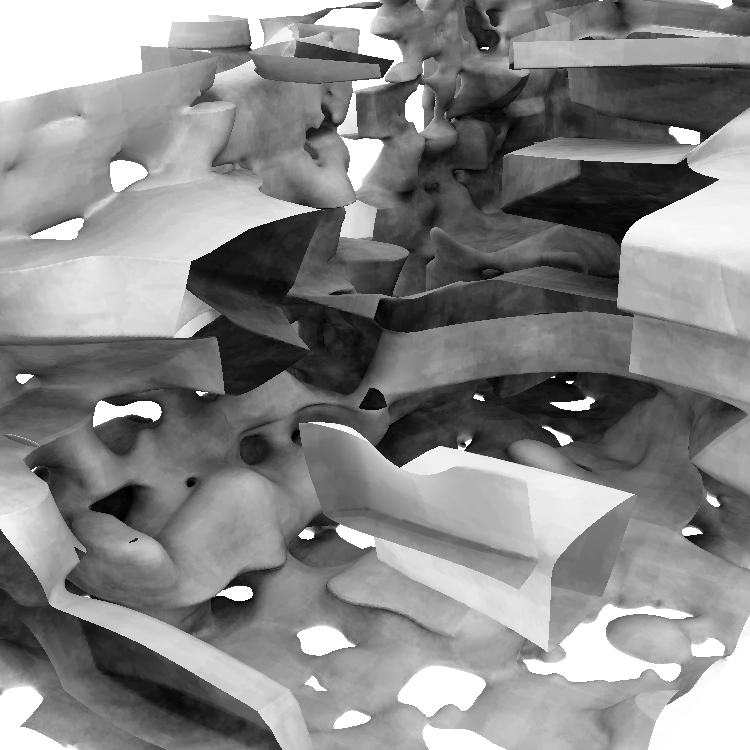}&
    \includegraphics[width=0.15\linewidth]{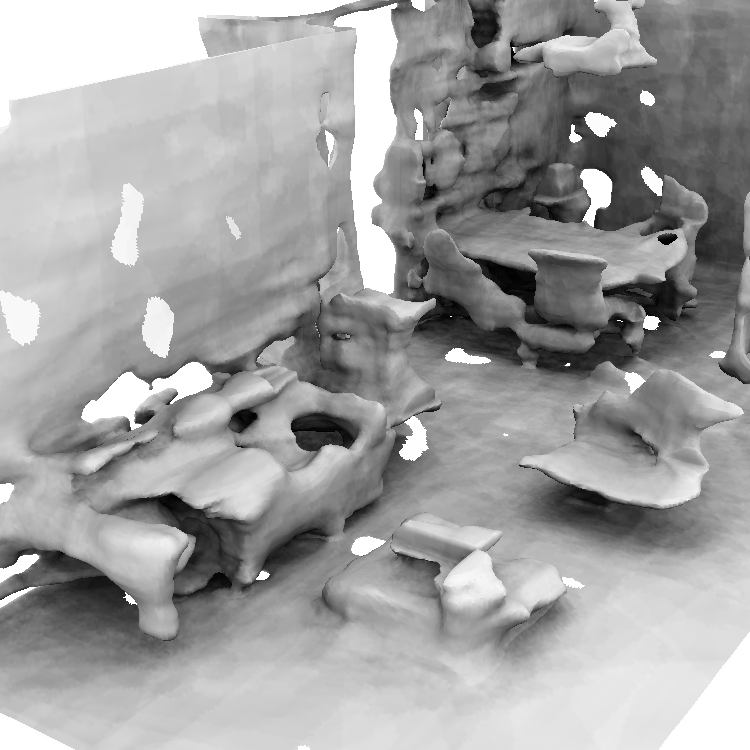}&
    \includegraphics[width=0.15\linewidth]{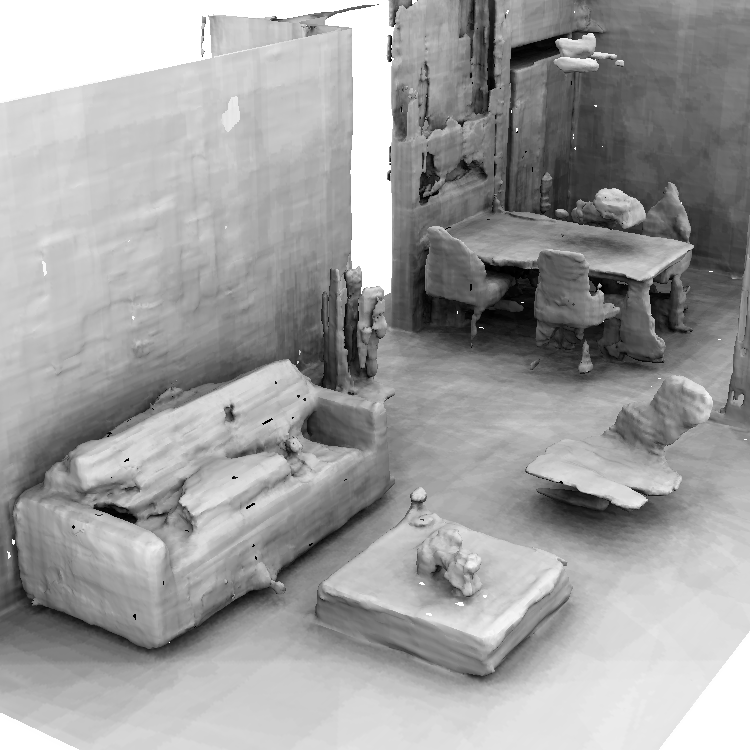}\\[-3pt]
    & & \small 38s & \small 5\,min 09\,s & \small 5\,min 22\,s & \small 17\,min\,32\,s\\[2pt]
    \rotatebox{90}{\quad\quad 100 pts/m$^2$}&
    \includegraphics[width=0.15\linewidth]{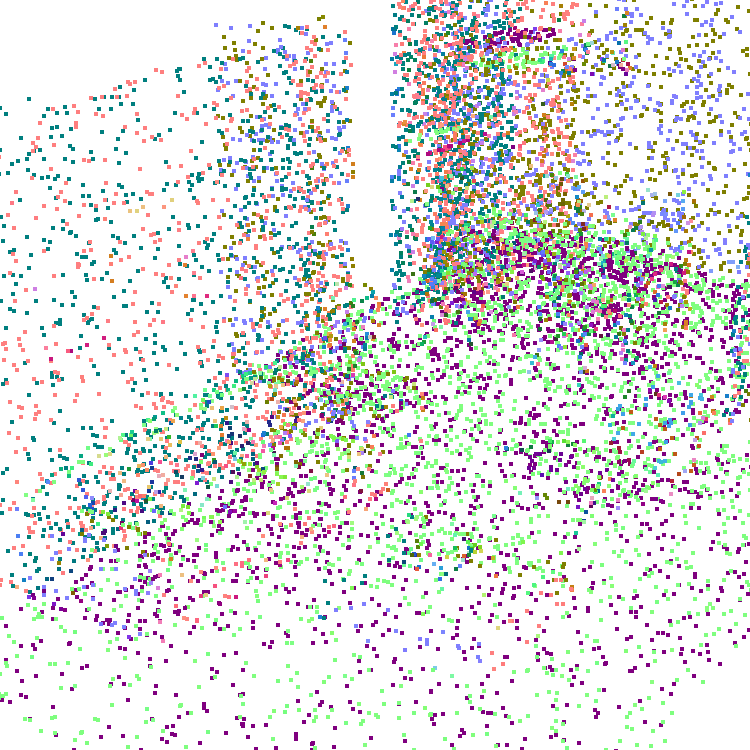}&
    \includegraphics[width=0.15\linewidth]{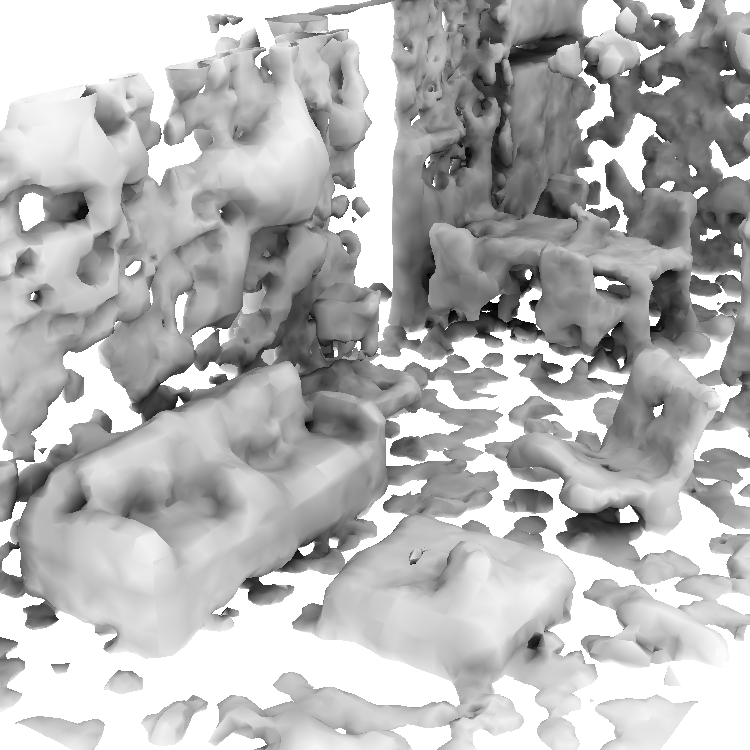}&
    \includegraphics[width=0.15\linewidth]{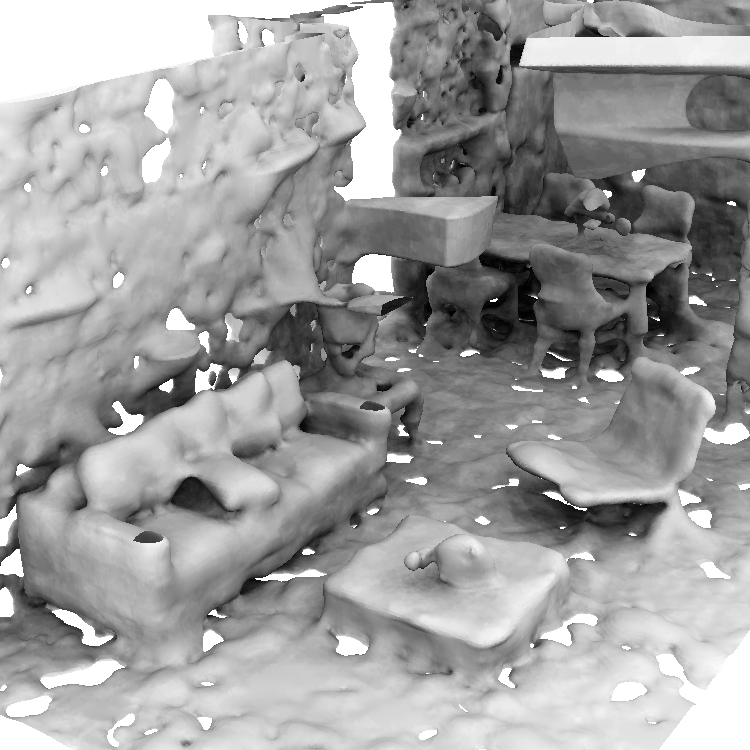}&
    \includegraphics[width=0.15\linewidth]{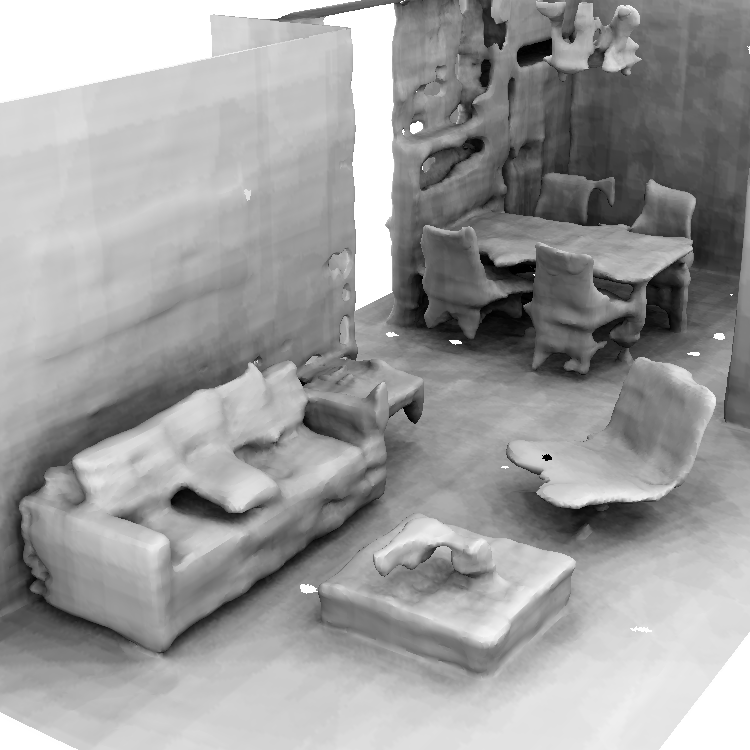}&
    \includegraphics[width=0.15\linewidth]{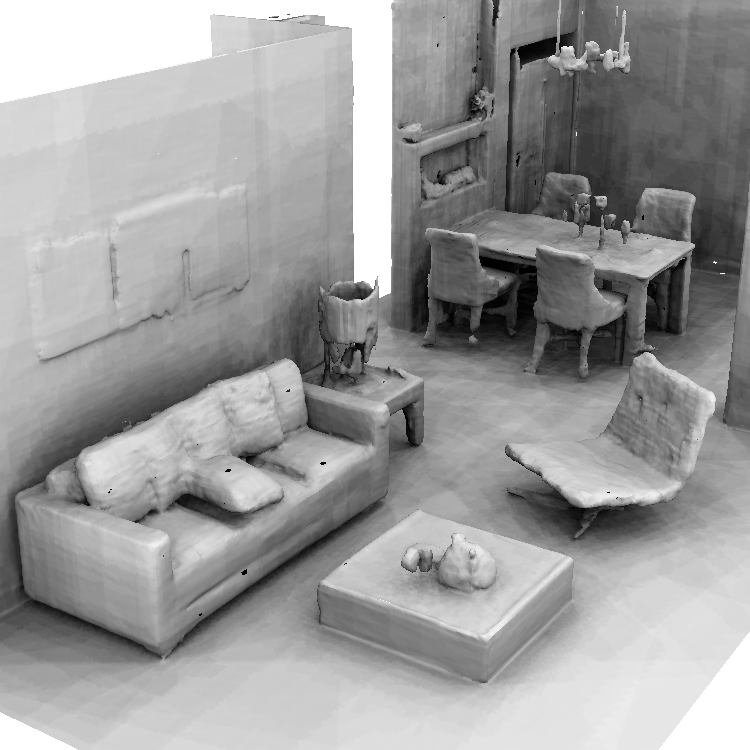}\\[-3pt]
    & & 1\small \,min 21\,s& \small 8\,min 11\,s & \small 5\,min 12\,s & \small 18\,min\,04\,s\\[2pt]
    \rotatebox{90}{\quad\quad 500 pts/m$^2$}&
    \includegraphics[width=0.15\linewidth]{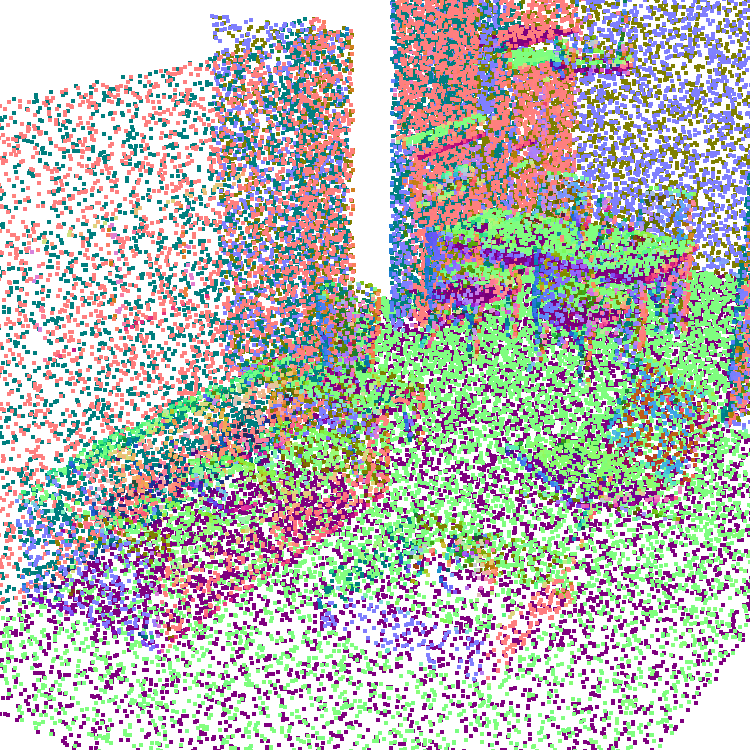}&
    \includegraphics[width=0.15\linewidth]{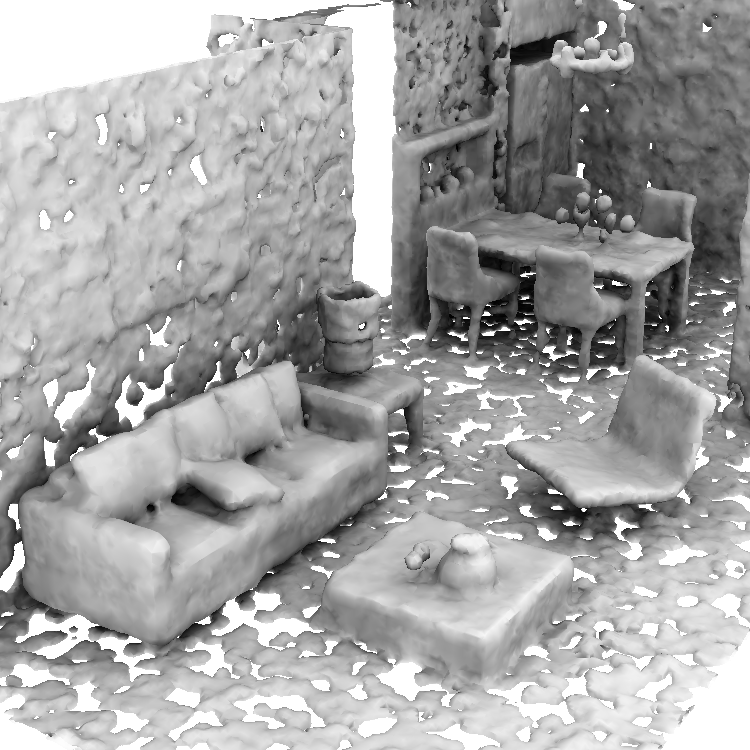}&
    \includegraphics[width=0.15\linewidth]{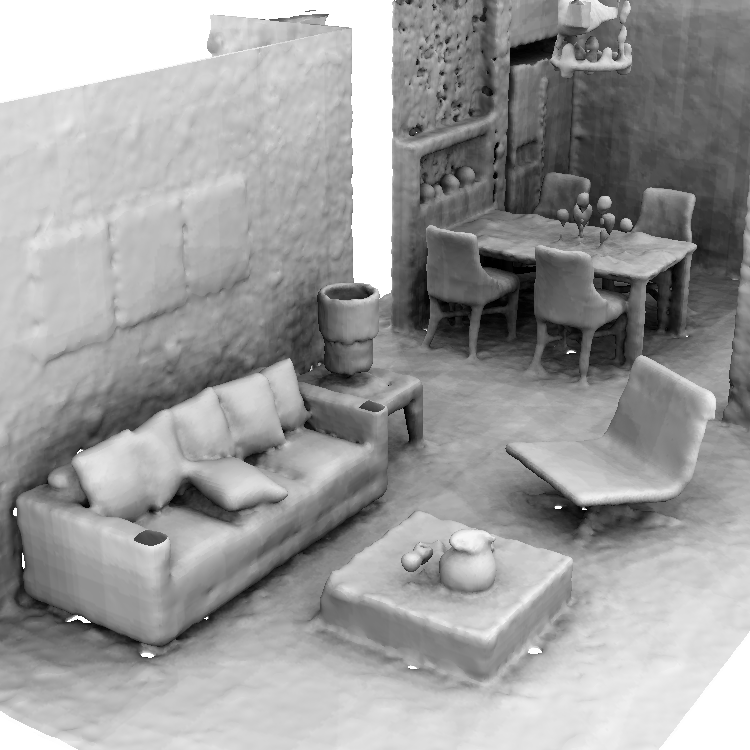}&
    \includegraphics[width=0.15\linewidth]{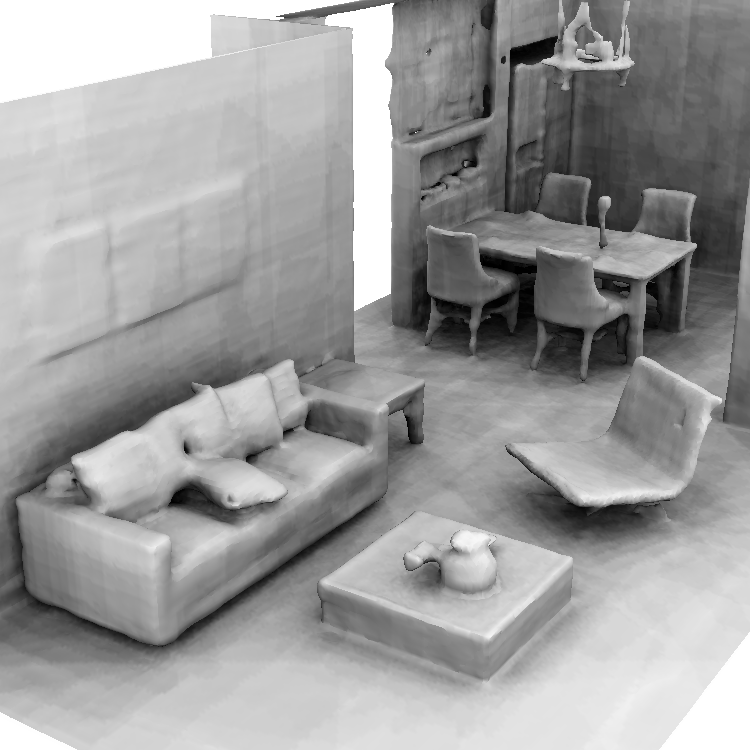}&
    \includegraphics[width=0.15\linewidth]{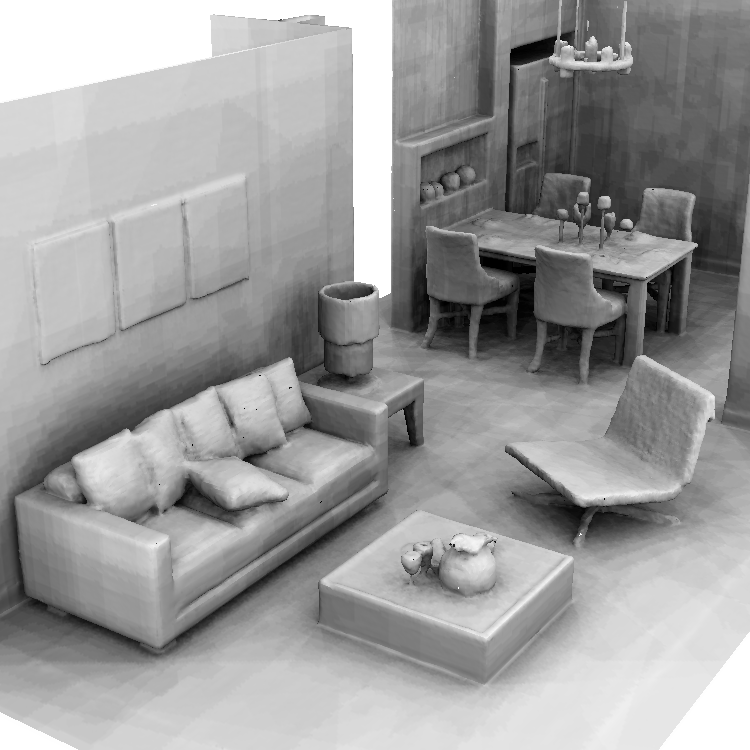}\\[-3pt]
    & & \small 3\,min 55\,s& \small 25\,min 17\,s & \small 5\,min 00\,s & \small 20\,min\,44\,s\\[2pt]
    \rotatebox{90}{\quad\quad 1000 pts/m$^2$}&
    \includegraphics[width=0.15\linewidth]{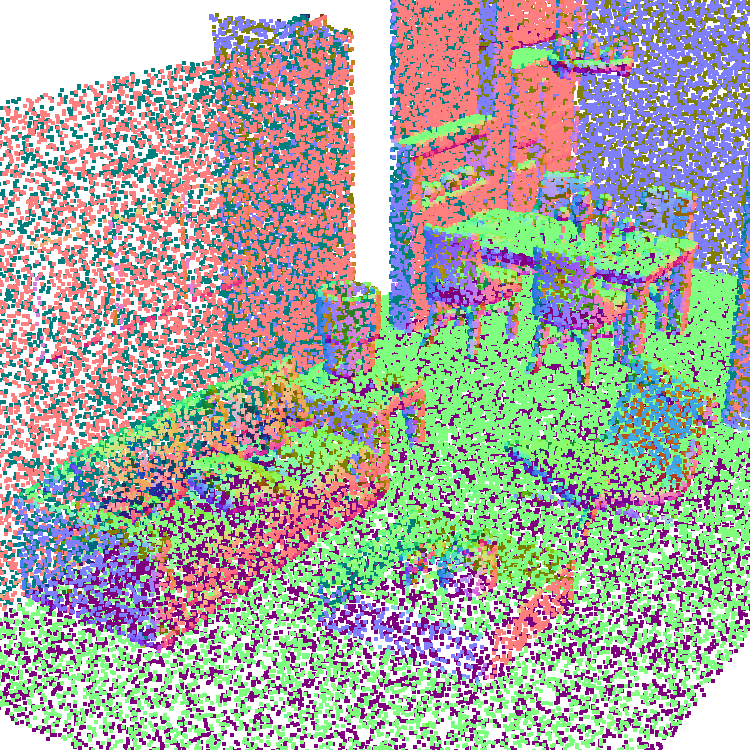}&
    \includegraphics[width=0.15\linewidth]{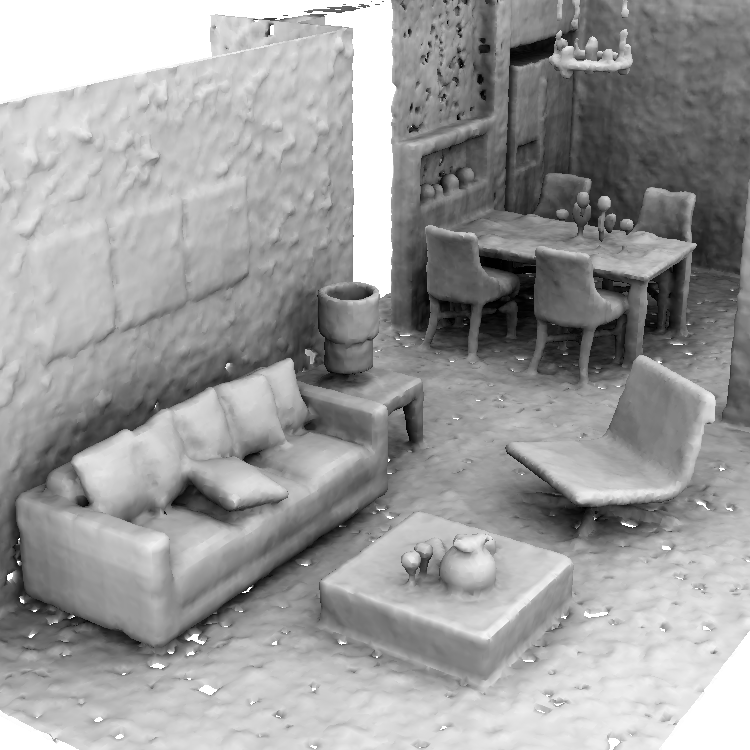}&
    \includegraphics[width=0.15\linewidth]{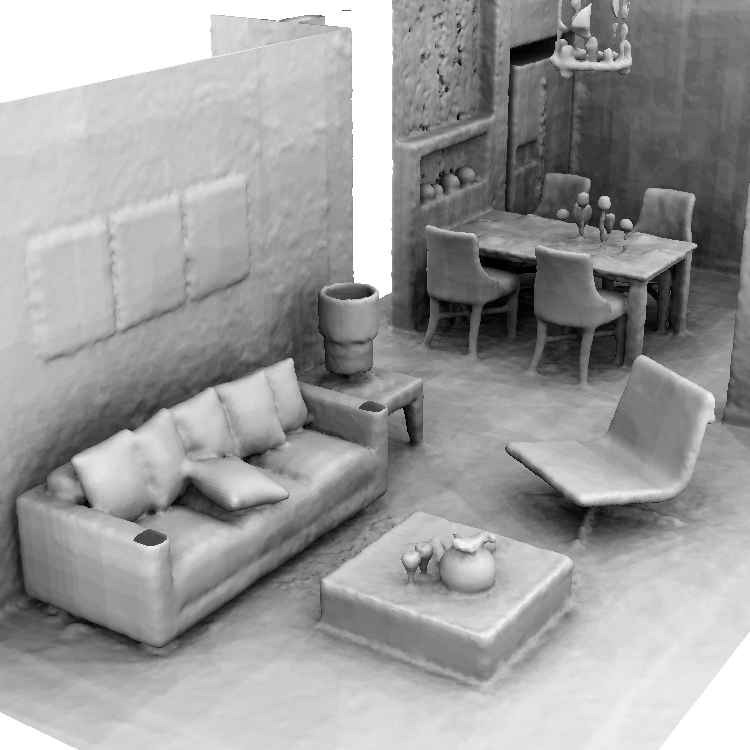}&
    \includegraphics[width=0.15\linewidth]{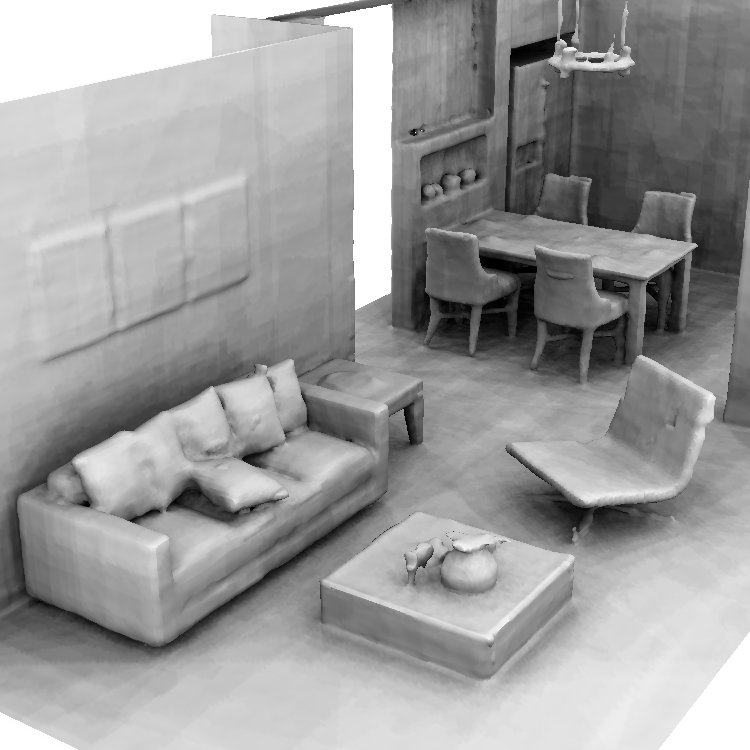}&
    \includegraphics[width=0.15\linewidth]{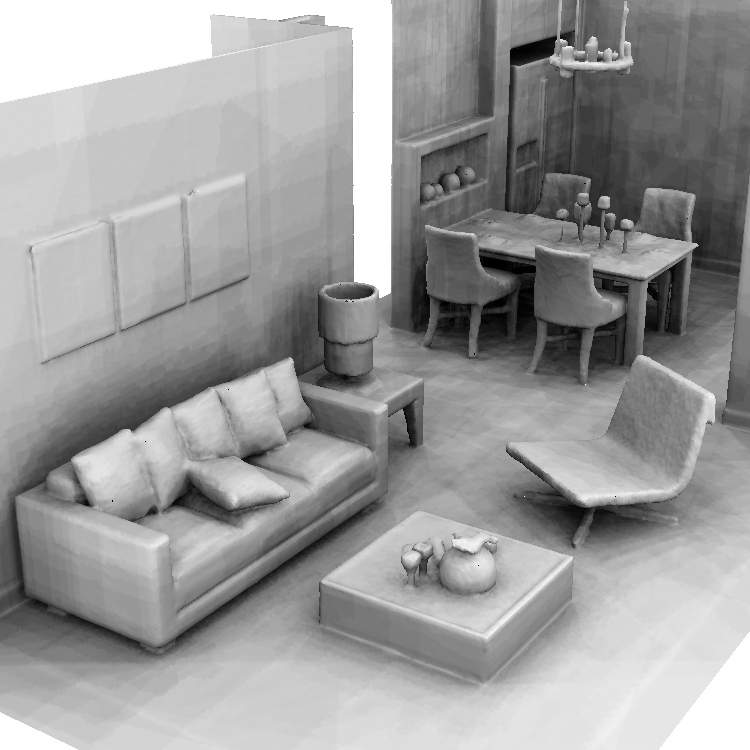}\\[-3pt]
    & & \small 5\,min\,05\,s & \small 46\,min\,44\,s & \small 5\,min\,08\,s & \small 23\,min\,59\,s
    \end{tabular}
    \vspace{-8pt}
    \caption{\textbf{SceneNet.} Partial view of a full scene. The color on point clouds indicates the orientation of normals.} 
    \label{fig:scenenet}
    \vspace{-3mm}
\end{figure*}

\subsection{Convolutions for implicit representations}
\label{sec:convmethods}

LIG \cite{Jiang2020CVPR} divides the input point cloud along a regular 3D grid to create 3D patches and capture local geometric shapes shared by several objects at a medium scale. For each of these patches, a 3D CNN then computes a local feature vector, which goes through a reduced IM-NET \cite{Chen2019Learning} for SDF decoding. However, later on, only the learned decoder is exploited; no local embedding is inferred. Given an input point cloud, latent vectors on the grid are optimized from scratch to minimize an objective function similar to the loss used for training. LIG additionally requires to be provided with oriented normals to make use of points known to be inside or outside the shape. This, however, may introduce artificial back-faces, which can partly be addressed in a postprocessing stage.
In contrast, we can work without normals, we directly operate with convolutions on surface points rather than on a regular grid, and we directly use inferred embeddings without any heavy optimization.

IF-Net \cite{Chibane2020CVPR} introduces a multi-scale pyramid of 3D convolutional encoders aligned on a discrete voxel grid and trained on voxels at different scales. The occupancy of a query point is decided by a decoder taking as input the interpolated features extracted at this point for each pyramid level.
In contrast, we do not discretize into voxels; we use point cloud convolution. Also, we learn how to interpolate the latent vectors rather than use a basic trilinear interpolation. Last, we provide results on scenes, not just on objects.

NDF \cite{Chibane2020Neural} uses the same multi-scale encoding as IF-Net but relies on a UDF rather than occupancy for decoding. It allows the generation of very dense points clouds that can directly be meshed into possibly open surfaces.

SG-NN \cite{dai2020sgnn} uses a sparse 3D convolution \cite{choy20194d} to learn a TSDF in a self-supervised setting, training for completion from partial scans.
In contrast, we use point convolution and infer occupancy rather than SDF, which is easier to learn.

ConvONet \cite{Peng2020ECCV} also uses a grid-based convolution, training an autoencoder that predicts occupancy. (It generalizes ONet \cite{Mescheder2019CVPR}, which only uses a single encoding and full connection.) For input point clouds, the encoder is a shallow PointNet \cite{Qi2017CVPR} operating on points rather than on a voxelized discretization, and the decoder is a 3D U-Net \cite{Cciccek2016MICCAI}. The occupancy of a 3D point is inferred from a trilinear interpolation of grid features. Besides 3D convolution, variants based on a combination of 2D convolutions in a few spatial directions are proposed. DP-ConvONet \cite{Lionar_2021_WACV} is a variant that considers a dynamic family of such directions.
SA-ConvONet \cite{tang2021sign} overfits a pre-trained ConvONet model on the input using a sign-agnostic optimization of the implicit field. It improves accuracy at the cost of computation time. 

\begin{figure*}
    \centering
    \includegraphics[width=\linewidth]{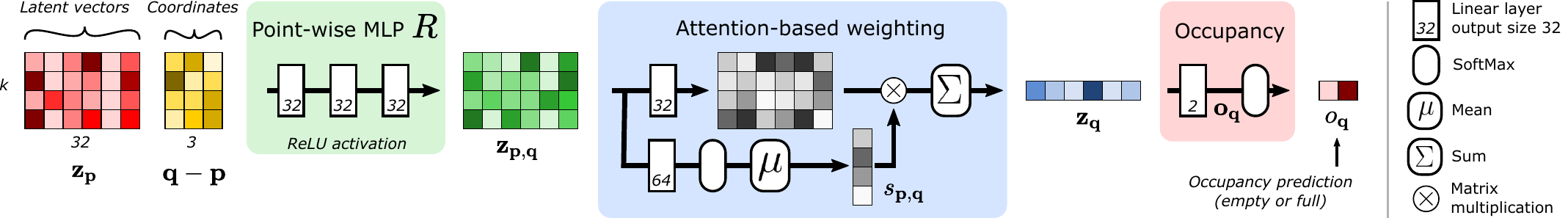}
    \vspace{-6mm}
    \caption{\textbf{Architecture.} The latent vectors $\latentv_{\point}$ (red squares) produced by the convolution-based encoder $\encoder$ of $k$ neighboring points $\point$ of a query point $\qpoint$ are:
    (1)~augmented with the relative query position $\qpoint \,{-}\, \point$ (yellow squares), 
    (2)~re-encoded with a 3-layer point-wise MLP $R$ (green frame) into relative latent vectors~$\latentv_{\point,\qpoint}$ (green squares), 
    (3)~combined (blue frame) with inferred weights $\significance_{\point,\qpoint}$ (gray squares) into a latent vector $\latentv_\qpoint$ (blue squares),
    (4)~decoded with a linear layer $D$ (pink frame) into occupancy logits $\occupancy_{\qpoint}$ and probablities $\occupancyp_{\qpoint}$ (pink squares).
    } 
    \label{fig:decoder}
    \vspace{-2mm}
\end{figure*}

As inference applies to grids, whose vertices or centers may be far from input points, the above methods lose the direct connection with the input surface samples. They are also suboptimal in that the latent vectors holding the information are uniformly distributed in space rather than concentrated where it matters the most, i.e., near the surface.
To address these issues, we use point convolution and compute latent vectors at each input point. We then interpolate occupancy decisions of nearest neighbors using learned weights.

AdaConv \cite{Ummenhofer2021Adaptive} uses point convolution like us but aggregates multi-scale information on an adaptive voxel grid, while we attach features to points, closer to the surface. Besides, it requires oriented normals, contrary to us.

RetrievalFuse \cite{Siddiqui2021RetrievalFuse} splits a scene along a regular grid and encodes each 3D chunk as a latent vector via convolutional layers. But rather than using them for decoding, it retrieves similar chunks from the training set and combines their distance field to create a surface, enhancing the completion capability.
In contrast, we are fully convolutional and the implicit function is directly obtained by interpolating inferred features, without the need to maintain the dataset samples used for training and with more generalization capacity.

Points2Surf \cite{Erler2020Points2Surf} collects, for each query point, both a patch of neighbors (which gives a convolution flavor) and globally-sampled input points to help to provide a sign to the local distance field. The local patch and the global sub-sampling go through an MLP to create latent vectors that are concatenated and decoded into a signed distance.
In contrast, we directly get non-local information as our receptive field is much larger. Besides, we are faster as we only compute a limited number of latent vectors (one per input point) that we later use for interpolation given a query point, while Points2Surf samples local+global points and goes through the whole encoder for each query point, i.e., a large number of times, that grows with the Marching-cubes resolution.

To infer occupancy or distance of a query point, methods that compute several latent vectors for a single object or scene either select the most appropriate latent vector to decode, typically in a multi-scale grid \cite{Ummenhofer2021Adaptive}, or interpolate the latent vectors of query neighbors \cite{Chibane2020CVPR, Jiang2020CVPR, Peng2020ECCV, Chibane2020Neural, Lionar_2021_WACV, tang2021sign}. We perform interpolation too, based on features computed on input points. However, given a query point, we do not interpolate the features themselves but the occupancy logits, as our experiments shows it leads to better results. Besides, we use a learned interpolation rather than the usual tri-linear interpolation \cite{Chibane2020CVPR, Jiang2020CVPR, Peng2020ECCV, Chibane2020Neural, Lionar_2021_WACV, tang2021sign} or the inverse-distance distance weighting \cite{Qi2017NIPS}. Although different in nature, learning has also been used in \cite{Siddiqui2021RetrievalFuse} to blend retrieved chunks.

\section{Our method}

\textbf{Goal.} Given as input a set of  3D points $\pointcloud$ sampled on a surface, possibly with noise, our goal is to construct a continuous function $\occupancyf:\Real^3 \rightarrow [0,1]$ indicating the probability of occupancy $\occupancyp_\qpoint \,{=}\, \occupancyf(\qpoint)$ at any given query point $\qpoint \in \Real^3$. We learn this function with a neural network using data consisting of point clouds sampled in the whole space and labeled with $0$ (in empty space) or $1$ (within the shape). The surface of the shape can then be extracted as the isosurface of the implicit function~$\occupancyf$ with occupancy level~0.5.

\textbf{Overview.} Our method consist of the following steps:
\begin{enumerate}[itemsep=-2pt,topsep=-1pt]

    \item We encode input points $\point \,{\in}\, \pointcloud$ into latent vectors $\latentv_\point$.
    
    \item Given an arbitrary query point $\qpoint$, we consider a neighborhood $\neighbors_\qpoint$ of input points in $\pointcloud$ to  interpolate from.

    \item For each neighbor $\point \,{\in}\, \neighbors_\qpoint$, we construct a relative latent vector $\latentv_{\point,\qpoint}$ from $\latentv_\point$ and local coordinates $\qpoint \,{-}\, \point$.\label{decodefirst}
    
    \item We extract significance weights $\significance_{\point,\qpoint}$ to sum the relative latent vectors $\latentv_{\point,\qpoint}$: $\latentv_\qpoint = \sum_{\point \in \neighbors_\qpoint} \significance_{\point,\qpoint} \,\latentv_{\point,\qpoint}$.
    
    \item We decode the resulting feature vector $\latentv_\qpoint$ as two full-empty logits $\occupancy_\qpoint$, and turn them into probabilities~$\occupancyp_\qpoint$.
    \label{decodelast}
    
\end{enumerate}\vspace{3pt}
These steps, illustrated on Figure~\ref{fig:decoder}, are detailed below.

\textbf{Absolute encoding.} 
A point convolution first produces a latent vector $\latentv_\point \,{=}\, \encoder(\point)$ for each input point $\point\,{\in}\,\pointcloud$. The encoder $E$ can be implemented by any point cloud segmentation backbone, only changing the last layer to yield a vector of some chosen dimension~$n$ as the size of vectors $\latentv_\point$. (In our experiments, the convolution backbone is FKAConv \cite{Boulch2020ACCV} and $n\,{=}\,32$.) To also use normals (optionally), the input points are just augmented with the 3 normal coordinates.

\textbf{Query neighborhood.} 
Given an arbitrary query point~$\qpoint$ (when training or to predict occupancy at test time), we construct a set of neighbors $\neighbors_\qpoint$ from input points $\pointcloud$. (In our experiments, $\neighbors_\qpoint$ is the $k$ nearest neighbors of $\qpoint$, with $k\,{=}\,64$.)

\textbf{Relative encoding.} 
We augment the latent vector $\latentv_\point$ of
each neighbor $\point \,{\in}\, \neighbors_\qpoint$ with the local coordinates $\qpoint \,{-}\, \point$ of query point $\qpoint$ relatively to~$\point$. These augmented latent vectors are then processed by an MLP $\relativizer$ to produce relative latent vectors $\latentv_{\point,\qpoint} = \relativizer(\latentv_\point \,\|\, \qpoint\,{-}\,\point)$, where $\|$ is the concatenation. (In our experiments, $\latentv_\point$ and $\latentv_{\point,\qpoint}$ have size $n\,{=}\,32$.)

\textbf{Feature weighting.}
As PRNet \cite{Wang2019NeurIPS}, we observe that the norm of embeddings $\latentv_{\point,\qpoint}$ tends to correlate with their significance, hinting how much an input point~$\point$ matters for deciding the occupancy of query point~$\qpoint$, given $\point$'s neighbors and the position of $\qpoint$ w.r.t.~$\point$.
We use it to infer significance weights for relative latents vectors $\latentv_{\point,\qpoint}$. Concretely, we use an attention mechanism (blue frame in Fig.\,\ref{fig:decoder}): The relative embeddings $\latentv_{\point,\qpoint}$ go through a linear layer parameterized by a weight vector~$\weightv$, also of size~$n$, producing relative weights $\weight_{\point,\qpoint} \,{=}\, \weightv \,\odot\, \latentv_{\point,\qpoint}$, that are normalized by softmax over $\neighbors_\qpoint$ into positive interpolation weights $\significance_{\point,\qpoint}$ summing to~$1$.
We actually use a multi-head strategy to obtain a form of ensembling. We learn $h$ independent linear layers, parameterized by $h$ corresponding weight vectors $(\weightv_i)_{i=1..h}$, producing $h$ relative weights $\weight_{\point,\qpoint,i} \,{=}\, \weightv_i \,\odot\, \latentv_{\point,\qpoint}$, that are finally softmaxed as $\significance_{\point,\qpoint,i}$ and averaged as $\significance_{\point,\qpoint} \,{=}\, \frac{1}{n} \sum_{i} \significance_{\point,\qpoint,i}$. (In our experiments, we use $h\,{=}\,64$.)

\textbf{Interpolation.} The feature vector 
$\latentv_\qpoint$ at query point $\qpoint$ is interpolated from the relative latent vectors $\latentv_{\point,\qpoint}$ of neighbors $\point$, as the weighted sum $\latentv_\qpoint \,{=} \sum_{\point \in \smash{\neighbors_\qpoint}} \!\significance_{\point,\qpoint} \,\latentv_{\point,\qpoint}$.

\textbf{Decoding.} A linear layer $\decoder$ decodes the feature vector 
$\latentv_{\qpoint}$ into occupancy scores $\occupancy_{\qpoint} \,{=}\, \decoder(\latentv_{\qpoint})$, which is a two-logit vector classifying position $\qpoint$ as occupied or not, that is then turned via softmax into occupancy probabilities~$\occupancyp_{\qpoint}$.

\textbf{Loss function.} To train the network, we use a cross-entropy loss that penalizes wrong occupancy predictions. Please note that using a binary cross-entropy, like in IF-Net \cite{Chibane2020CVPR} or ConvONet \cite{Peng2020ECCV}, leads to identical 
results.

\section{Refinements}

\textbf{Adapting to high density.}
We train our network with a fixed number $\Ntrain$ of input points for easy mini-batching. (In our experiments, $\Ntrain \,{=}\,$3k  or  10k.) 
At test time, if the surface is more densely sampled, the receptive field of the backbone may lack enough global context to decide which side of the surface is full or empty, unless oriented normals are also provided with points. 
A way to broaden enough the receptive field is to downsample the input point cloud, but it then naturally leads to a loss of details.

To reduce this effect, we rely on test-time augmentation (TTA) \cite{krizhevsky2012imagenet}, which can be seen as a form of ensembling: we average several runs on different subsamples.
However, aggregating final results, as often done in TTA \cite{Shanmugam2021BetterAggregation}, would be very time consuming in our case as we would have to do it to answer the occupancy of each query, basically multiplying the inference running time by the number of subsamples.

Instead, we perform TTA at latent vector level, thus running several times only the first step of our approach (absolute encoding), before query decoding. It depends on the number of input points (to attach a latent vector on), rather than on the number of query points, which is much larger.
Concretely, we randomly create enough subsamples so that each point $\point \,{\in}\, \pointcloud$ is seen at least $\Nview$ times, and average each $\latentv_\point$ over all samples. (In experiments, $\Nview \,{=}\, 10$.) The subsamples are randomly generated by sequentially picking a point $\point \,{\in}\, \pointcloud$ with a priority that is the opposite of the number of times $\point$ appears in previous subsamples.

\begin{figure}[t]
    \centering
    \setlength{\tabcolsep}{3pt}
    \begin{tabular}{cccc}
        GT & Input 
        & ConvONet & \OURS\ (ours)  \\[-3pt]
        \includegraphics[width=0.23\linewidth]{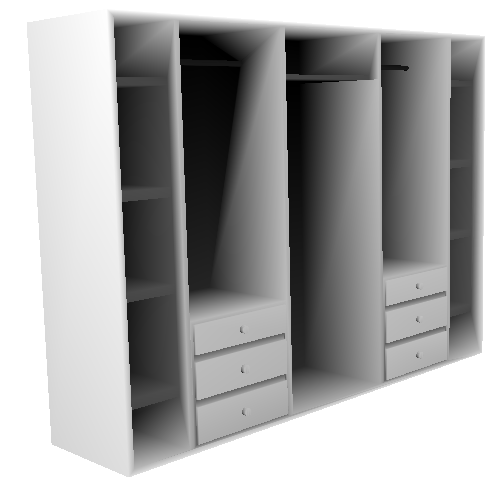}&
        \includegraphics[width=0.23\linewidth]{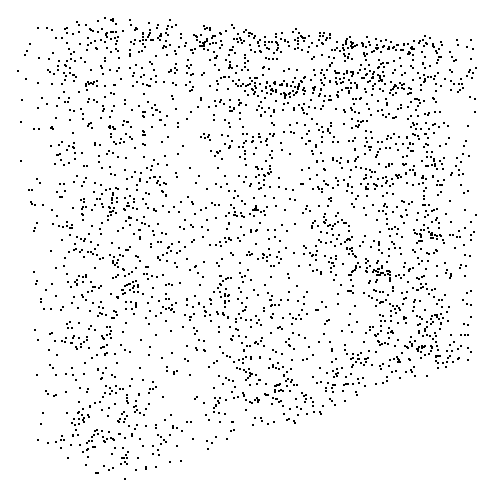}&
        \includegraphics[width=0.23\linewidth]{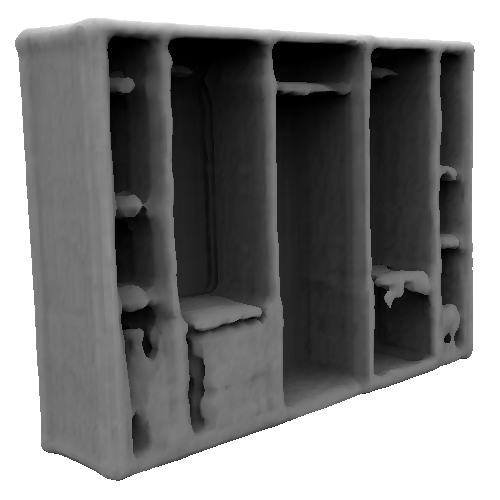}&
        \includegraphics[width=0.23\linewidth]{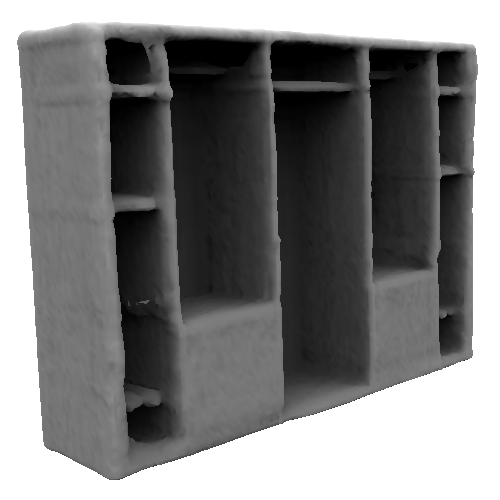}
        \\
        \includegraphics[width=0.23\linewidth]{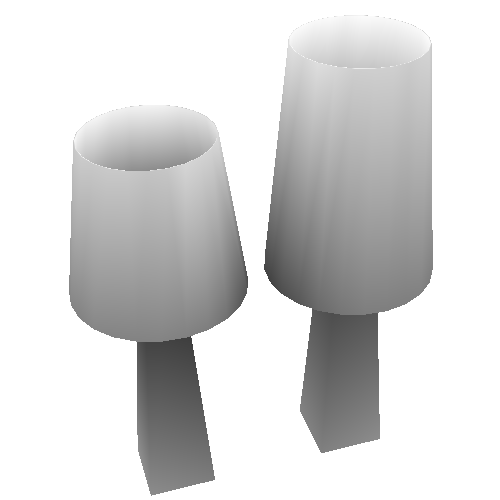}&
        \includegraphics[width=0.23\linewidth]{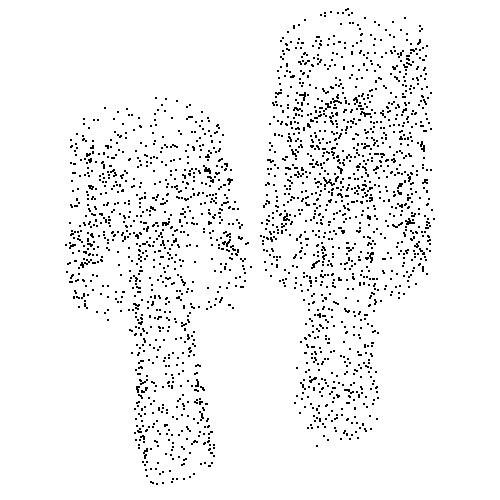}&
        \includegraphics[width=0.23\linewidth]{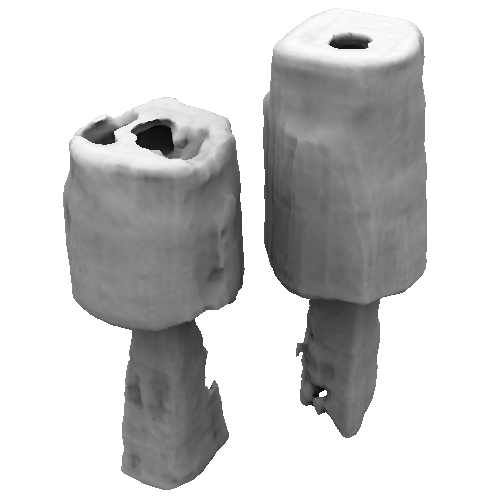}&
        \includegraphics[width=0.23\linewidth]{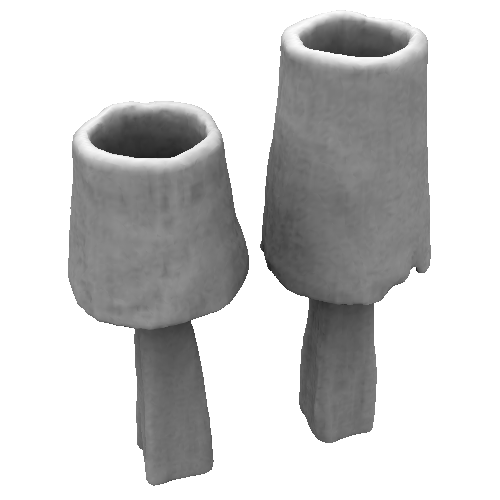}
        \\
        \includegraphics[width=0.23\linewidth, angle=90]{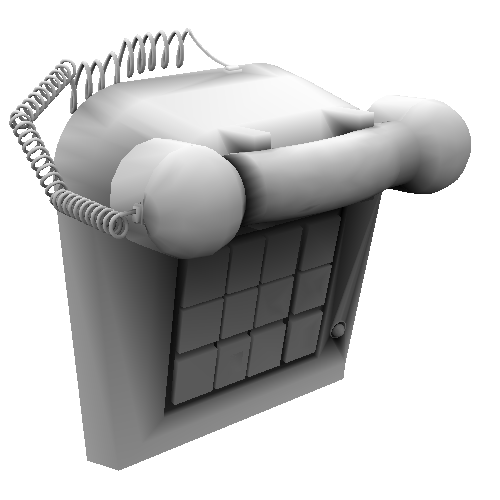}&
        \includegraphics[width=0.23\linewidth, angle=90]{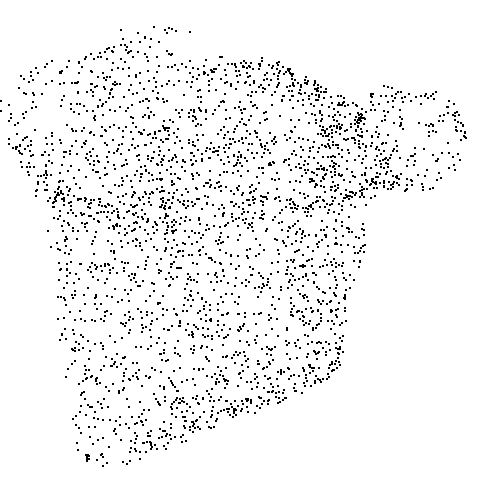}&
        \includegraphics[width=0.23\linewidth, angle=90]{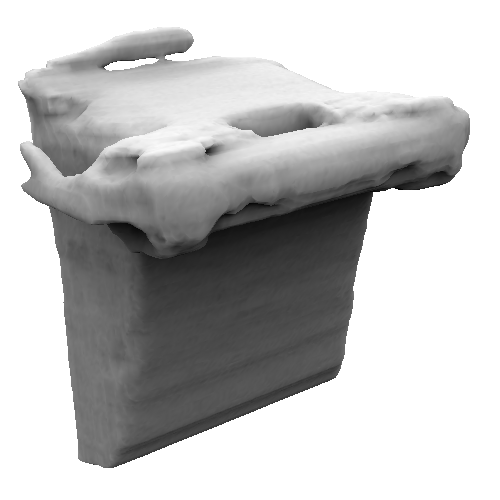}&
        \includegraphics[width=0.23\linewidth, angle=90]{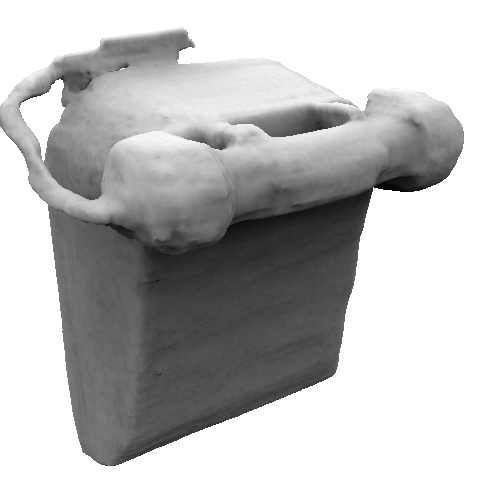}
        \\
        \includegraphics[width=0.23\linewidth]{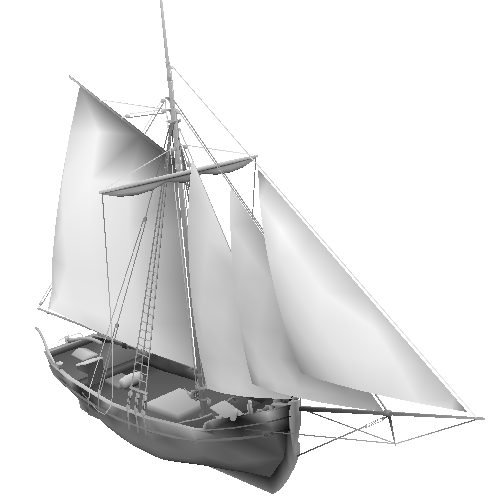}&
        \includegraphics[width=0.23\linewidth]{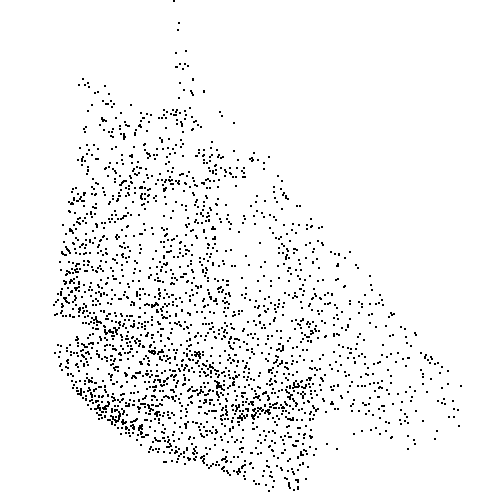}&
        \includegraphics[width=0.23\linewidth]{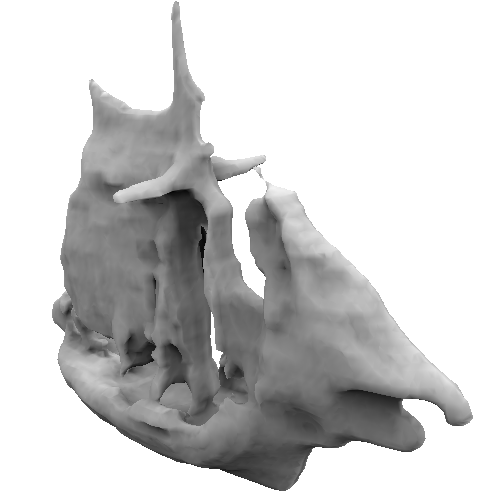}&
        \includegraphics[width=0.23\linewidth]{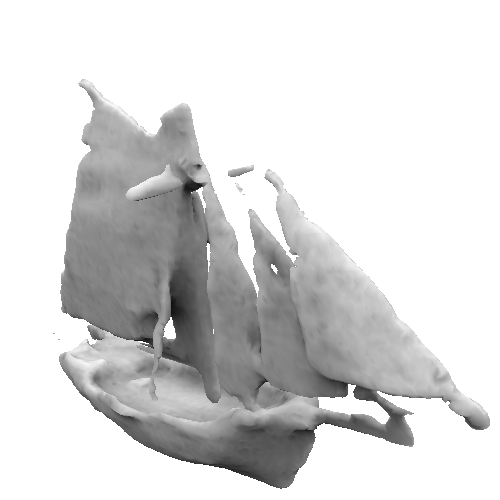}
    \end{tabular}
    \vspace{-5pt}
    \caption{\textbf{ShapeNet.} The methods train and test on 3k noisy pts.}
    \label{fig:shapenet}
    \vspace{-2mm}
\end{figure}

\textbf{Adapting to large size.}
As our method is convolutional, it naturally adapts to input point clouds $\pointcloud$ of arbitrary size. 
Yet, while $\pointcloud$ may contain millions of points, GPU memory limits in practice the number of points $\Ntest$ that can be treated together by the backbone. (We use $\Ntest \,{=}\,$100k.) 

As with semantic segmentation \cite{Boulch2020ACCV}, we can use a sliding-window with overlapping chunks of $\pointcloud$ of maximum size $\Ntest$.
Alternatively, as above, we can make subsamples of $\pointcloud$ by iteratively picking a low-priority point $\point \,{\in}\, \pointcloud$ and its $\Ntest{-}1$ nearest neighbors. (In our experiments, $\Nview \,{=}\, 3$.)

\textbf{Scene scaling.}
At inference time, the scale of the input point cloud may differ from the scales in the training set. As point-based backbones can be sensitive to variations of scale and density, we rescale the input such that the average distance between a point and its nearest neighbor is the same both in the training set and in the test point cloud.

\section{Experiments}

We experiment both on objects and scenes, in different point density regimes, with or without normal information depending on the baseline methods we compare with.

Because existing methods often perform well in some setting but not in others, most published papers tend to evaluate on different datasets or in specific configurations: number of train/test points, added noise, normals, generalization, etc. Some methods are also too slow to be evaluated on full datasets and report results only on dataset fractions. To be fair with these methods, we evaluate in their setting (when enough information is provided to do so) rather than impose them specific settings. It also illustrates the ability of our method to adapt to various configurations.

\subsection{Datasets, baselines and metrics}

\textbf{ShapeNet} \cite{Chang2015ARXIV}, as pre-processed by \cite{Choy2016ECCV}, contains watertight meshes of shapes in 13 classes, with train/val splits and 8500 objects for testing. As \cite{Peng2020ECCV}, we sample 3000 points from each mesh (at each epoch) and apply a Gaussian noise with zero mean and standard deviation 0.05.

\textbf{Synthetic Rooms} \cite{Peng2020ECCV} has 5000 synthetic scenes with random walls and populated with ShapeNet objects. We use \cite{Peng2020ECCV}'s protocol for sampling 10k pts on the meshes to create train/val/test data, with noise as for ShapeNet. Shapes are scenes in terms of complexity, objects in terms of size.

\textbf{ABC}~\cite{Koch_2019_CVPR} is a set of CAD models, mainly mechanical parts. We use splits and point preprocessing from~\cite{Erler2020Points2Surf}:\,4950 shapes for training, 100 for validation and 100 for testing.

\textbf{Famous}~\cite{Erler2020Points2Surf} contains 22 shapes of various origins, e.g., from the Stanford 3D Scanning Repository~\cite{krishnamurthy1996fitting}.

\textbf{Thingi10k}~\cite{zhou2016thingi10k}, as prepared by \cite{Erler2020Points2Surf}, has 100 shapes.

\textbf{SceneNet}  \cite{Handa2016Understanding,handa2016scenenet}
is a synthetic dataset of indoor scenes. Data prepared in the same way as \cite{huang2018robust} yield 34 scenes.

\textbf{MatterPort3D}~\cite{chang2017matterport3d} has indoor scenes too. We use the same 2 scenes as prepared and used by~\cite{tang2021sign}: with 65k pts.

\textbf{Baselines} are drawn among the state-of-the-art methods presented in Section~\ref{sec:convmethods}. We also compare to SPR \cite{Kazhdan2013SIGGRAPH}, a popular, non-learning-based reconstruction method that requires oriented normals (which is a strong hypothesis) and, possibly, a trimming parameter tuning (factor 6 in Tab.\,\ref{tab:quantit}).

\textbf{Our method}, unless otherwise stated, uses the FKAConv backbone \cite{Boulch2020ACCV}, feature size $n\,{=}\,32$ as in ConvONet \cite{Peng2020ECCV} or LIG \cite{Jiang2020CVPR}, $k\,{=}\,64$ neighbors, $h\,{=}\,64$ interpolation heads, and does not use normals nor TTA.

\textbf{Mesh Generation}, for implicit functions, is done with the Marching cubes \cite{Lorensen1987CG} with resolution $256^3$ for objects, 1\,cm for SceneNet, 2\,cm for MatterPort3D.

\textbf{Metrics.} We use the following common metrics: volumetric \textbf{IoU}, symmetric Chamfer L1-distance $\times 10^{2}$ (\textbf{CD}), normal consistency (\textbf{NC}), i.e., mean absolute cosine of normals in one mesh and normals at nearest neighbors in the other mesh, and F-Score \cite{Tatarchenko2019CVPR} with threshold value 1\% (\textbf{FS}). Surface metrics are approximated by point sampling.

\begin{table}[t]
\centering
\def\logits{relative}
\def\features{global}
\setlength{\tabcolsep}{4pt}
\begin{tabular}{l@{~}c@{~~}c@{~~}|ccc}
\toprule
    \em (a) Point backbone            &&&  IoU\,$\uparrow$  & CD\,$\downarrow$ & NC\,$\uparrow$ \\
\midrule
    Residual PointNet           &&& 0.661 & 10.583 & 0.817	\\
	\bf FKAConv                     &&& \tabfirst  0.882 & \tabfirst \hphantom{0}4.069 & \tabfirst  0.929		\\
\toprule
    \rlap{\em (b) No. interpolation neighbors}            &&&  IoU\,$\uparrow$  & CD\,$\downarrow$ & NC\,$\uparrow$\\
\midrule
    $k = \hphantom{00}1$   &&& 0.799 & 6.951 & 0.867 \\
    $k = \hphantom{00}8$   &&& 0.819 & 6.723 & 0.912 \\
    $k = \hphantom{0}\bf 64$ &&& \tabfirst 0.882 & 4.069	& 0.929	\\
    $k = 128$  &&& 0.876 & \tabfirst 3.611	& \tabfirst  0.930\\
\toprule
    \em (c) Interpol.\ features & \em{glob.} & \em{rel.} 
    &  IoU\,$\uparrow$  & CD\,$\downarrow$ & NC\,$\uparrow$ \\
    \midrule
        Max	                        & & \checkmark & 0.882 & 4.069	& 0.929	\\
        Mean	                    & & \checkmark & \tabsecond 0.883 & 3.703	& 0.933	\\
        Mean                        & \checkmark & & 0.854 & 5.331	 & 0.902	\\
        Inverse distance	        & & \checkmark & 0.877 & 3.947	& \tabsecond 0.935	\\
        Inverse distance            & \checkmark & & 0.851 & 4.724	& 0.912	\\
        Single-head attention	    & & \checkmark & 0.879 & \tabfirst 3.686	& 0.934	\\ 
        \bf Multi-head attention 	& & \checkmark & \tabfirst 0.895 & \tabsecond 3.702	& \tabfirst 0.938	\\
\bottomrule
\end{tabular}
\vspace{-2mm}
\caption{\textbf{Alternative study.} We train on ShapeNet chairs, without normals, 3k input points, with noise, and unless otherwise stated, FKAConv backbone, $k\,{=}\, 64$ neighbors, and max interpolation. We test on 10 models from each of the 13 ShapeNet classes. We interpolate either global features $\latentv_{\point}$ or relative features $\latentv_{\point,\qpoint}$.}
\label{tab:ablations}
\vspace{-2mm}
\end{table}

\subsection{Alternative and ablation studies}

To justify our algorithmic choices, we experiment on ShapeNet in generalization mode, training on chairs but evaluating on all the classes. We use the same train/test split as \cite{Mescheder2019CVPR, Peng2020ECCV}, evaluating on 130 shapes (10 per class).

As can be seen in Table~\ref{tab:ablations}(a), the convolutional backbone FKAConv \cite{Boulch2020ACCV} is more efficient by a large margin than the PointNet-based segmentation network with residual connections \cite{Qi2017CVPR, Mescheder2019CVPR}, which loses small scale information \cite{Qi2017NIPS}. 

Though interpolating from $k\,{=}\,64$ neighbors rather than $k\,{=}\,128$ has a slightly worse CD and NC (cf.\ Tab.\,\ref{tab:ablations}(b)), it has a better IoU and it is faster; we use this setting in the following. We note we get better results with a multi-head attention (using $h\,{=}\,64$ rather than $h\,{=}\,1$) and when interpolating relative rather than global features (cf.\ Tab.\,\ref{tab:ablations}(c)).

Last, Tab.\,\ref{tab:points2surf} and Fig.\,\ref{fig:real_world} show the benefits of the TTA strategy with models trained with 3k and 10k points on ABC.

\begin{table*}[t]
    \centering
    \setlength{\tabcolsep}{1pt}
    \vspace*{-0.5mm}
    \scalebox{0.97}{
    \begin{tabular}{@{}lr@{~~}|@{~}ccc@{~}|@{~}ccccc@{~}|@{~}ccccc@{}}
     & \llap{Test set} & \multicolumn{3}{c@{~}|@{~}}{ABC (100 shapes)} & \multicolumn{5}{c@{~}|@{~}}{Famous (22 shapes)} & \multicolumn{5}{c}{Thingi10k (100 shapes)}\\
    Method & \llap{Noise setting} & no-n. & var-n. & max-n. & no-n. & med-n. & max-n. & sparse & dense & no-n. & med-n. & max-n. & sparse & dense \\
    \toprule
    DeepSDF & \small\cite{Park2019CVPR}            & 8.41  & 12.51 & 11.34 & 10.08 & 9.89  & 13.17 & 10.41 & 9.49 & 9.16 & 8.83 & 12.28 & 9.56 & 8.35\\
    AtlasNet & \small\cite{Groueix2018CVPR}    & 4.69  & \hphantom{0}4.04  & 4.47  & 4.69  & 4.54  & 4.14  & 4.91  & 4.35 & 5.29 & 5.19 &  4.90 & 5.64 & 5.02\\
    SPR & \small\cite{Kazhdan2013SIGGRAPH}  & 2.49  & \hphantom{0}3.29  & 3.89  & 1.67  & 1.80  & 3.41  & 2.17  & 1.60 & 1.78 & 1.81 & 3.23 & 2.35 & 1.57\\
    Points2Surf & \small\cite{Erler2020Points2Surf} & 1.80  & \hphantom{0}2.14  & 2.76  & 1.41 & \tabsecond 1.51  & \tabfirst  2.52  & 1.93  & \tabfirst 1.33 &  1.41 & 1.47 & 2.62 & 2.11 & \tabfirst 1.35\\
    \midrule
    \multicolumn{2}{l@{~}|@{~}}{\OURS\ \ $\Ntrain{=}\Ntest{=}$3k} 
    & 1.87  & \hphantom{0}2.26  & 2.90  & 1.56  & 1.75  & 2.99  & 1.99  & 1.70 & 1.47 & 1.64 & 3.21 & 2.00 & 1.55\\
    \multicolumn{2}{l@{~}|@{~}}{\OURS\ \ $\Ntrain{=}\Ntest{=}$3k, $\hphantom{0}\Nview{=}$10~~}    
    & 1.77  & \tabsecond \hphantom{0}2.10  & \tabsecond 2.68  & \tabsecond 1.40  & 1.54  & 2.93  & \tabfirst 1.78  & \tabsecond 1.50 & \tabsecond 1.39 & \tabsecond 1.46 & \tabsecond 2.55 & \tabfirst 1.83 & 1.40 \\
    \multicolumn{2}{l@{~}|@{~}}{\OURS\ \ $\Ntrain{=}\Ntest{=}$10k} 
    & \tabsecond 1.72  & \hphantom{0}2.15  & 2.72  & 1.57  & 1.61  & 3.04  & 1.92  & 1.57 & 1.50 & 1.57 & 2.82 & 2.08 & 1.51\\
    \multicolumn{2}{l@{~}|@{~}}{\OURS\ \ $\Ntrain{=}\Ntest{=}$10k, $\Nview{=}$10} 
    & \tabfirst 1.70  & \tabfirst \hphantom{0}2.01  & \tabfirst 2.50  & \tabfirst 1.34 & \tabfirst 1.50  & \tabsecond 2.75 & \tabsecond 1.89  & \tabsecond 1.50 & \tabfirst 1.35 & \tabfirst 1.44 & \tabfirst 2.34 & \tabsecond 1.95 & \tabsecond 1.38\\
    \bottomrule
    \end{tabular}}
    \vspace{-6pt}
    \caption{\textbf{ABC, Famous, Thingi10k.} Training on ABC shapes with 10 scans, variable Gaussian noise ($\sigma$ uniformly picked in $[0,0.05L]$, $L$ largest box length). Chamfer distance $\times\,100$ on ABC, Famous and Thingi10k test sets, as prepared by \cite{Erler2020Points2Surf}: `no-n.' (no noise), `var-n.' (variable noise, as training), `max-n.' ($\sigma \,{=}\, 0.05L$), `med-n.' ($\sigma \,{=}\, 0.01L$), `sparse' (5 scans), 'dense' (30 scans). Only SPR uses normals.}
    \label{tab:points2surf}
    \vspace{-3mm}
\end{table*}

\begin{table}[t]
\centering
\begin{tabular}{l@{~}r|cccc}
\toprule
Method                             & & IoU\,$\uparrow$ & CD$\downarrow$ &NC\,$\uparrow$ & FS$\uparrow$ \\
\midrule
ONet      & \small\cite{Mescheder2019CVPR}  & 0.761 & 0.87 & 0.891 & 0.785 \\
ConvONet  &\small\cite{Peng2020ECCV}           & 0.884 & 0.44 & 0.938 & 0.942 \\
DP-ConvONet & \small\cite{Lionar_2021_WACV}     & \tabsecond 0.895 & \tabsecond 0.42 & \tabsecond 0.941 & \tabsecond 0.952 \\
\midrule
\OURS & \llap{(ours)}        
& \tabfirst 0.926 & \tabfirst 0.30 & \tabfirst 0.950 & \tabfirst 0.984\\
\bottomrule
\end{tabular}
\vspace{-6pt}
\caption{\textbf{ShapeNet.} The methods train and test on 3k noisy pts.}\label{tab:shapenet}
\medskip\smallskip
\begin{tabular}{l@{~}r|cccc}
\toprule
Method                             & & IoU\,$\uparrow$   & CD\,$\downarrow$     & NC\,$\uparrow$ & FS\,$\uparrow$ \\
\midrule
ONet      & \small\cite{Mescheder2019CVPR}  & 0.475 & 2.03 & 0.783 & 0.541 \\
SPR    & \small\cite{Kazhdan2013SIGGRAPH}  & -     & 2.23 & 0.866 & 0.810 \\ 
SPR trimmed&\small\cite{Kazhdan2013SIGGRAPH} & -     & 0.69 & 0.890 & 0.892 \\ 
ConvONet   &\small\cite{Peng2020ECCV}          & \tabsecond 0.849 & \tabsecond 0.42 & \tabsecond 0.915 & \tabsecond 0.964 \\
DP-ConvONet &\small\cite{Lionar_2021_WACV}      & 0.800 & \tabsecond 0.42 & 0.912 & 0.960 \\
\midrule
\OURS & \llap{(ours)}                        & \tabfirst 0.884 & \tabfirst 0.36 & \tabfirst 0.919 & \tabfirst 0.980\\
\bottomrule
\end{tabular}
\vspace{-6pt}
\caption{\textbf{Synthetic Rooms.} Learning-based methods train and test on 10k noisy pts. Only SPR uses normals. Numbers from \cite{Peng2020ECCV,Lionar_2021_WACV}.}\label{tab:syntheticrooms}
\label{tab:quantit}
\vspace{-3mm}
\end{table}

\begin{table}[t]
    \begin{tabular}{c@{~}l@{~}r|ccc}
        \toprule
        \hspace{-1mm}pts/m$^2$\hspace{1mm} & Method && CD\,$\downarrow$  & NC\,$\uparrow$  & FS\,$\uparrow$ \\
        \midrule
        20  & SPR     & \small\cite{Kazhdan2013SIGGRAPH} & 5.27 & 0.772 & 0.4392 \\
            & Neural Splines& \small\cite{Williams2021NeuralSplines} & 3.76 & 0.815 & 0.6563 \\
            & LIG       & \small\cite{Jiang2020CVPR} & \tabsecond 1.52 & \tabsecond 0.923 & \tabsecond 0.8757 \\
            & \OURS & \llap{(ours)} & \tabfirst 0.84 & \tabfirst 0.960 & \tabfirst 0.9600 \\
        \midrule
        100 & SPR     & \small\cite{Kazhdan2013SIGGRAPH} & 1.96 & 0.853 & 0.7709\\
            & Neural Splines& \small\cite{Williams2021NeuralSplines}& 1.15 & 0.931 & 0.9228\\
            & LIG       & \small\cite{Jiang2020CVPR} & \tabsecond 0.97 & \tabsecond 0.961 & \tabsecond 0.9643\\
            & \OURS & \llap{(ours)} & \tabfirst 0.57 & \tabfirst 0.984 & \tabfirst 0.9941\\
        \midrule
        500 & SPR     & \small\cite{Kazhdan2013SIGGRAPH} & 0.86 & 0.936 & 0.9787\\
            & Neural Splines& \small\cite{Williams2021NeuralSplines}& \tabsecond 0.60 & \tabsecond 0.982& \tabsecond 0.9958\\
            & LIG       & \small\cite{Jiang2020CVPR} & 0.87 & 0.975 & 0.9773\\
            & \OURS & \llap{(ours)} & \tabfirst 0.53 & \tabfirst 0.992 & \tabfirst 0.9987\\
        \midrule
        1000& SPR     & \small\cite{Kazhdan2013SIGGRAPH} & 0.73 & 0.967 & 0.9957\\
            & LIG       & \small\cite{Jiang2020CVPR} & \tabsecond 0.84 & \tabsecond 0.978 & \tabsecond 0.9750\\
            & \OURS & \llap{(ours)} & \tabfirst 0.53 & \tabfirst 0.993 & \tabfirst 0.9987\\
        \midrule
        \rowcolor{gray!25}
        &  Oracle  & \llap{(4M pts)} & 0.50 & 0.995 & 0.9998\\
        \bottomrule
    \end{tabular}
    \vspace{-6pt}
    \caption{\textbf{SceneNet.} LIG and {\OURS} train on ShapeNet with 10k pts with normals (no noise). Test is on SceneNet with normals (no noise). Neural Splines uses a grid size of 1024, 10k Nystr\"om samples, $8{\times}8{\times}8$ chunks. Numbers differ from \cite{Jiang2020CVPR} as we had to regenerate the unavailable watertight meshes: we used \cite{huang2018robust} with resolution 500k, higher than in \cite{Jiang2020CVPR}, 
    getting finer and thinner details where CAD models have no volume;
    as \cite{Jiang2020CVPR}, we ignore scenes with volume-to-area ratio $>$ 0.13, getting 34 scenes.
    `Oracle' is the ground truth evaluated against itself (two different samplings).}
    \label{tab:scene:scenenet}
    \vspace{-4mm}
\end{table}

\subsection{Reconstruction}

\textbf{Reconstruction without normals.}
Because of long running times, only a few published methods evaluate on the whole ShapeNet dataset. We outperform them on all metrics with a significant margin (Table~\ref{tab:shapenet}). We reconstruct finer details (Figure~\ref{fig:shapenet}) and we do not have the same tendency as ConvONet to fill volumes; we can instead generate more easily thin surfaces, which explain our superior IoU. We outperform other methods as well on Synthetic Rooms (Table~\ref{tab:syntheticrooms}), where also we capture much finer details.

\textbf{Generalization.}
LIG is specifically designed for scalability and generality. It learns to reconstruct small shape patches from a given dataset, and then applies it to any new object or scene. Points2Surf is a patch-learning method too, although its requirement for a global view of the input and its running time make it less suited for scene reconstruction.

We compare to LIG, training both methods on ShapeNet objects (with normals as LIG requires them) and testing on SceneNet. We generalize better (Tab.\,\ref{tab:scene:scenenet}) at all densities, capturing finer details and not erasing thin objects (Fig.\,\ref{fig:scenenet}).

We compare to Points2Surf, training on ABC in the same setting. We outperform Points2Surf on most of their settings (Tab.\,\ref{tab:points2surf}), both on ABC and when generalizing to Famous and Thingi10k. Points2Surf outperforms {\OURS} only on very noisy or dense inputs, and only with a small margin.%

\textbf{Scene reconstruction without normals.}
We compare to SA-ConvONet on MatterPort3D scenes (Fig.\,\ref{fig:matterport}) in their same actual setting (downsampling to 65536 pts). Our reconstruction is less smooth than SA-ConvONet but has finer details. As SA-ConvONet overfits many networks at inference time on top of ConvONet, it is notably slower too.

\subsection{Discussion and limitations}

Our approach is suited both for single-object and whole-scene reconstruction. However, although it can cope with a substantial variation of point density, it cannot complete shapes when large parts are missing. Apart from a few methods like \cite{dai2017complete, dai2018scancomplete, dai2020sgnn, Siddiqui2021RetrievalFuse}, only object-targeted methods can presently do it, for classes known at training time, but they cannot reconstruct scenes at all.

Inferring surface orientation, when normals are not provided, requires wide context information. But a high density may reduce the receptive field, yielding orientation failures and artifacts. Our TTA only partly addresses the issue; handling it directly at backbone level would be better.

Nevertheless, \OURS\ reaches the state of the art for both object and scene reconstruction, with or without oriented normals. It shows good generalization capabilities to shapes and scenes that are very different from the training set.

More details and visuals on the method and on the experiments are in the supplementary material.

\smallskip
\noindent\textbf{Acknowledgments } to Gilles Puy for fruitful discussions.

{\small
\bibliographystyle{ieee_fullname}
\bibliography{biblio}
}

\clearpage
\appendixwithtoc 
\section*{Supplementary material}
\makeatletter
\renewcommand\paragraph{%
  \@startsection{paragraph}{4}{\z@}%
     {1.1ex \@plus1.0ex \@minus.5ex}%
     {-0.5em}%
     {\normalfont\normalsize\bfseries}}
\makeatother



\section{Implementation details}

\paragraph{Code avalability.}

The code of our method is available at \url{https://github.com/valeoai/POCO}.

\paragraph{Framework and hardware.}

Our code uses PyTorch as deep learning framework.
All experiments were done with a single NVIDIA RTX 2080 Ti GPU with 11GB memory.

\paragraph{Backbone.}

We used FKAConv \cite{Boulch2020ACCV} as convolutional backbone, with default parameters (number of layers, number of layer channels). Only the latent vector size~$n$, i.e., the output dimension of the backbone, was changed. It was set to 32, which is also the output dimension of all linear layers of the occupancy decoder (except the last one, which outputs the occupancy).
As a comparison, methods like {\ConvONet}~\cite{Peng2020ECCV} and LIG \cite{Jiang2020CVPR} also use latent vectors of size~32.

\paragraph{Architecture.}
The network architecture is described in Figure~5 of the main paper. We phrase here some parts of it.

The input size of the relative encoder (green area in Figure~5) is the size of the latent vectors (i.e., the backbone output size) plus the size of point coordinates, i.e., $32\,{+}\,3\,{=}\,35$. All linear layers have an output size of 32, except the multi-head layer for the computation of significance weights, of output size~$h\,{=}\,64$, and the final occupancy layer, of output size~2, corresponding to classes empty and full.
The layer activations all are ReLUs.
Batch norms are only used in the backbone, i.e., the absolute encoder $\encoder$; there are none in the relative encoder $\relativizer$, nor in the decoder $\decoder$.

\paragraph{Point sampling at training time} 
is not part of {\OURS}. We reused existing dataset samplings (from ConvONet~\cite{Peng2020ECCV} and Points2Surf~\cite{Erler2020Points2Surf}) to compare on the same training data. The other datasets are only used for inference.

\paragraph{Training settings.}

We train using Adam~\cite{Kingma2015ICML} with learning rate~$10^{-3}$.
The training batch size is 16 for 3k input points and 8 for 10k input points.
We train for 600k iterations.

\section{Meshing for occupancy}\label{sec:meshoccup}

Mesh generation, for implicit functions, generally relies on the Marching cubes (MC) algorithm \cite{Lorensen1987CG}, evaluating occupancy on a regular 3D grid.

\paragraph{Marching cubes based on refinements (MC-refin).}
Recently, the MC variant used in ONet \cite{Mescheder2019CVPR} has often been used due to its higher speed. It operates on a coarse grid but locally refines the resolution thanks to a heuristics:
Unless all corners of a cube at a given resolution agree on being empty of full, i.e., as soon as two corners of a cube disagree on occupancy, the cube is subdivided into 8 subvoxels. The initial grid is typically of size $32^3$, and it is typically refined (subdivided) up to two times, leading to a local resolution equivalent to a $128^3$ grid. The resulting mesh, after MC, is furthermore simplified \cite{Garland1998Simplifying} and refined using first and second order gradient information \cite{Mescheder2019CVPR}. While the heuristics may miss thin details, this MC with refinement (MC-refin) leads to a much faster running time than plain MC, with a factor up to $8^2$ when using up to two refinement steps.

\paragraph{Marching cubes based on region growing (MC-regro).}
To ensure we have little chances of missing refinements, in particular for locally complex surfaces or thin volumes, we use a different strategy. We consider from the outset a fine-grained resolution but, to prevent many useless queries in large empty or full regions, we adopt a region-growing approach (MC-regro). The seeds are the input points, for which we compute the occupancy. We then compute the occupancy for query points that are both in the close neighborhood (voxels at distance at most 2 grid steps) of both a location in the empty volume and a location in the full volume, i.e., close to the surface. And we iterate.

Besides, with the Marching cubes algorithm, a vertex is placed on the edge of a cube by linearly interpolating the two scalar values at the edge's endpoints. But contrary to distance fields, occupancy fields may have sharp transitions. Consequently, opposite-side endpoints frequently have values close to 0 and 1, and vertices tend to be placed in the middle of segments, creating discretization effects. To prevent it, we perform a dichotomic search along edges to better locate the occupancy transition.
We operate 10 dichotomies, which is more than enough in most cases.

In general, reconstructions with MC-regro are qualitatively better than MC-refin on scenes, but similar on objects. In fact, quantitative results on ShapeNet show a similar reconstruction accuracy of {\OURS} with either MC-refin or MC-regro. The reason probably is that thin details have little impact on the different metrics. This ability to capture thin details makes MC-regro generally slower than MC-refin (see Section~\ref{sec:runtime}, Table~\ref{tab:timerecons}).

\section{Running times}\label{sec:runtime}
Some running times are given in Figures 1 and~4 in the paper, as well as here in Tables~\ref{tab:timings} and~\ref{tab:timerecons}. 

The time for the backbone to extract features is negligible (${<}\, 1$\%). The bottleneck is the decoding, as we have to respond to many occupancy queries depending on the resolution of the Marching cubes (MC). And to answer an MC query, the bottleneck is the computation of nearest neighbors, which currently is not optimized, requiring communications between the GPU and the CPU. (It could probably be optimized by pre-computing neighbors at low MC resolution to reduce the GPU-CPU communication overhead.)

In contrast, grid-based methods such as those based on ConvONet \cite{Peng2020ECCV, Lionar_2021_WACV, tang2021sign} do not need such an optimization as they do not depend on nearest neighbors. However, while our approach requires extracting one feature per point for encoding (typically a few thousands points for an object), these other methods extract one feature per grid cell, typically $64^3 \,{\approx}\, 262$k. Besides, as we show in the paper, losing input points induces a loss of details.

\begin{table}[t]
    \centering    
    \begin{tabular}{l|r}
    Method and setting & \multicolumn{1}{c}{Time} \\
    \midrule
    \relax{Points2Surf} &  \\
    \quad Full reconstruction (single thread) & 23 min 48 s \\
    \quad Full reconstruction (1 thread per model) & 10 min 15 s \\
    \midrule
    \relax{{\OURS} ($\Ntrain\,{=}\,\Ntest\,{=}\,3$k, $\Nview\,{=}\,10$)} &  \\
    \quad Only inference of latent vectors & 38 s \\
    \quad Full reconstruction (single thread) & 4 min 27 s
    \end{tabular}
    \vspace*{-2mm}
    \caption{\textbf{Running time} for reconstructing the 4 models of the Real-World dataset (50k pts each) using Points2Surf or {\OURS}.}
    \label{tab:timings}
\end{table}

\begin{table}[t]
\centering\setlength{\tabcolsep}{3pt}
\begin{tabular}{l@{}cc@{}r}
Method & MC-refin & MC-regro & Time \\
\midrule
    SA-ConvONet    & \checkmark & & 245.7 s\\
    LIG (5k iter.) & \checkmark & & 104.5 s\\
    LIG (3k iter.) & \checkmark & &  66.2 s\\
    Points2Surf    & \checkmark & &  38.4 s\\
    SPR            & \checkmark & &  14.9 s\\
    Neural Splines & \checkmark & &  12.7 s\\
    ConvONet       & \checkmark & &   0.6 s\\
\midrule
    {\OURS}        & & \checkmark &  10.7 s\\
    {\OURS}        & \checkmark & &   2.5 s\\
\end{tabular}
    \vspace*{-2mm}
\caption{\textbf{Average reconstruction time} of different methods for ShapeNet shapes from 3k~points using the same $128^3$ grid size for the Marching cubes (MC), although with different heuristics and MC variants. MC-refin is the commonly-used MC variant in \cite{Mescheder2019CVPR} that operates on a $32^3$ grid and potentially refines it locally twice into a local resolution equivalent to a $128^3$ grid. MC-regro is our region-growing variant of the Marching cubes that directly operates on a $128^3$ grid, although sparsely (see Section~\ref{sec:meshoccup}).}
\label{tab:timerecons}
\vspace*{-2mm}
\end{table}

\paragraph{Impact of test-time augmentations.} 
Although in this case, because of the high point-cloud density (50k pts), we apply the test-time augmentation (TTA) strategy and run the latent vector inference on many different point cloud subsamples (such that each point is seen at least $\Nview\,{=}\,10$ times), our method is still significantly faster than Points2Surf.

In fact, as our encoding time is negligible compared to the numerous decoding queries for meshing with MC, our TTA strategy at feature level brings little slowdown, e.g., +5\% for $\Nview\,{=}\,10$, compared to $\Nview\,{=}\,1$.

\paragraph{Overall reconstruction time.} 
In Table~\ref{tab:timerecons}, we report the average reconstruction time of different methods.
To be fair, given that mesh generation via occupancy queries is a running time bottleneck, we compare the methods using the same MC algorithm, namely MC-refin with a coarse grid of size $32^3$ that can be refined up to twice, i.e., into a grid of size $128^3$. We also report the running time of {\OURS} with our MC-regro variant on a grid of size $128^3$.
As said in Section~\ref{sec:meshoccup}, the quantitative results of {\OURS} with either MC-refin or MC-regro are similar.

On ShapeNet with medium-density points clouds (3k points per shape), we rank second behind ConvONet for speed.
Note however that LIG is faster on denser scenes (see Figure~4 of the main paper) as the computation time per patch is constant, while our kNN search based on a kd-tree gets slower. 
(It could be faster by pre\-com\-put\-ing neighbors in the data loader, to limit GPU-CPU exchanges.)

\section{Receptive field}

A question that naturally arises to understand the power and benefits of different approaches is the size of the receptive field for inferring occupancy features.

Because it is based on nearest neighbors, the receptive field of the backbone varies based on the scene geometry. It naturally tends to augment with the number of layers but sometimes, as when a separate group of points are mutual neighbors, the local receptive field does not increase.

\begin{figure}[ht]
    \centering
    \vspace{-2mm}
    \includegraphics[width=0.8\linewidth]{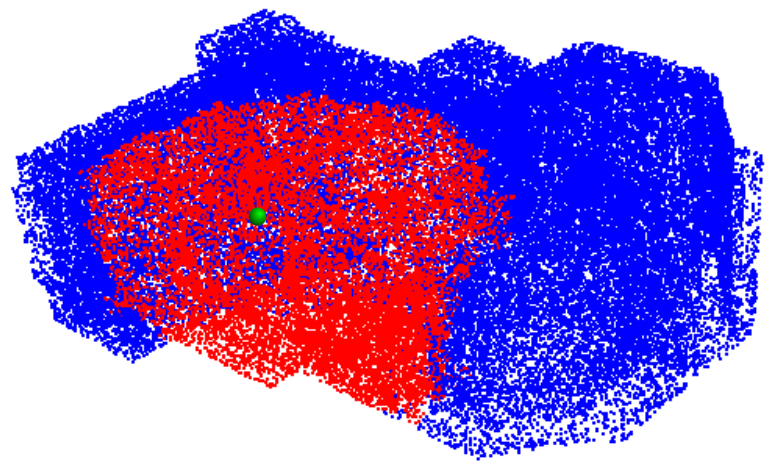}
    \vspace{-1mm}
    \caption{\textbf{Receptive field of the FKAConv backbone} on a point cloud from SceneNet with density 100 pts/m$^2$. The receptive field of the point marked in green is colored in red.}
    \label{fig:receptive}
    \vspace*{-2mm}
\end{figure}

To evaluate the actual (in fact, maximum) receptive field of a given point, we apply the following procedure:
\begin{enumerate}[topsep=1pt,itemsep=-2pt]
    \item We use a variant of the network without ReLUs and where the convolutions are replaced with averaging.
    \item We apply the loss on a single output location.
    \item We back-propagate the loss signal.
    \item We identify input points receiving a non-zero gradient.
\end{enumerate}
On a SceneNet living-room scene, with density 100 pts/m$^2$, we obtain an average receptive field of 29k points when looking at non-zero gradient (see Figure~\ref{fig:receptive}).
If we only look at points for which the back-propagated gradient has a norm greater than $10^{-7}$ (i.e., a significant gradient), then the receptive field encompasses 16k points.

\section{Experiments}

\subsection{Choice of compared methods and datasets}

Following the success of methods such as AtlasNet \cite{Groueix2018CVPR} and DeepSDF \cite{Park2019CVPR}, a dozen of new learning-based reconstruction methods have been published every year.

\begin{table}[t]
\centering
\setlength{\tabcolsep}{3pt}
\noindent\begin{tabular}{l@{~}l|l|l}
Train & Test & Train set & Test set \\
\midrule
object & object & ShapeNet & ShapeNet \\
\midrule
object & object & ABC & ABC, \emph{Thingi10k,} \\
& & & \quad \emph{RealWorld, Famous}\\
\midrule
object & scene & ShapeNet & \emph{SceneNet} \\
\midrule
scene & scene
& Synthetic & Synthetic Rooms, \\
& & \quad Rooms & \quad \emph{MatterPort3D} 
\end{tabular}
\vspace{-2mm}
\caption{\textbf{Datasets used for training and testing.} \emph{Italic:} datasets used in a generalization setting, including from objects to scenes.}
\label{tab:useddatasets}
\vspace{-4mm}
\end{table}

As said in the main paper, existing methods often perform well in some settings but not in others. Consequently, most published papers tend to evaluate on different datasets (see Table~\ref{tab:datasets}) or in specific configurations: low or high density of train/test points, with or without added noise and outliers, with or without oriented normals, training specifically for a class of shapes or generalizing to any shape, addressing single object or whole scene reconstruction, etc.  Some methods are also too slow to be evaluated on full datasets and report results only on dataset fractions.  Last, although most methods make code available, some do not offer pre-trained models, or scripts, or parameters (at the time of writing).
This makes comparisons particularly difficult. 

We chose to compare to some of the most cited or most recent methods. To be fair with these methods, we evaluate in their setting (when enough information is provided to do so) rather than impose them other specific settings. It also illustrates the ability of our method to adapt to various configurations. Method codes are referenced in Section~\ref{sec:assetsmethods}.

The datasets that we used in our experiments are listed in Table~\ref{tab:useddatasets}.  Datasets references are in Section~\ref{sec:assetsmethods}.
\begin{itemize}[nosep]
    \item On SceneNet, we chose points with normals, which allows comparing to LIG (which requires normals).
    \item On MatterPort3D, we chose points without normals, allowing comparison to SA-ConvONet but not to LIG.
    \item On ShapeNet, we chose in the main paper points without normals and with noise, allowing comparison to ConvONet; in this supplement, we use points with normals and without noise, allowing to compare to LIG.
\end{itemize}

\begin{table*}[p]
\centering\newcommand{\rot}[1]{\rotatebox{90}{#1}}
\let\CM\cmark
\newcommand{\ALL}{$\blacksquare$}
\newcommand{\HAL}{$\squarebotblack$}
\newcommand{\FEW}{$\square$}
\newcommand{\alo}{$\bullet$}
\newcommand{\feo}{$\circ$}
\newcommand{\ALO}{$\bullet^\bot$}
\newcommand{\FEO}{$\circ^\bot$}
\newcommand{\ALn}{$\blacksquare\hphantom{^\bot}$}
\newcommand{\HAn}{$\squarebotblack\hphantom{^\bot}$}
\newcommand{\FEn}{$\square\hphantom{^\bot}$}
\newcommand{\ALN}{$\blacksquare^\bot$}
\newcommand{\FEN}{$\square^\bot$}
\begin{tabular}{l@{~}r!{\vrule width 2pt}c|c|c!{\vrule width 2pt}c|c|c|c|c|c|c!{\vrule width 2pt}c|c|c|c|c|c!{\vrule width 2pt}}
&&&&&
\multicolumn{7}{c!{\vrule width 2pt}}{Objects} &
\multicolumn{6}{c!{\vrule width 2pt}}{Scenes}
\\
\cline{6-18} 
&&
\rot{normals required} &
\rot{code unavailable} &
\rot{pre-training unavaila\rlap{ble}} &
\rot{3D Warehouse} &
\rot{ABC} &
\rot{D-Faust} &
\rot{Famous} &
\rot{ShapeNet} &
\rot{Thingi10K} &
\rot{ThreeDScans} &
\rot{3DFront} &
\rot{MatterPort3D} &
\rot{ScanNet} &
\rot{SceneNet} &
\rot{Synthetic Rooms} &
\rot{Tanks and Temples~~}
\\
\toprule
AdaConv& \cite{Ummenhofer2021Adaptive}&\CM&  &\CM&      &      &      &      &      &      &      &      &      &      &      &      & \ALN \\
\midrule
ConvONet    & \cite{Peng2020ECCV}    &   &   &   &      &      &      &      & \ALn &      &      &      & \FEn & \ALn &      & \ALL &      \\
\midrule
DeepLS & \cite{Chabra2020DeepLS}     &   &\CM&\CM& \FEW &      &      &      &      &      &      &      &      &      &      &      &      \\
\midrule
DeepSDF       & \cite{Park2019CVPR}  &   &   &\CM&      &      &      &      & \ALn &      &      &      &      &      &      &      &      \\          
\midrule
DefTet       & \cite{gao2020deftet}  &   &\CM&\CM&      &      &      &      & \ALn &      &      &      &      &      &      &      &      \\
\midrule
DP-ConvONet& \cite{Lionar_2021_WACV}  &   &   &   &      &      &      &      & \ALn &      &      &      &      &      &      & \ALL &      \\
\midrule
IF-NET    & \cite{Chibane2020CVPR}   &   &   &   &      &      &      &      & \ALn &      &      &      &      &      &      &      &      \\
\midrule
IGR    & \cite{Gropp2020ICML}        &   &   &   &      &      &      &      &      &      &      &      &      &      &      &      &      \\
\midrule
IM-NET    & \cite{Chen2019Learning}  &   &   &   &      &      &      &      & \FEn &      &      &      &      &      &      &      &      \\
\midrule
LDIF          & \cite{Genova2020CVPR}&   &   &\CM&      &      &      &      & \ALn &      &      &      &      &      &      &      &      \\
\midrule
LIG         & \cite{Jiang2020CVPR}   &\CM&   &   &      &      &      &      & \ALN &      &      &      & \ALN &      & \ALN &      &      \\
\midrule
MetaSDF  & \cite{sitzmann2020metasdf}&   &   &\CM&      &      &      &      & \FEn &      &      &      &      &      &      &      &      \\
\midrule
NDF        & \cite{Chibane2020Neural}&   &   &\CM&      &      &      &      & \FEn &      &      &      &      &      &      &      &      \\
\midrule
Neural-Pull& \cite{Baorui2021NeuralPull}&&   &   &      & \FEW &      & \FEW &      &      &      &      &      &      &      &      &      \\
\midrule
Neural Splines& \cite{Williams2021NeuralSplines}&\CM&&& &      &      &      & \FEN &      &      &      &      &      &      &      &      \\
\midrule
ONet         & \cite{Mescheder2019CVPR}& &   &   &      &      &      &      & \ALN &      &      &      &      &      &      &      &      \\
\midrule
Points2Surf&\cite{Erler2020Points2Surf}& &   &   &      & \FEW &      & \FEW &      & \FEW &      &      &      &      &      &      &      \\
\midrule
RetrievalFuse& \cite{Siddiqui2021RetrievalFuse}&&&\CM&  &      &      &      & \ALn &      &      & \ALL*& \ALL*&      &      &      &      \\
\midrule
SA-ConvONet & \cite{tang2021sign}    &   &   &   &      &      &      &      & \FEn &      &      &      & \FEn & \FEn &      & \FEW &      \\
\midrule
SAIL-S3  & \cite{Zhao2021SignAgnostic}&  &\CM&\CM&      &      &      &      & \FEn &      & \FEW &      &      &      &      &      &      \\
\midrule
SAL  & \cite{Atzmon2020CVPR}         &   &   &   &      &      & \ALL &      &      &      &      &      &      &      &      &      &      \\
\midrule
SALD  & \cite{Atzmon2021ICLR}        &   &\CM&\CM&      &      & \ALL &      &      &      &      &      &      &      &      &      &      \\
\midrule
SAP  & \cite{Peng2021SAP}            &   &   &   &      &      &      &      & \ALn &      &      &      &      &      &      &      &      \\
\midrule
ScanComplete & \cite{dai2018scancomplete}&&  &   &      &      &      &      &      &      &      &      &      & \ALL*&      &      &      \\
\midrule
SG-NN  & \cite{dai2020sgnn}          &   &   &   &      &      &      &      &      &      &      &      & \ALL*&      &      &      &      \\
\midrule
SPR  & \cite{Kazhdan2013SIGGRAPH}    &\CM&   &   &      &      &      &      &      &      &      &      &      &      &      &      &      \\
\bottomrule\raisebox{10pt}{}%
\OURS       & (ours)                 &   &   &   &      & \FEW &      & \FEW & \ALn & \FEW &      &      & \FEn &      & \ALN & \ALL &      \\
\end{tabular}
\vspace{-2mm}
\caption{\textbf{Datasets used for the \emph{evaluation} of 3D reconstruction methods \emph{from point clouds} in their published paper}, if freely available and \emph{$>$\,10 shapes are used}, and availability of code or pre-trained models suited for testing on the datasets (at the time of writing).
\newline \hspace*{4mm} 
\ALL: test on all/many shapes of the dataset ($> 1000$),
\FEW: test on a few shapes ($\leq 100$ or a single category),
$^\bot$: test with ground-truth normals as input, *: actual scans rather than uniformly sampled points.
\newline \hspace*{4mm} 
Tests on a given dataset may however be done in different settings (number of sampled points, amount of added noise or outliers, use many shapes but excluded classes or objects, etc.). For instance, many different numbers can be found in various publications for the performance of ONet on the ShapeNet dataset.}
\label{tab:datasets}
\end{table*}

\begin{figure*}[t]
\vspace*{-7mm}
\setlength{\tabcolsep}{2.8pt}
\begin{tabular}{@{}ccc|ccc|ccc@{}}
    \multicolumn{3}{c|}{512 pts} & \multicolumn{3}{c|}{2048 pts} & \multicolumn{3}{c}{8192 pts} \\
    Input & LIG & \OURS & Input & LIG & \OURS & Input & LIG & \OURS \\
    \midrule
    \includegraphics[width=0.10\linewidth]{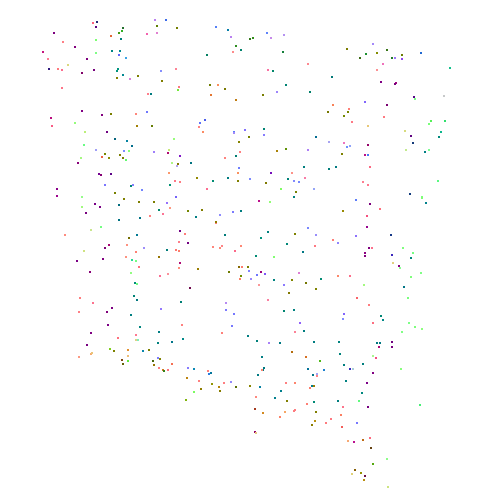}&
    \includegraphics[width=0.10\linewidth]{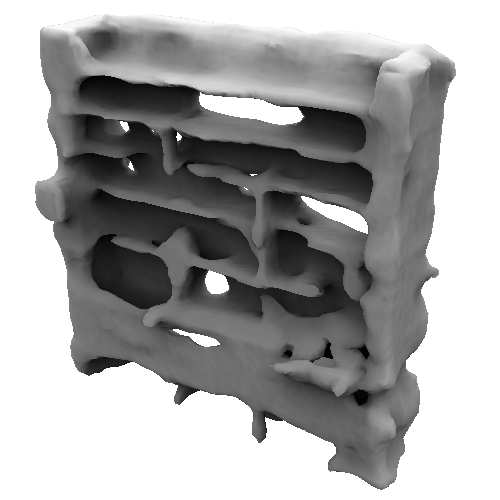}&
    \includegraphics[width=0.10\linewidth]{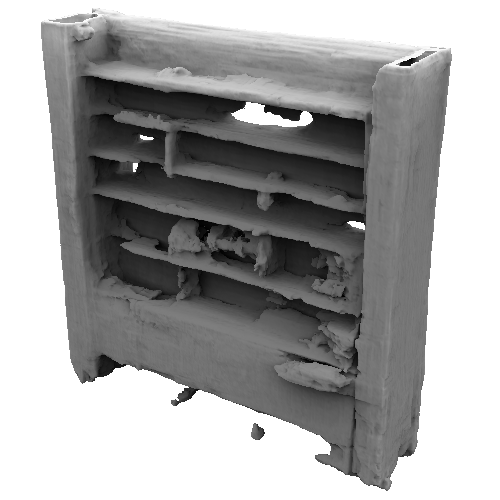}&
    \includegraphics[width=0.10\linewidth]{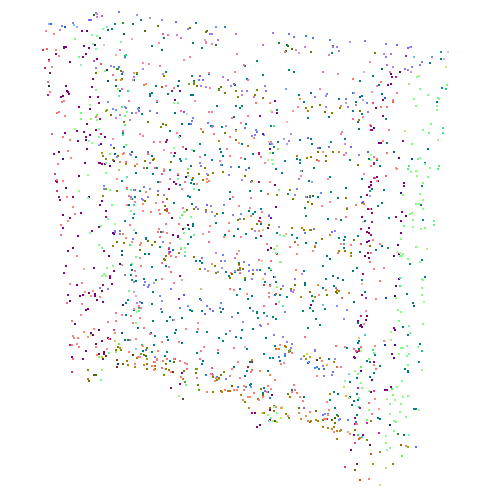}&
    \includegraphics[width=0.10\linewidth]{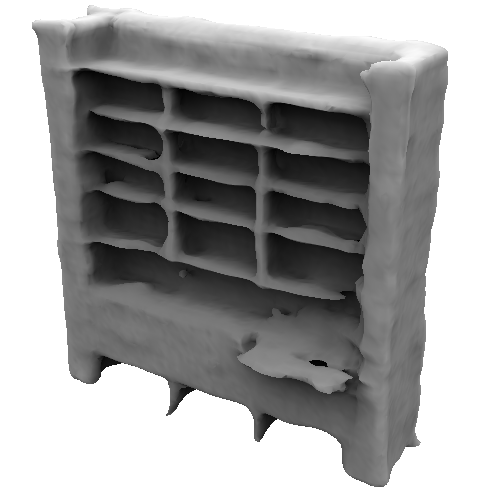}&
    \includegraphics[width=0.10\linewidth]{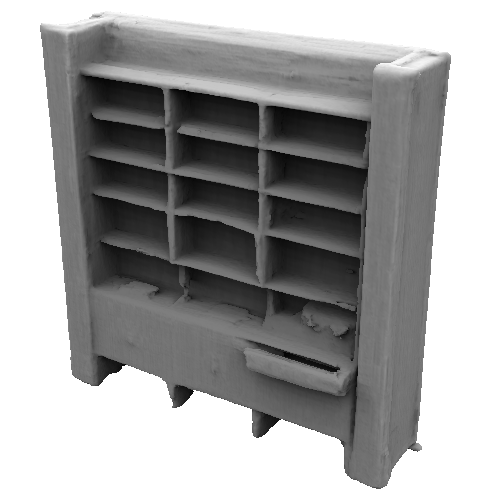}&
    \includegraphics[width=0.10\linewidth]{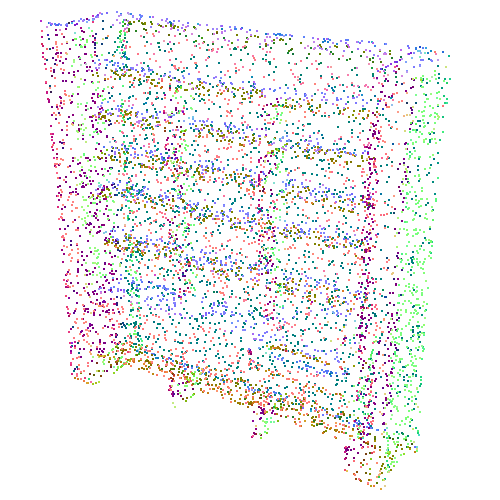}&
    \includegraphics[width=0.10\linewidth]{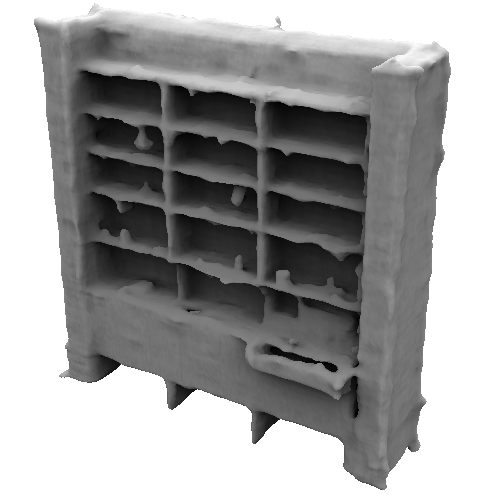}&
    \includegraphics[width=0.10\linewidth]{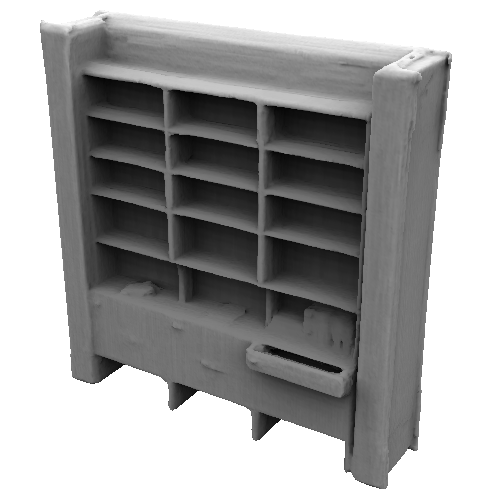}\\
    \includegraphics[width=0.10\linewidth]{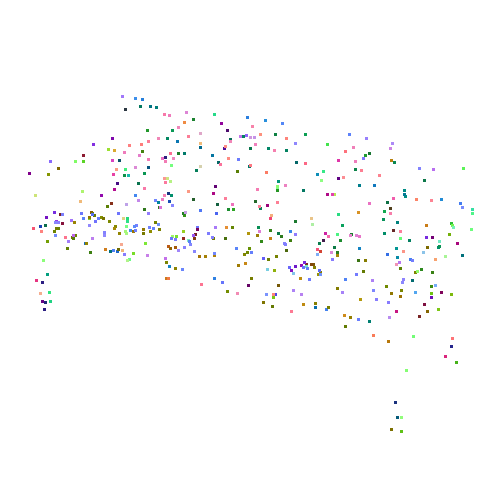}&
    \includegraphics[width=0.10\linewidth]{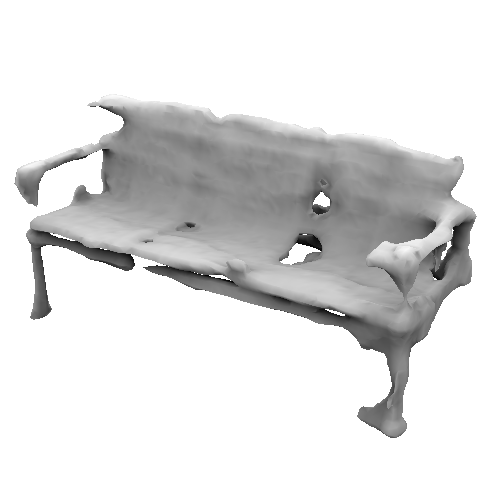}&
    \includegraphics[width=0.10\linewidth]{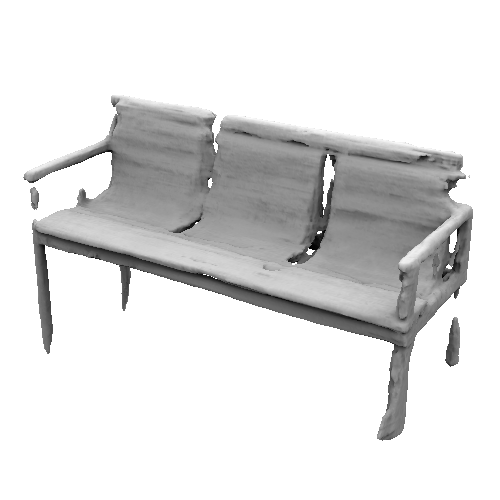}&
    \includegraphics[width=0.10\linewidth]{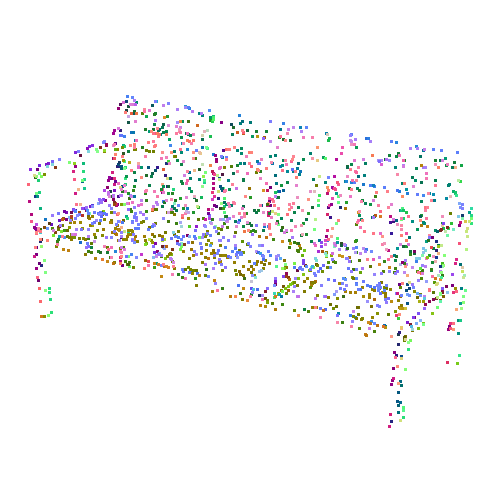}&
    \includegraphics[width=0.10\linewidth]{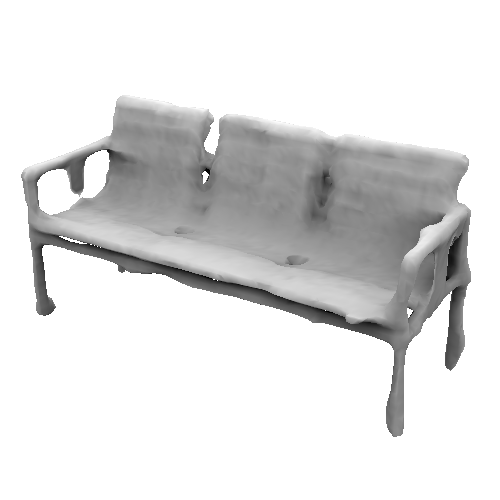}&
    \includegraphics[width=0.10\linewidth]{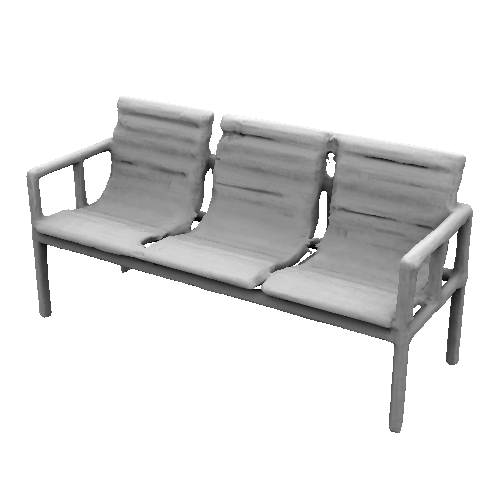}&
    \includegraphics[width=0.10\linewidth]{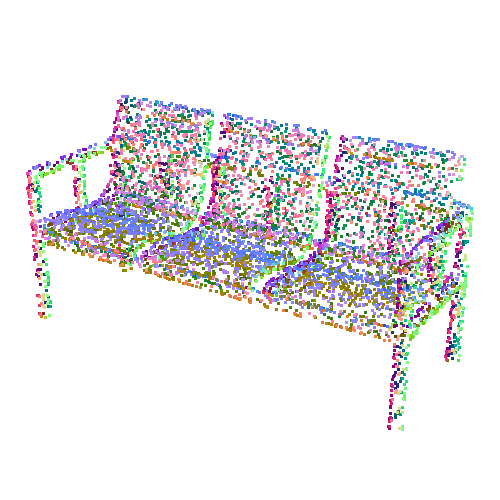}&
    \includegraphics[width=0.10\linewidth]{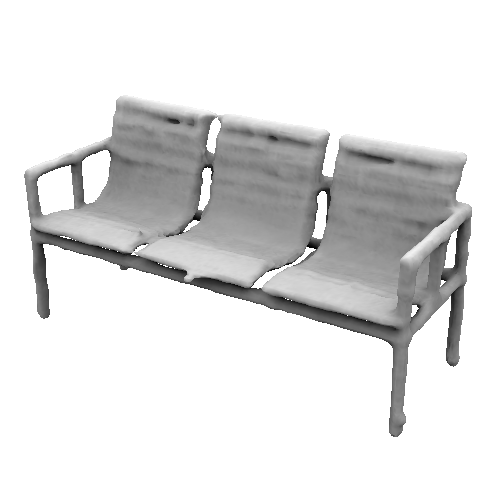}&
    \includegraphics[width=0.10\linewidth]{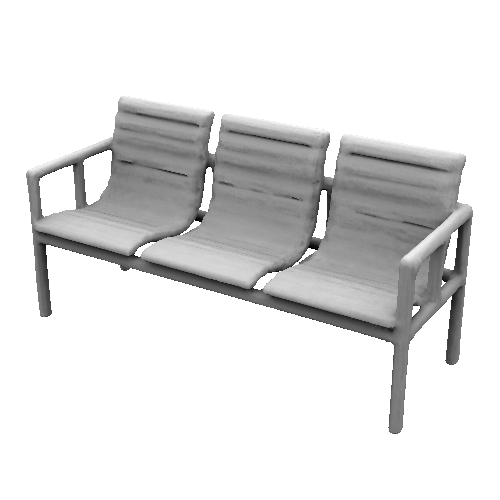}\\
    \includegraphics[width=0.10\linewidth]{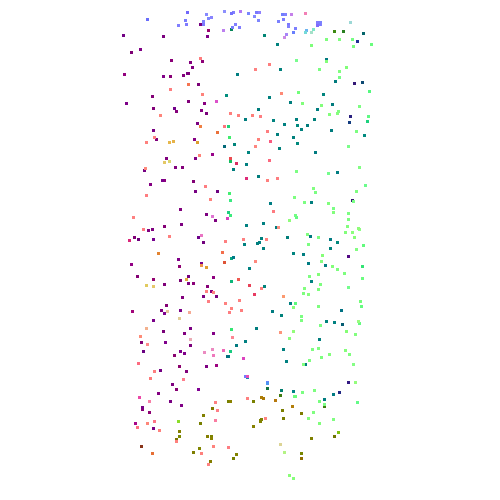}&
    \includegraphics[width=0.10\linewidth]{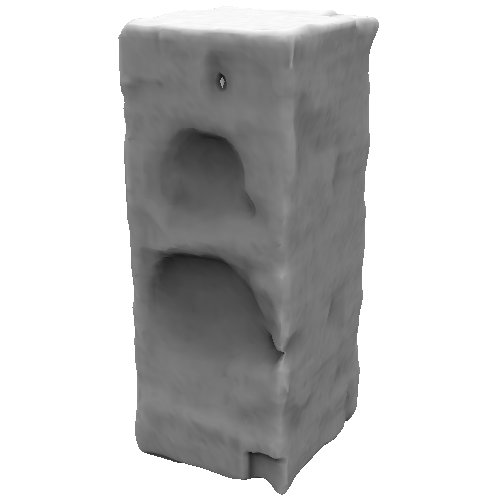}&
    \includegraphics[width=0.10\linewidth]{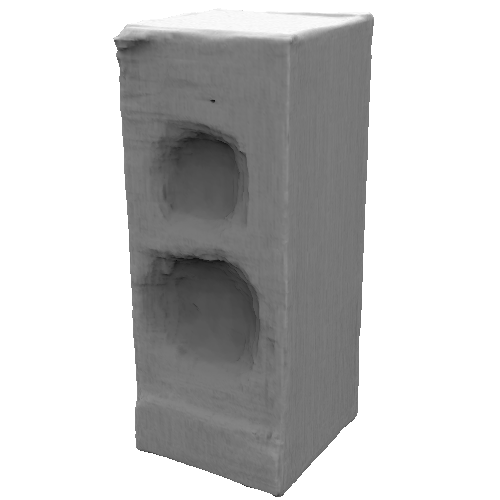}&
    \includegraphics[width=0.10\linewidth]{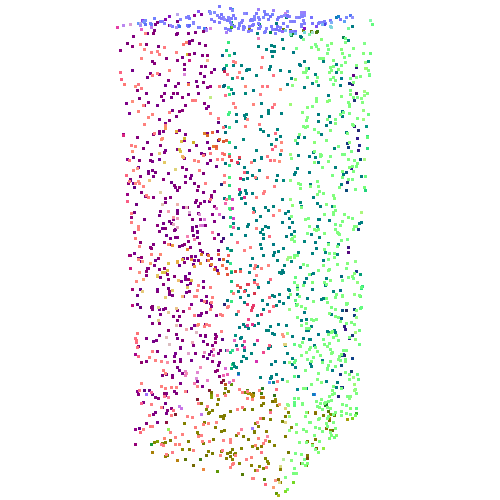}&
    \includegraphics[width=0.10\linewidth]{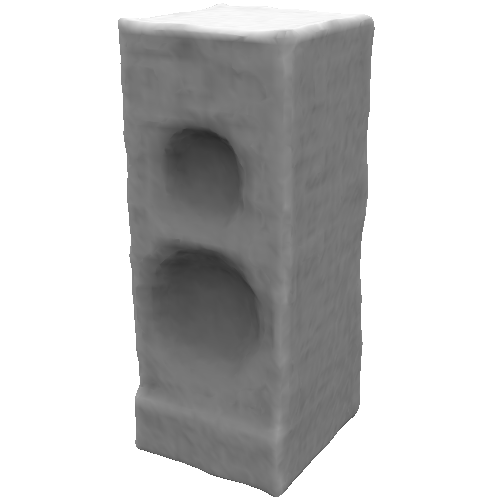}&
    \includegraphics[width=0.10\linewidth]{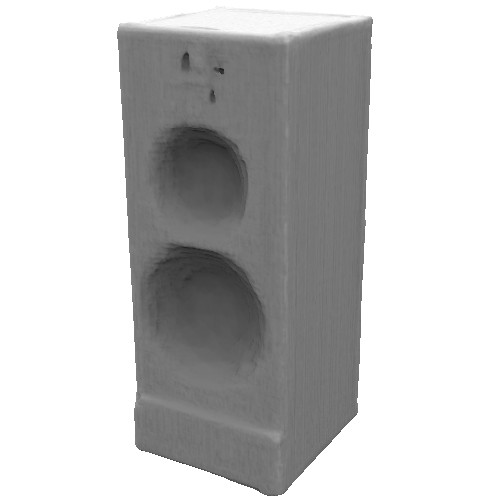}&
    \includegraphics[width=0.10\linewidth]{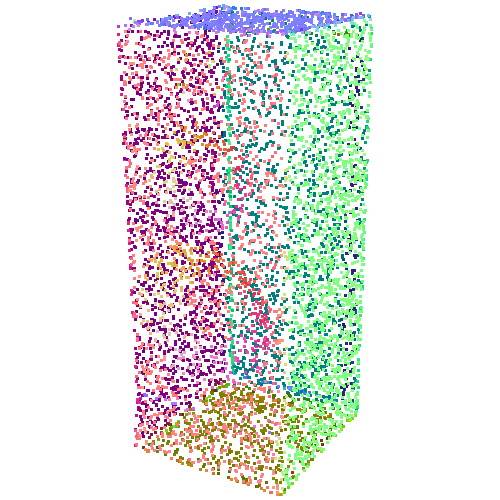}&
    \includegraphics[width=0.10\linewidth]{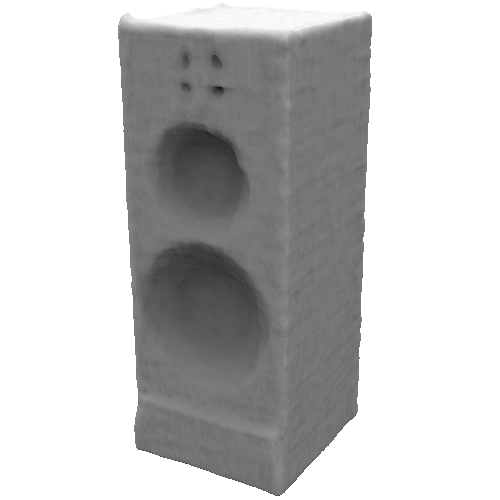}&
    \includegraphics[width=0.10\linewidth]{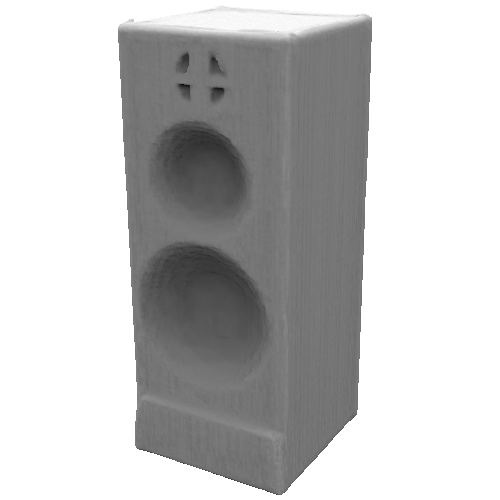}\\
    \includegraphics[width=0.10\linewidth]{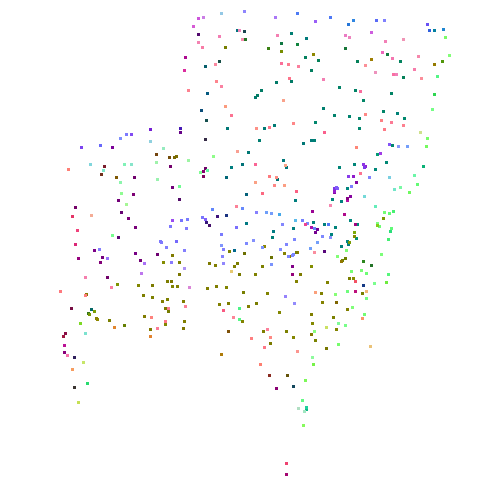}&
    \includegraphics[width=0.10\linewidth]{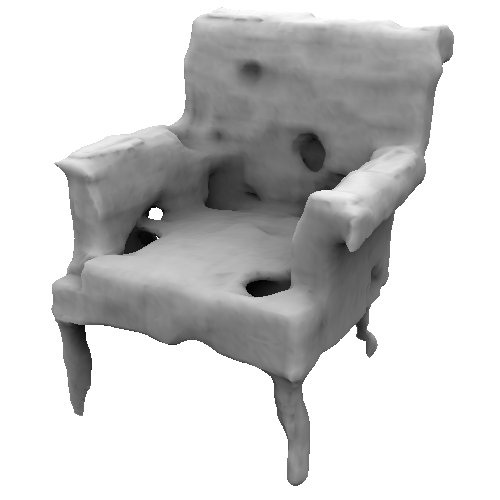}&
    \includegraphics[width=0.10\linewidth]{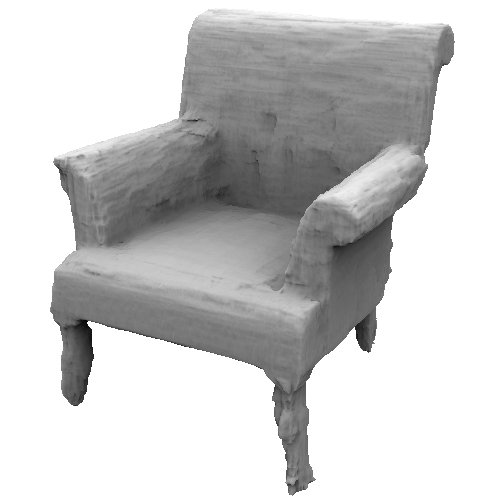}&
    \includegraphics[width=0.10\linewidth]{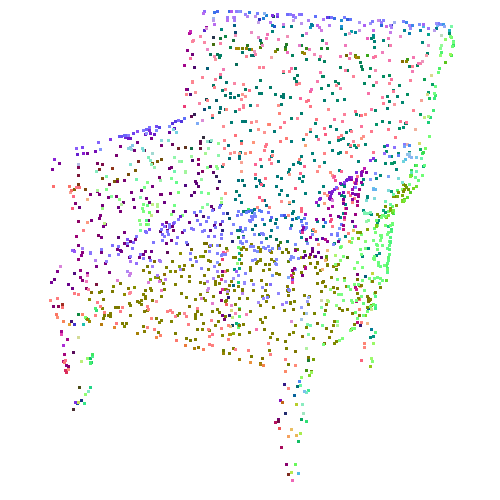}&
    \includegraphics[width=0.10\linewidth]{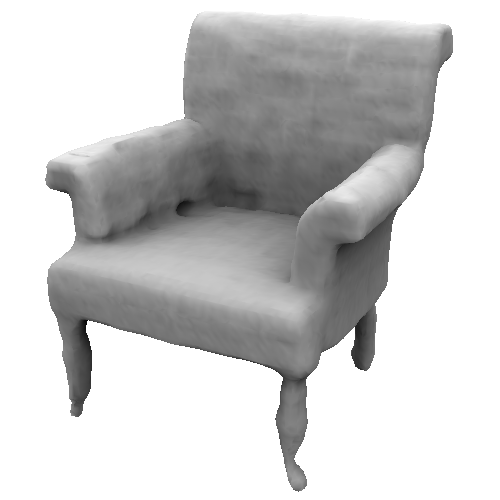}&
    \includegraphics[width=0.10\linewidth]{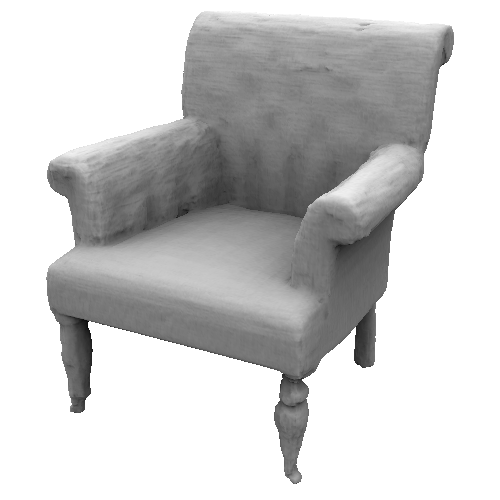}&
    \includegraphics[width=0.10\linewidth]{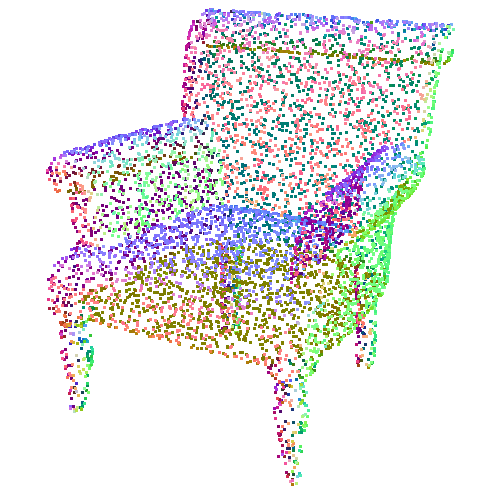}&
    \includegraphics[width=0.10\linewidth]{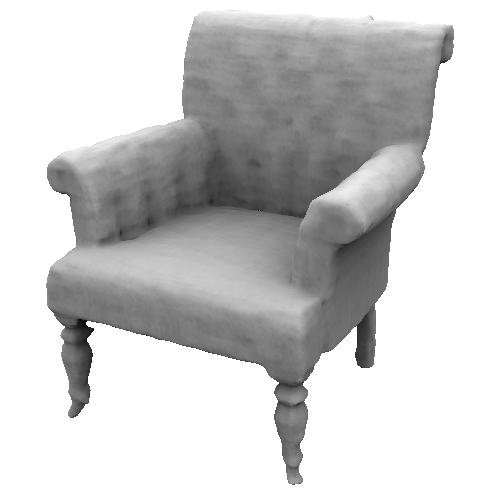}&
    \includegraphics[width=0.10\linewidth]{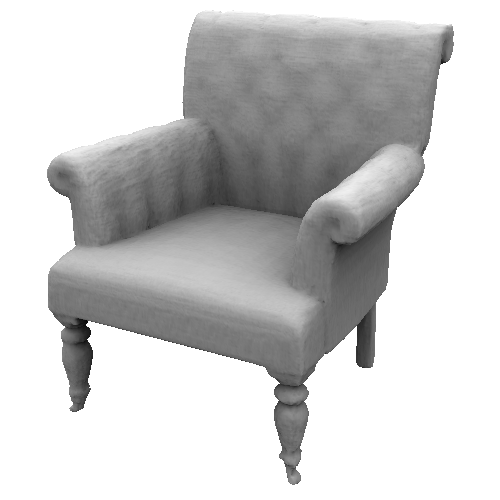}\\
    \includegraphics[width=0.10\linewidth]{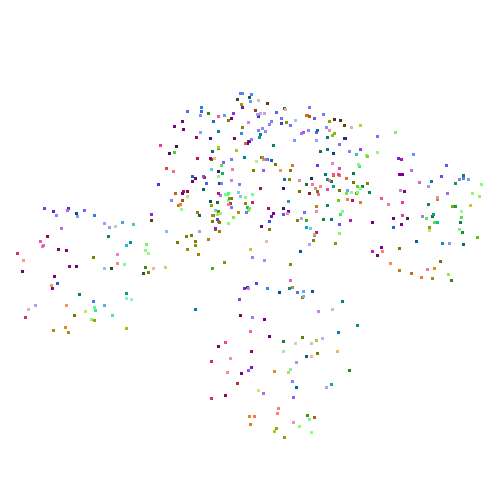}&
    \includegraphics[width=0.10\linewidth]{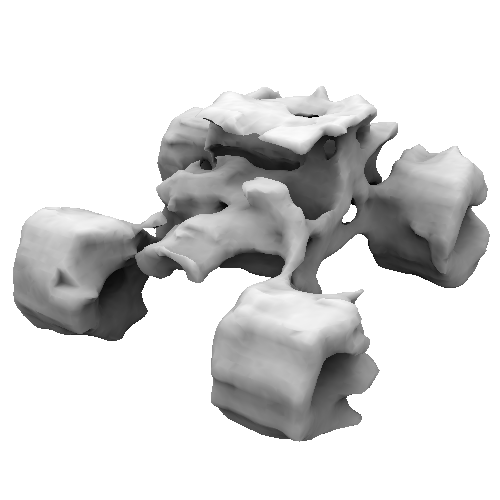}&
    \includegraphics[width=0.10\linewidth]{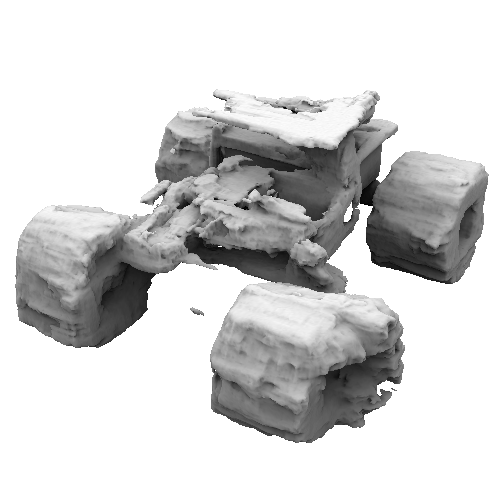}&
    \includegraphics[width=0.10\linewidth]{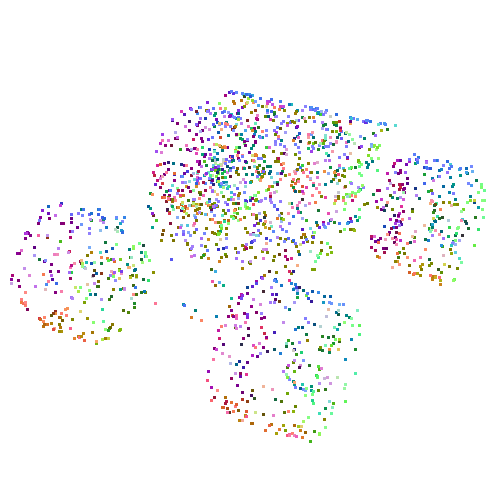}&
    \includegraphics[width=0.10\linewidth]{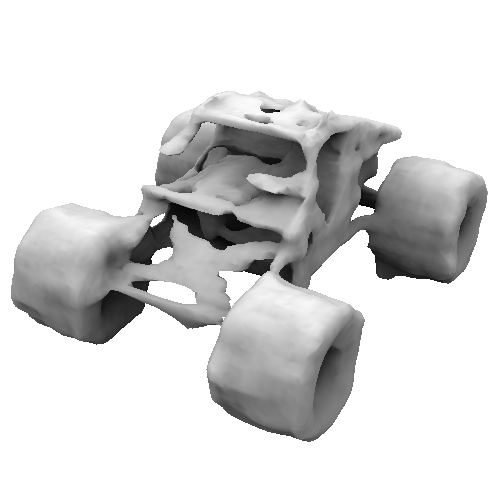}&
    \includegraphics[width=0.10\linewidth]{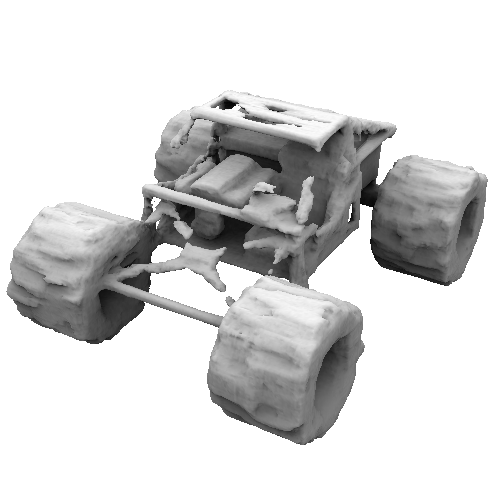}&
    \includegraphics[width=0.10\linewidth]{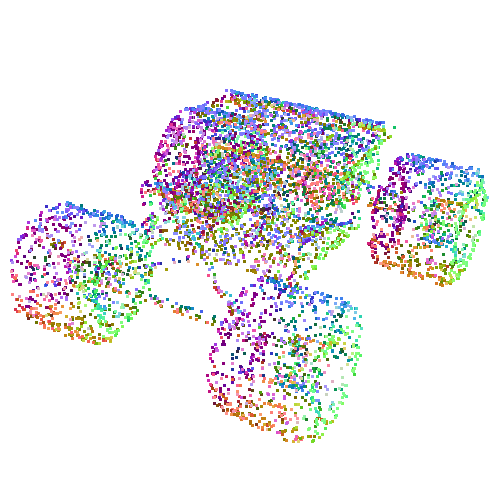}&
    \includegraphics[width=0.10\linewidth]{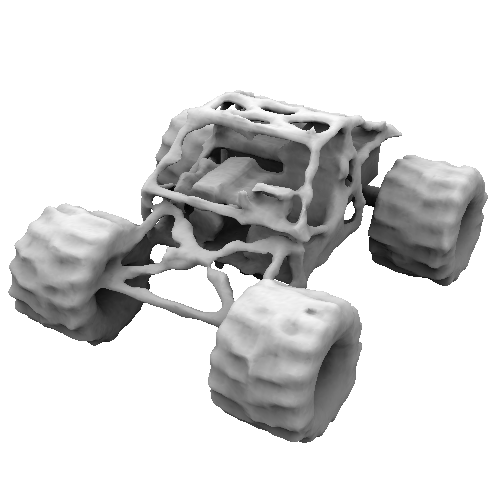}&
    \includegraphics[width=0.10\linewidth]{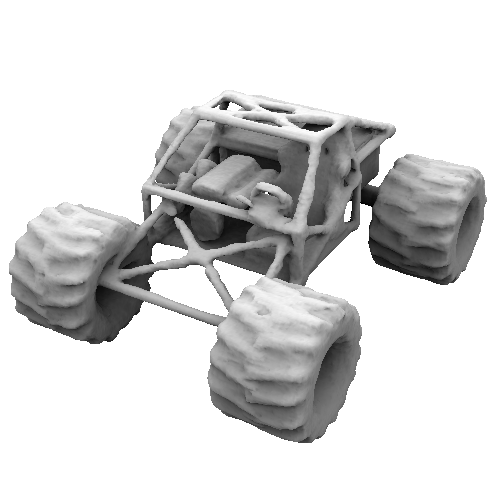}\\
    \includegraphics[width=0.10\linewidth]{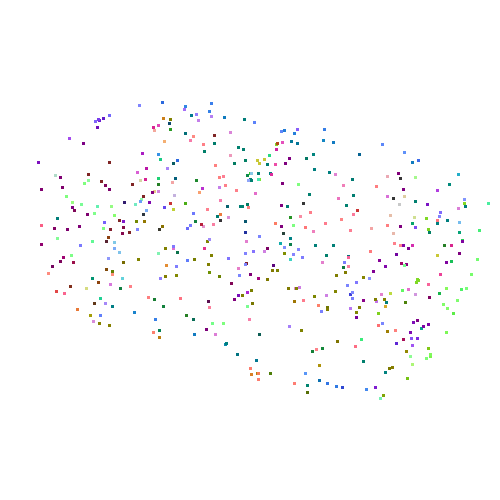}&
    \includegraphics[width=0.10\linewidth]{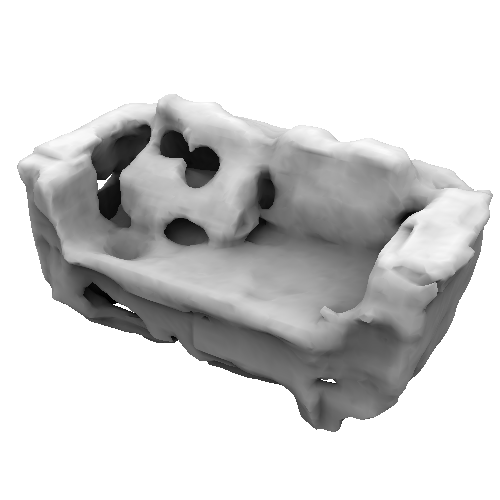}&
    \includegraphics[width=0.10\linewidth]{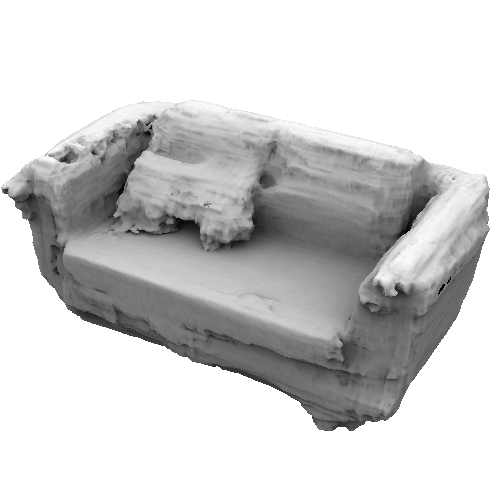}&
    \includegraphics[width=0.10\linewidth]{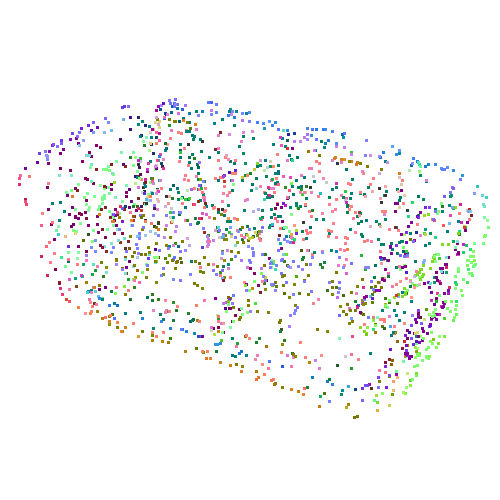}&
    \includegraphics[width=0.10\linewidth]{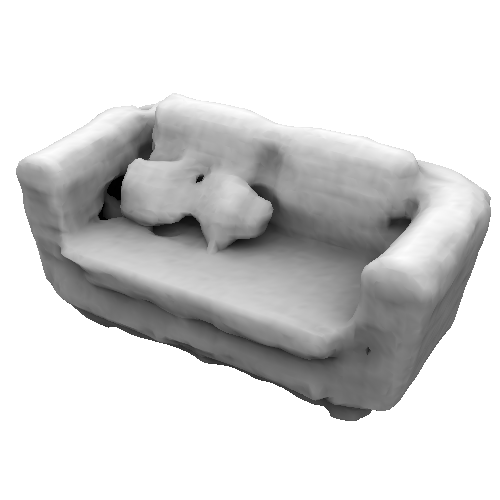}&
    \includegraphics[width=0.10\linewidth]{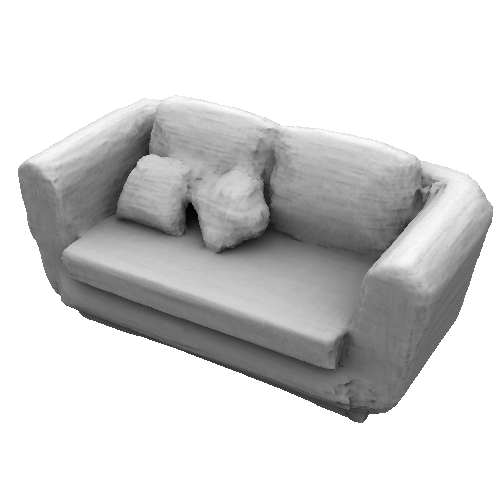}&
    \includegraphics[width=0.10\linewidth]{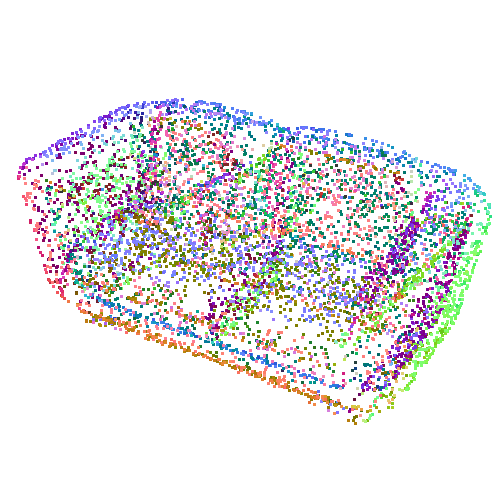}&
    \includegraphics[width=0.10\linewidth]{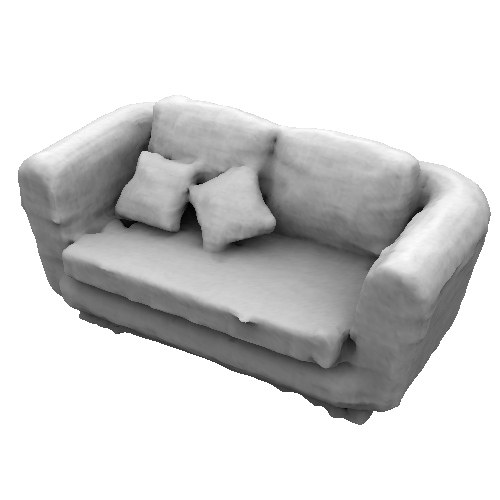}&
    \includegraphics[width=0.10\linewidth]{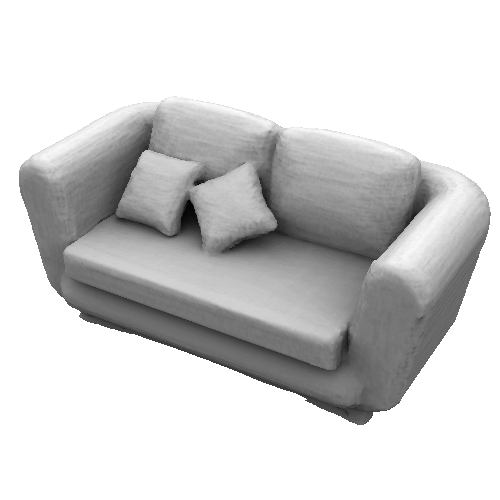}\\
    \includegraphics[width=0.10\linewidth]{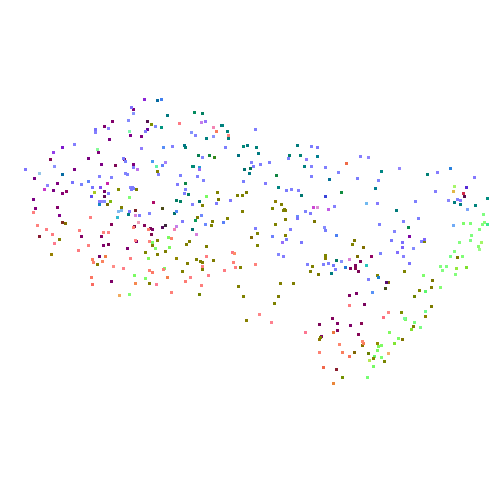}&
    \includegraphics[width=0.10\linewidth]{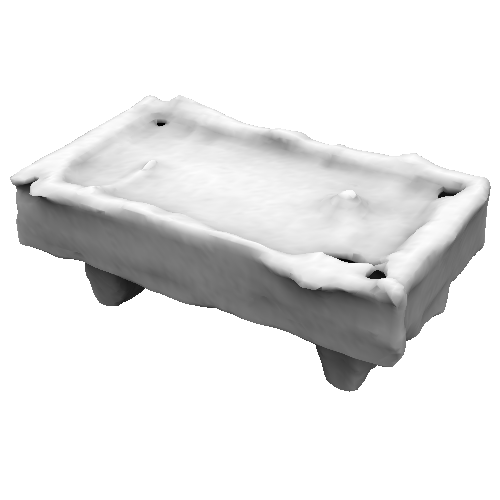}&
    \includegraphics[width=0.10\linewidth]{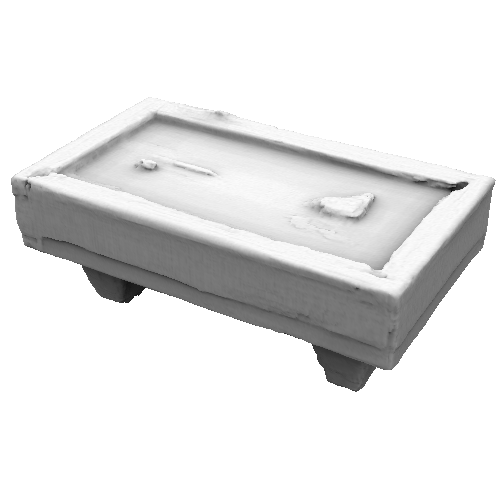}&
    \includegraphics[width=0.10\linewidth]{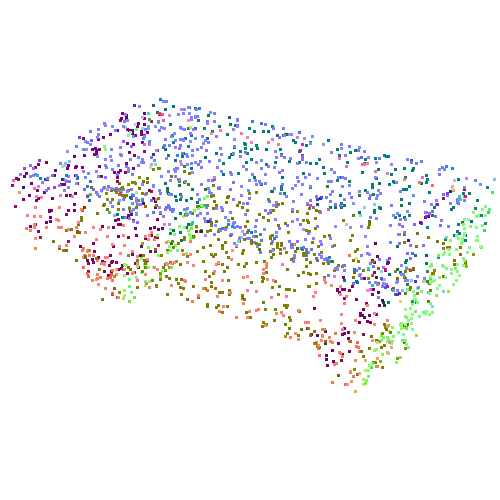}&
    \includegraphics[width=0.10\linewidth]{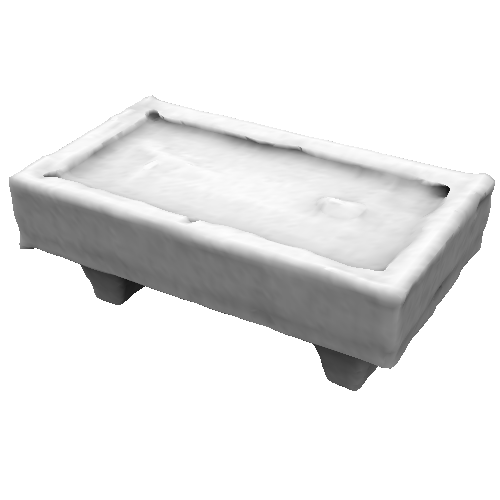}&
    \includegraphics[width=0.10\linewidth]{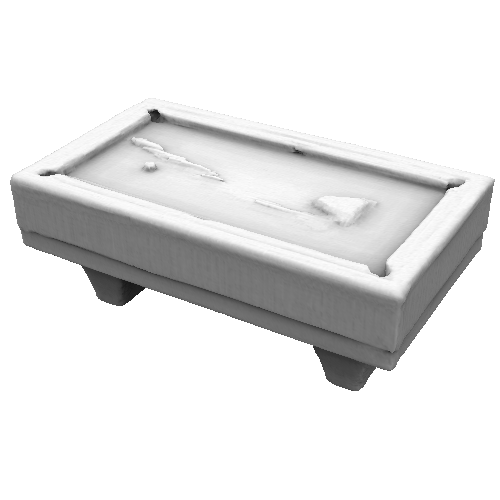}&
    \includegraphics[width=0.10\linewidth]{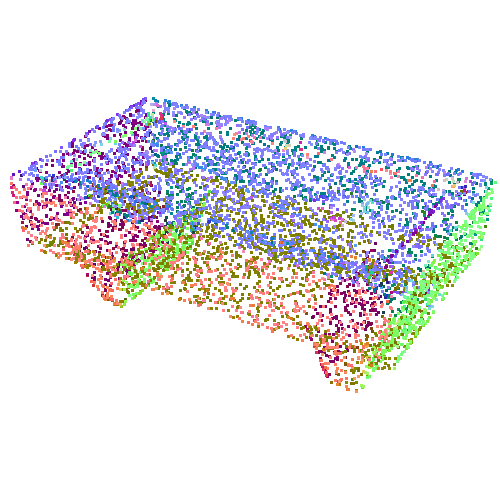}&
    \includegraphics[width=0.10\linewidth]{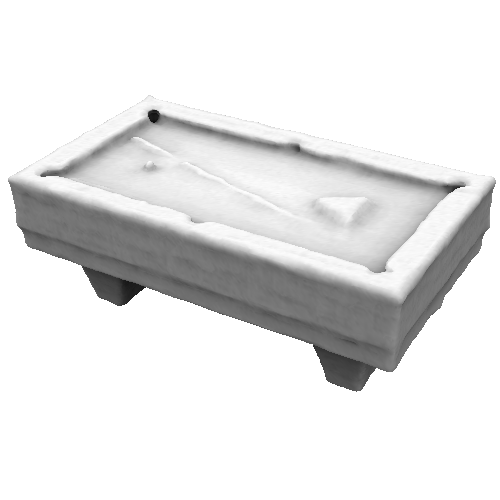}&
    \includegraphics[width=0.10\linewidth]{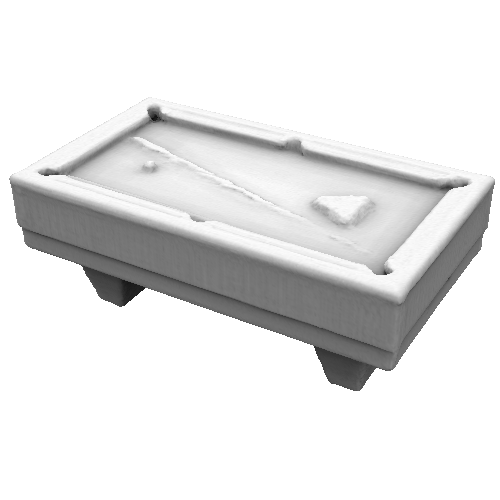}\\
    \includegraphics[width=0.10\linewidth]{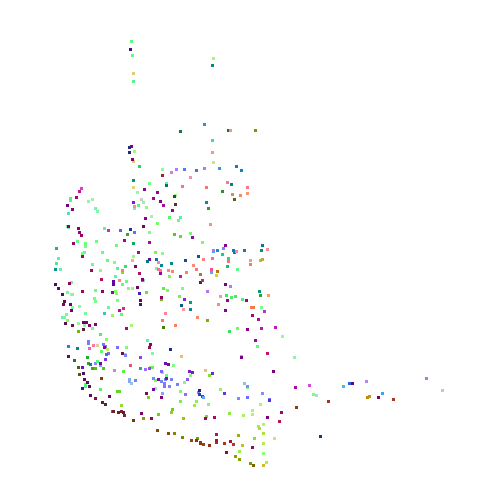}&
    \includegraphics[width=0.10\linewidth]{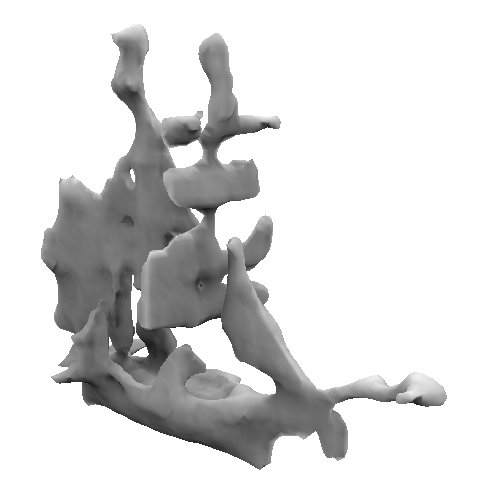}&
    \includegraphics[width=0.10\linewidth]{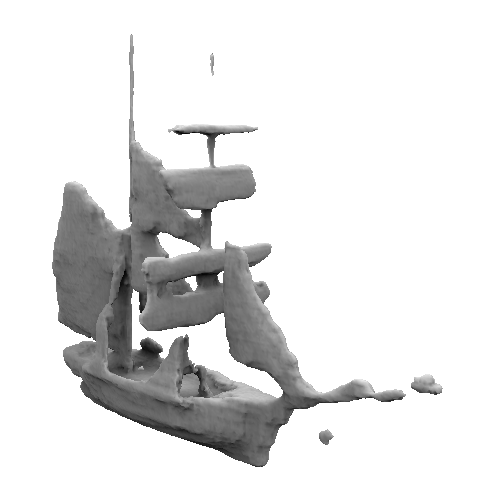}&
    \includegraphics[width=0.10\linewidth]{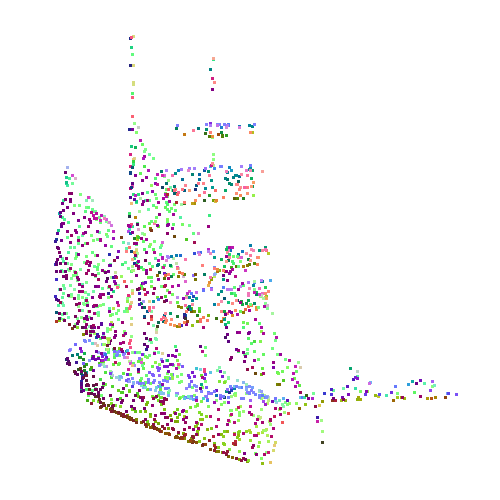}&
    \includegraphics[width=0.10\linewidth]{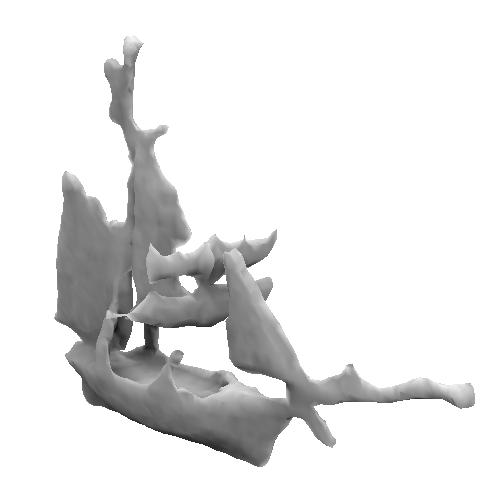}&
    \includegraphics[width=0.10\linewidth]{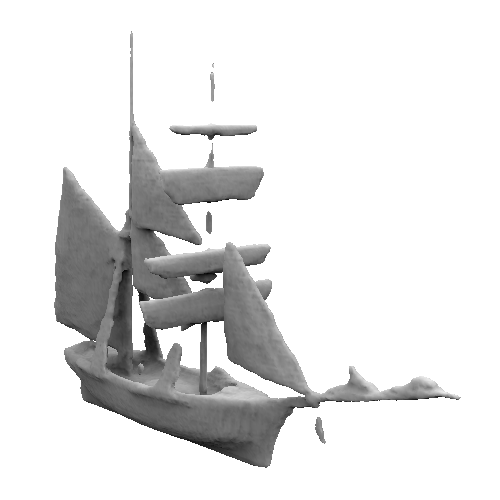}&
    \includegraphics[width=0.10\linewidth]{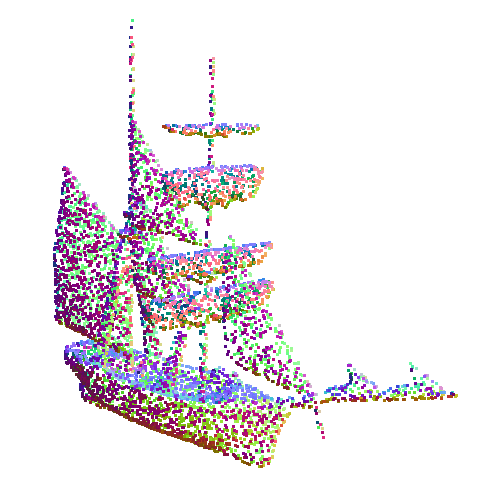}&
    \includegraphics[width=0.10\linewidth]{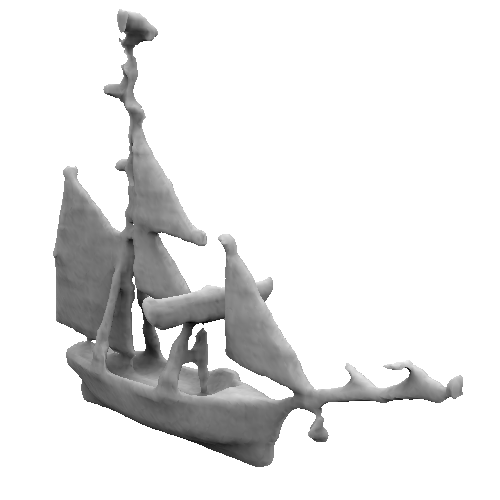}&
    \includegraphics[width=0.10\linewidth]{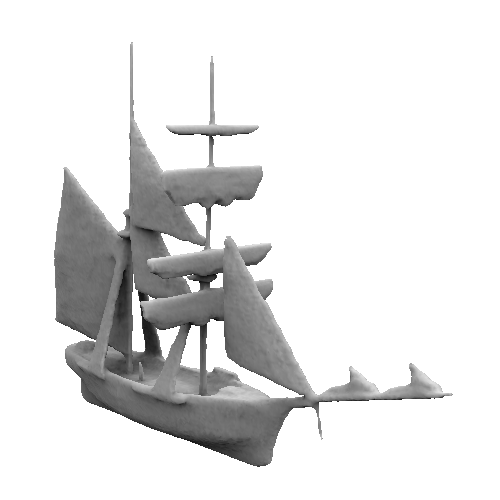}\\
    \includegraphics[width=0.10\linewidth]{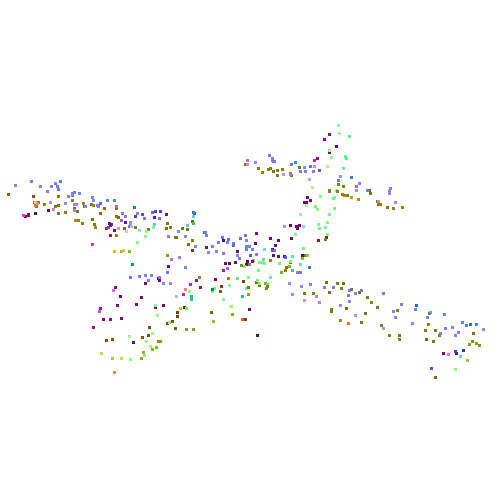}&
    \includegraphics[width=0.10\linewidth]{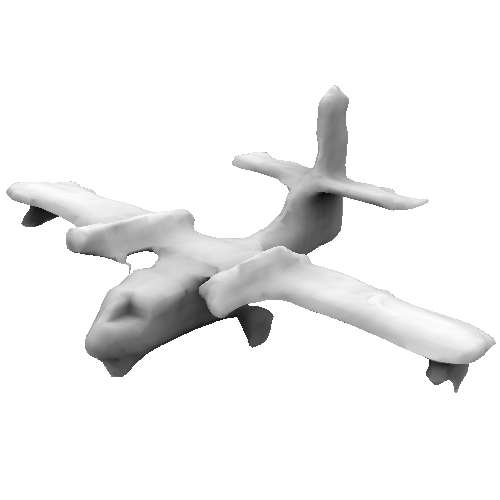}&
    \includegraphics[width=0.10\linewidth]{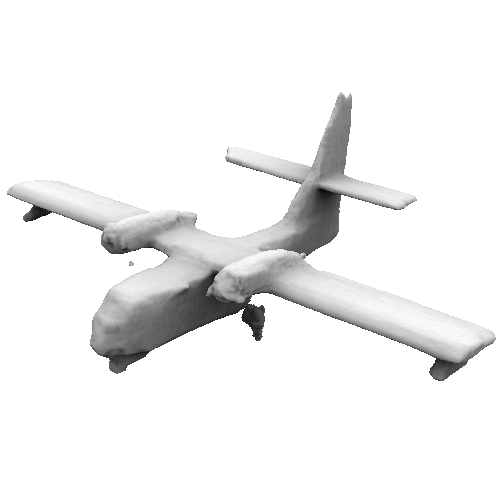}&
    \includegraphics[width=0.10\linewidth]{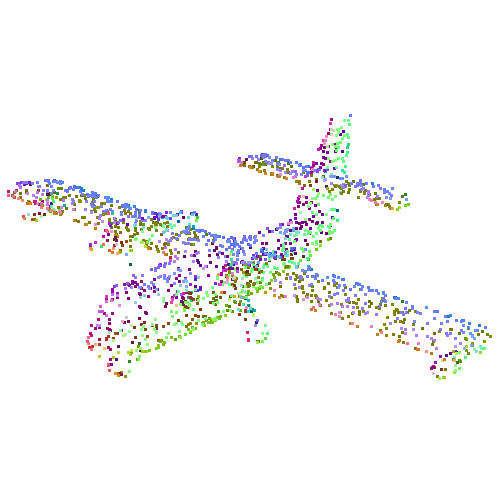}&
    \includegraphics[width=0.10\linewidth]{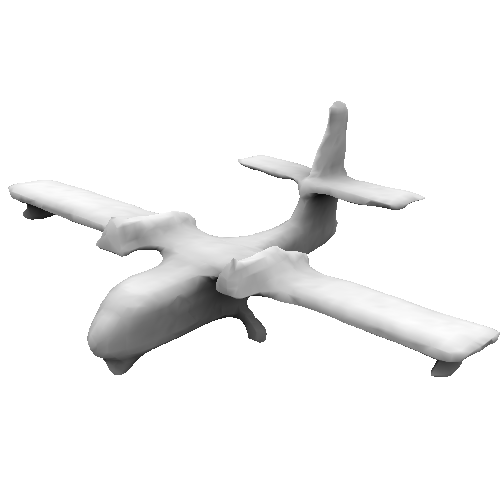}&
    \includegraphics[width=0.10\linewidth]{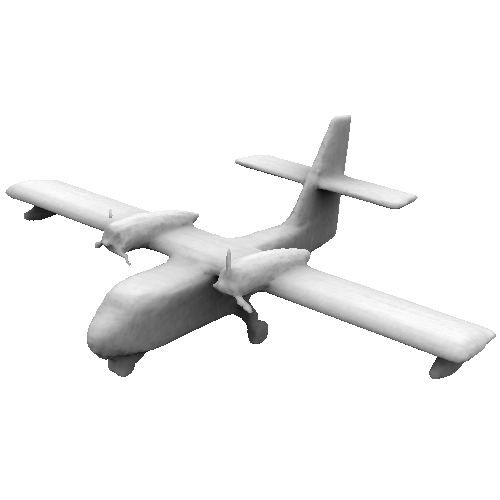}&
    \includegraphics[width=0.10\linewidth]{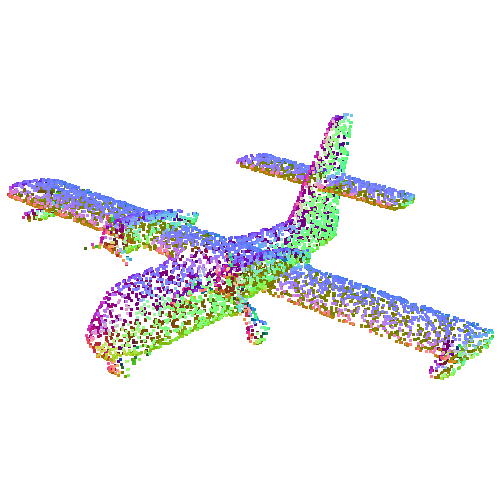}&
    \includegraphics[width=0.10\linewidth]{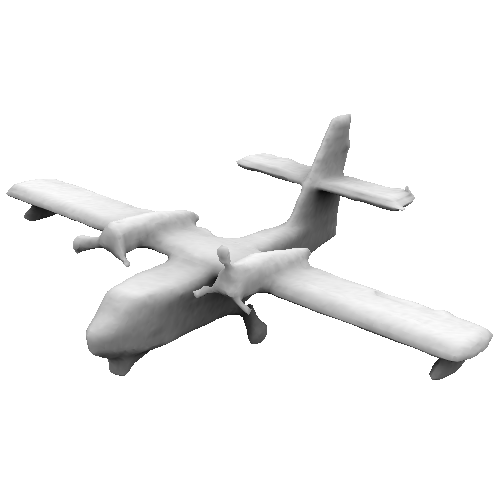}&
    \includegraphics[width=0.10\linewidth]{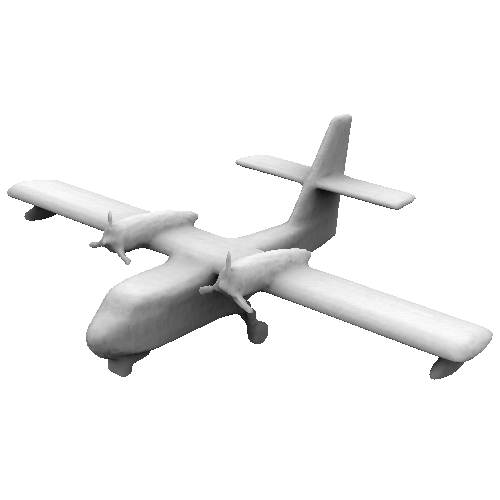}\\
    \includegraphics[width=0.10\linewidth]{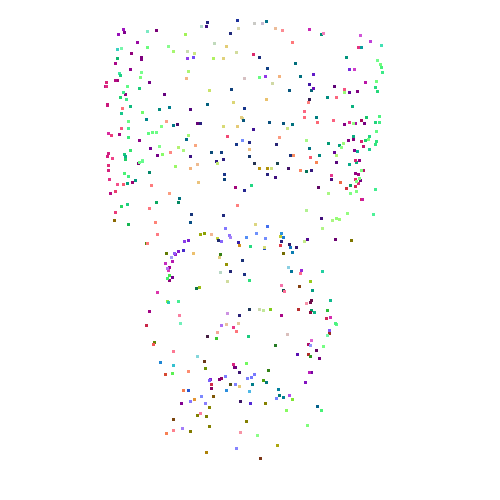}&
    \includegraphics[width=0.10\linewidth]{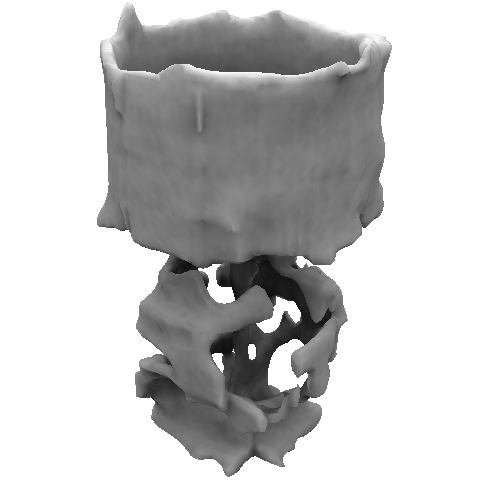}&
    \includegraphics[width=0.10\linewidth]{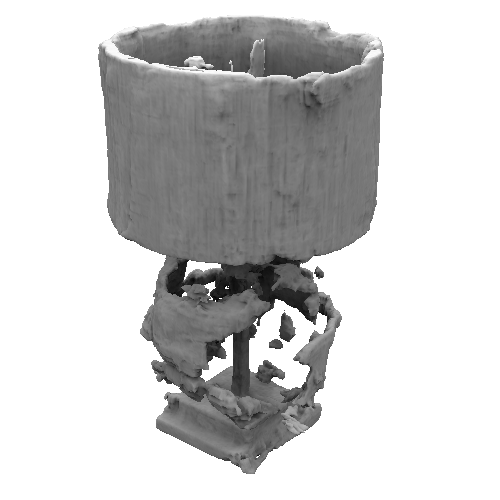}&
    \includegraphics[width=0.10\linewidth]{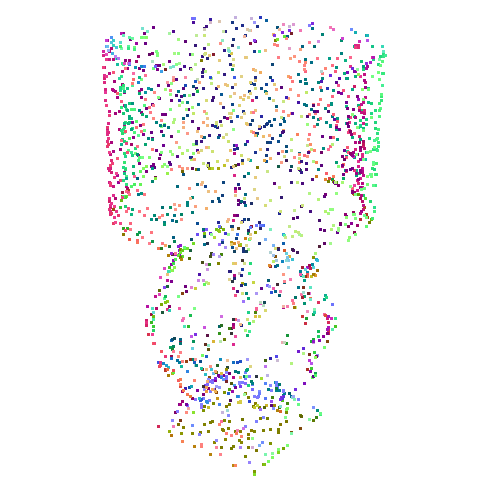}&
    \includegraphics[width=0.10\linewidth]{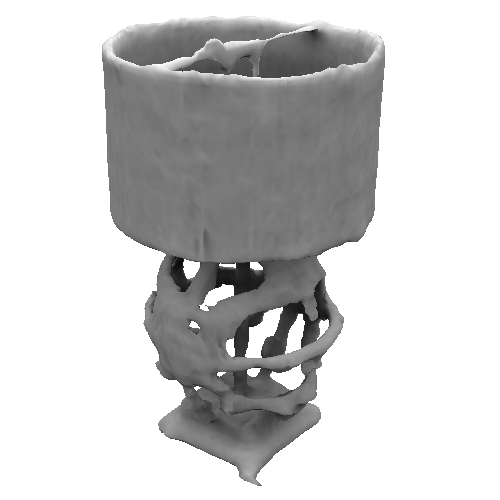}&
    \includegraphics[width=0.10\linewidth]{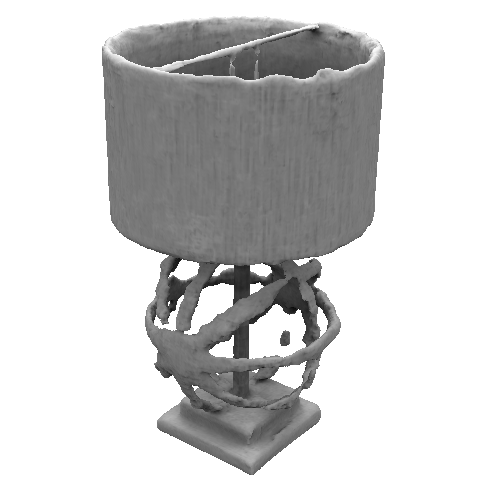}&
    \includegraphics[width=0.10\linewidth]{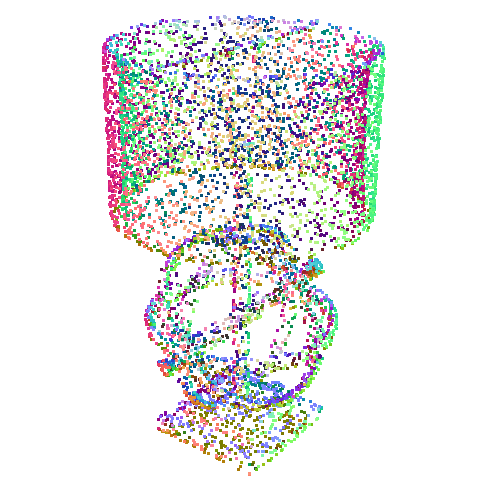}&
    \includegraphics[width=0.10\linewidth]{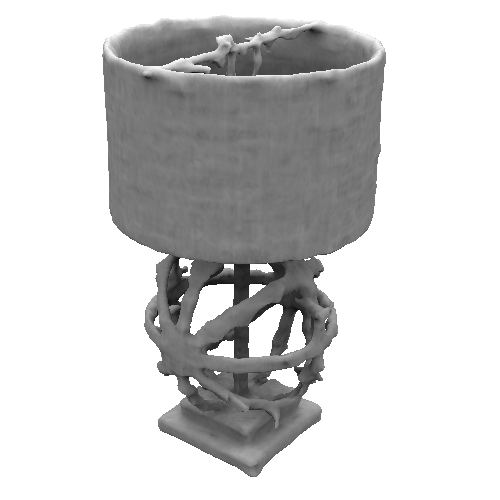}&
    \includegraphics[width=0.10\linewidth]{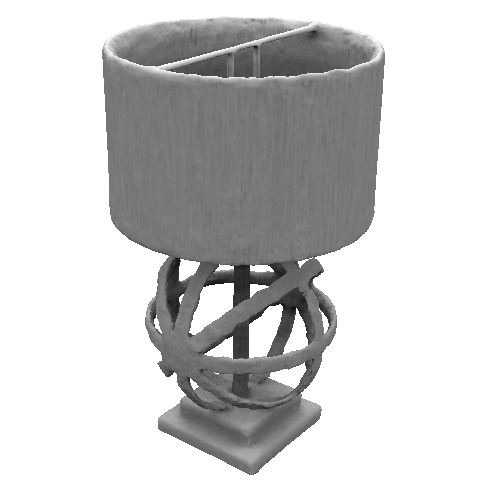}\\
    \includegraphics[width=0.10\linewidth]{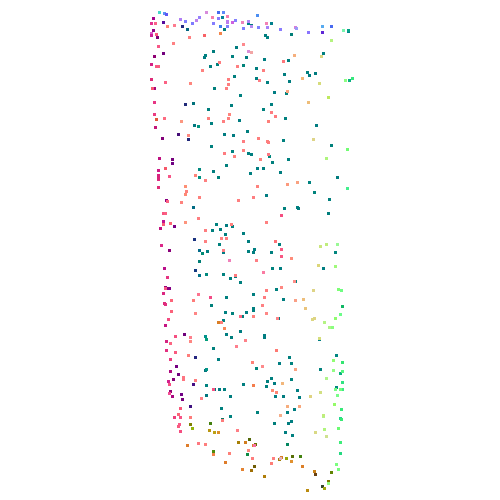}&
    \includegraphics[width=0.10\linewidth]{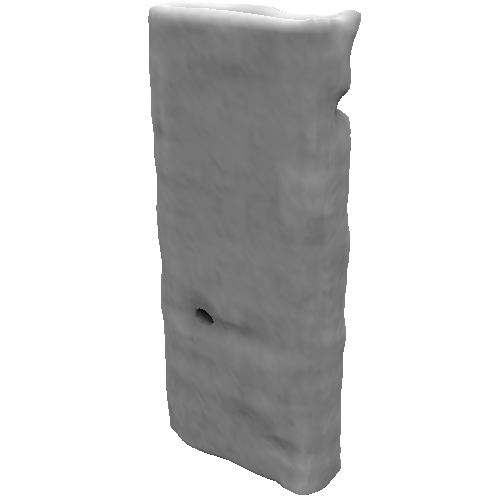}&
    \includegraphics[width=0.10\linewidth]{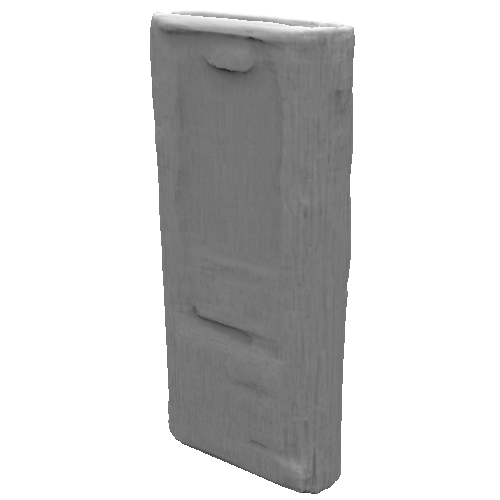}&
    \includegraphics[width=0.10\linewidth]{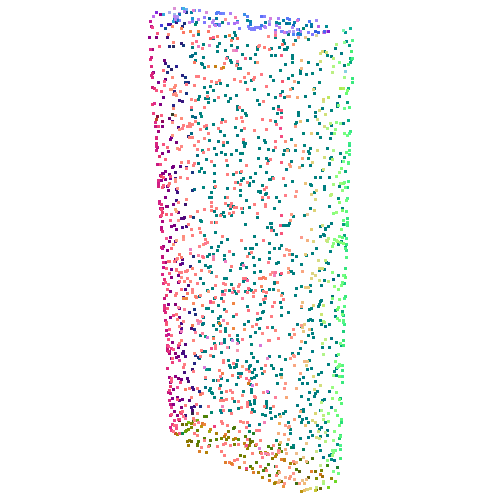}&
    \includegraphics[width=0.10\linewidth]{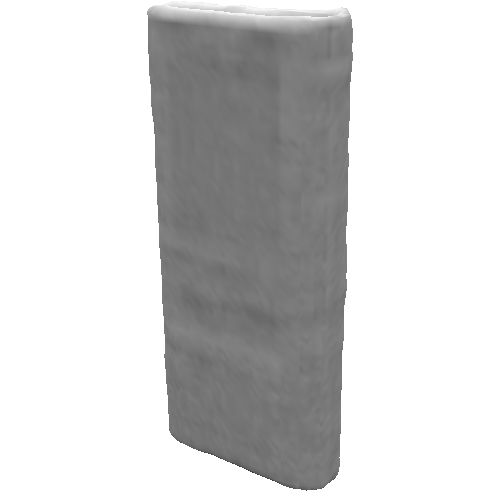}&
    \includegraphics[width=0.10\linewidth]{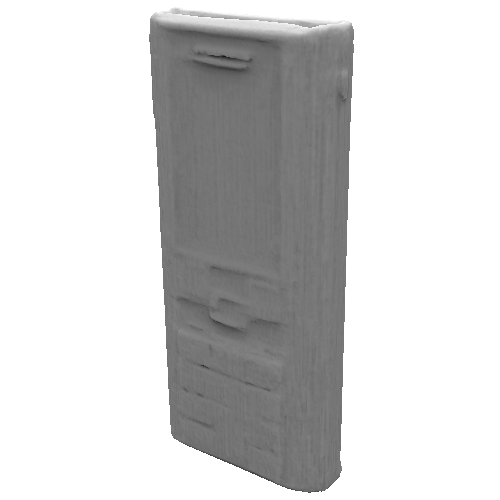}&
    \includegraphics[width=0.10\linewidth]{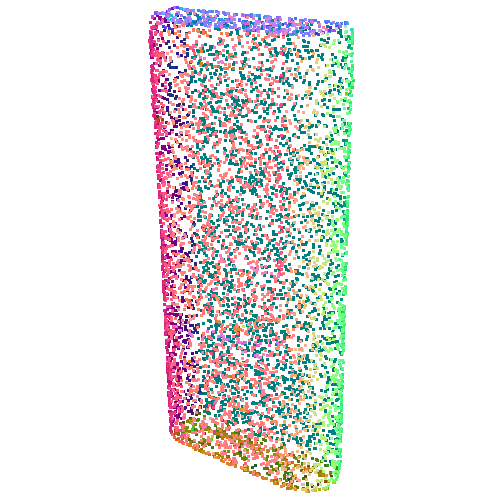}&
    \includegraphics[width=0.10\linewidth]{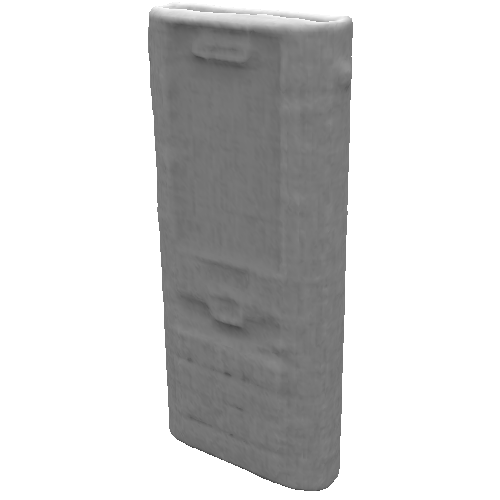}&
    \includegraphics[width=0.10\linewidth]{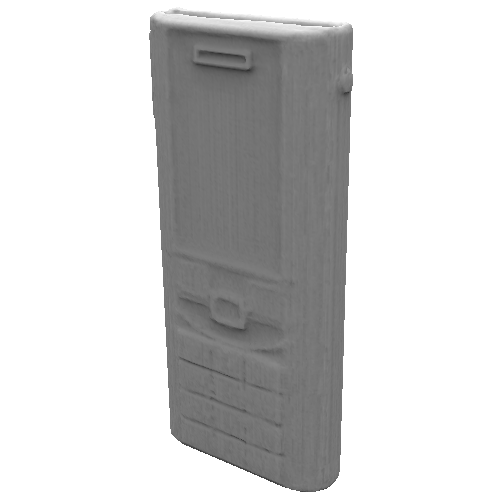}\\
    \includegraphics[width=0.10\linewidth]{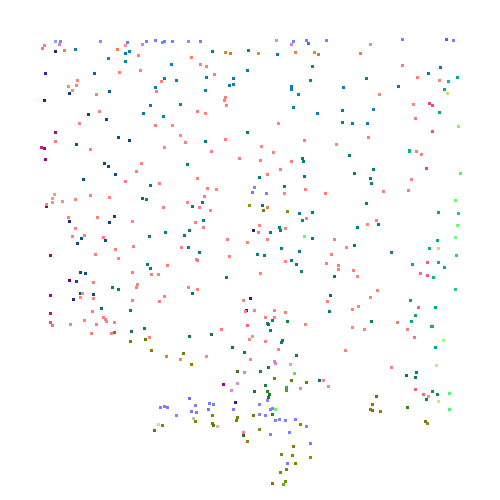}&
    \includegraphics[width=0.10\linewidth]{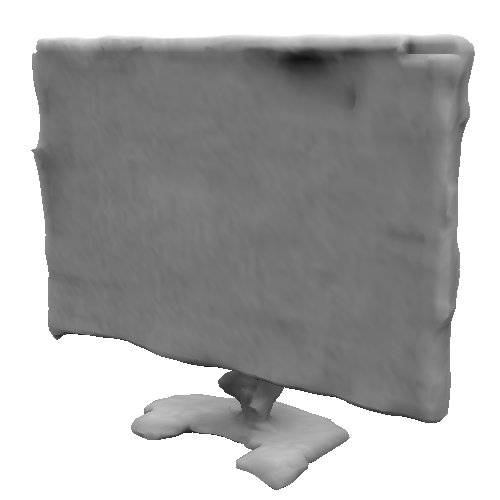}&
    \includegraphics[width=0.10\linewidth]{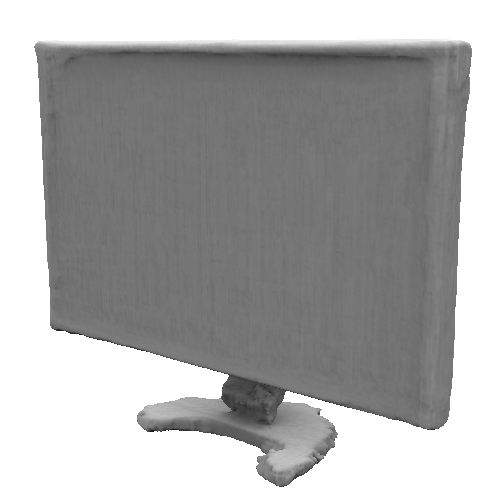}&
    \includegraphics[width=0.10\linewidth]{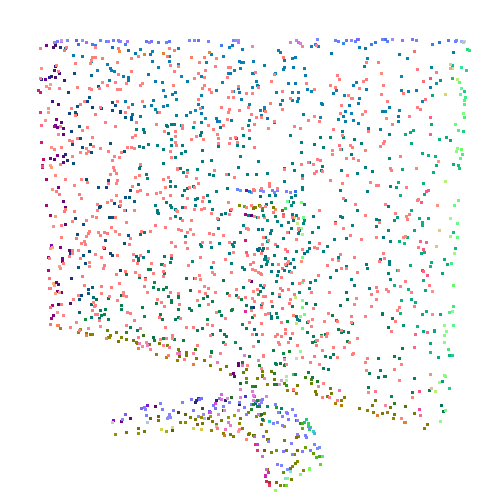}&
    \includegraphics[width=0.10\linewidth]{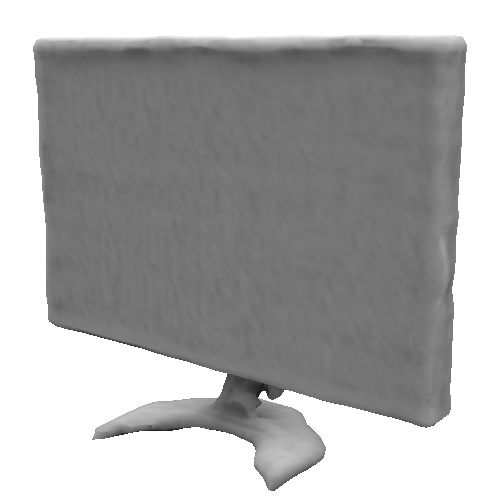}&
    \includegraphics[width=0.10\linewidth]{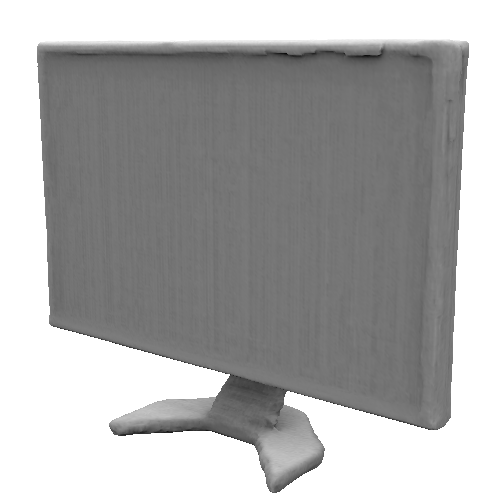}&
    \includegraphics[width=0.10\linewidth]{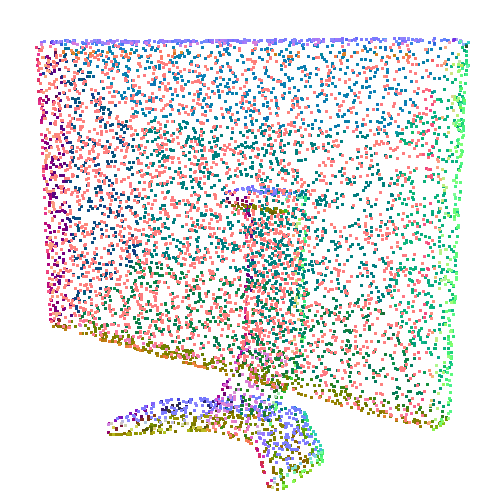}&
    \includegraphics[width=0.10\linewidth]{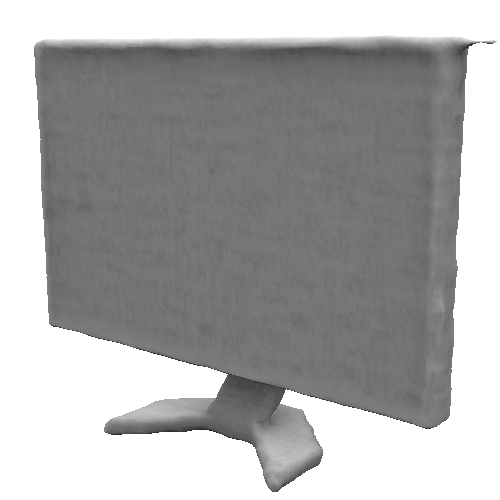}&
    \includegraphics[width=0.10\linewidth]{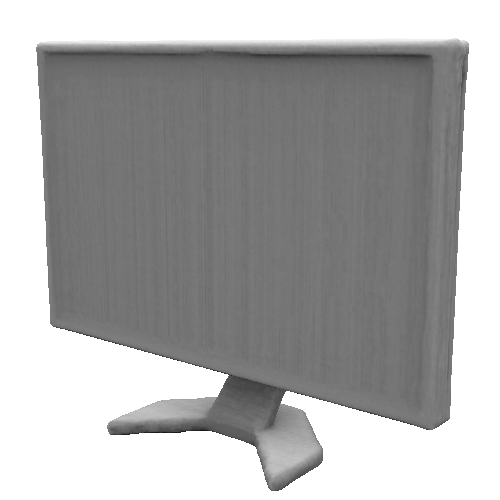}
\end{tabular}
\caption{\textbf{ShapeNet reconstructions} (input with normals), LIG (part size 0.20 for 512 and 2048 pts, 0.10 for 8192 pts) and {\OURS} (ours).}
\label{fig:supp:shapenet_lig}
\end{figure*}

\begin{figure*}[p]
    \centering
    \caption{\textbf{Reconstruction fragment of a SceneNet scene} with varying input point densities for SPR, LIG and {\OURS}.}
    \label{fig:scenenet2}
    \begin{tabular}{ccccc}
    & Input & SPR & LIG &{\OURS}
    \\
    \rotatebox{90}{\hspace{12mm} 20 pts/m$^2$} &
    \includegraphics[width=0.20\linewidth]{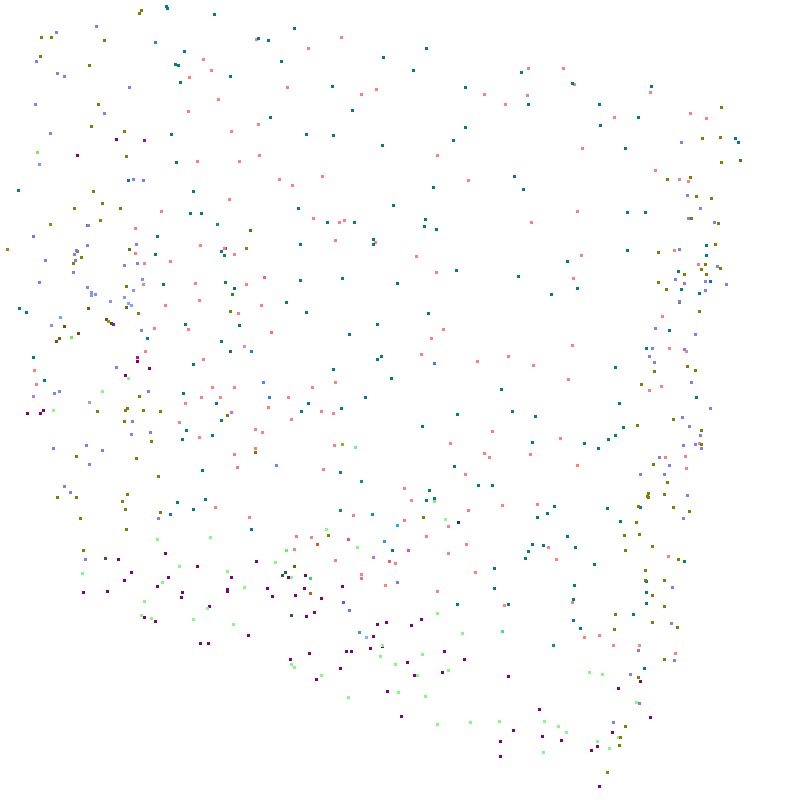}&
    \includegraphics[width=0.20\linewidth]{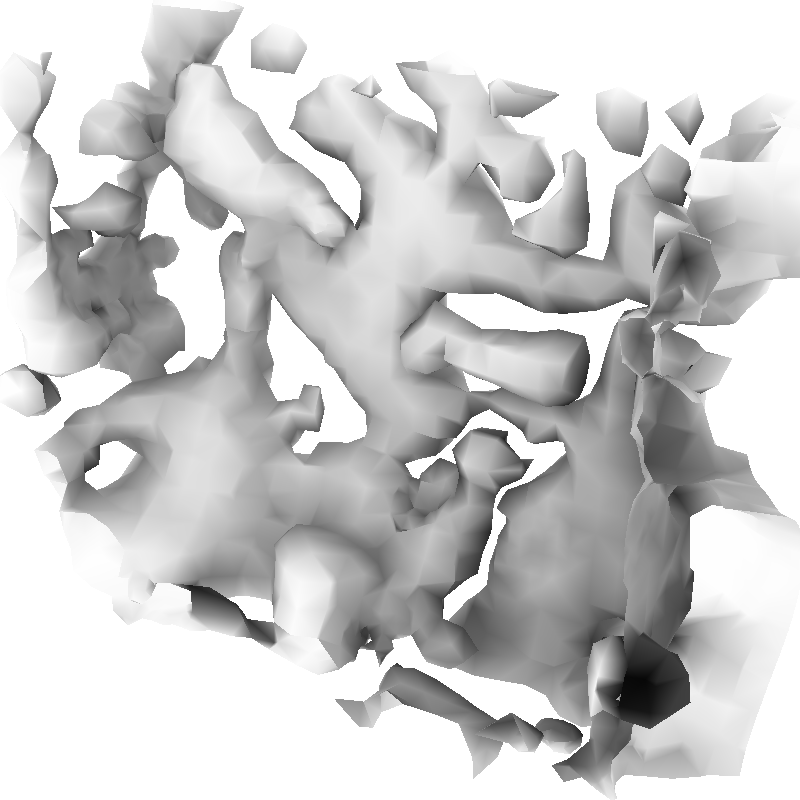}&
    \includegraphics[width=0.20\linewidth]{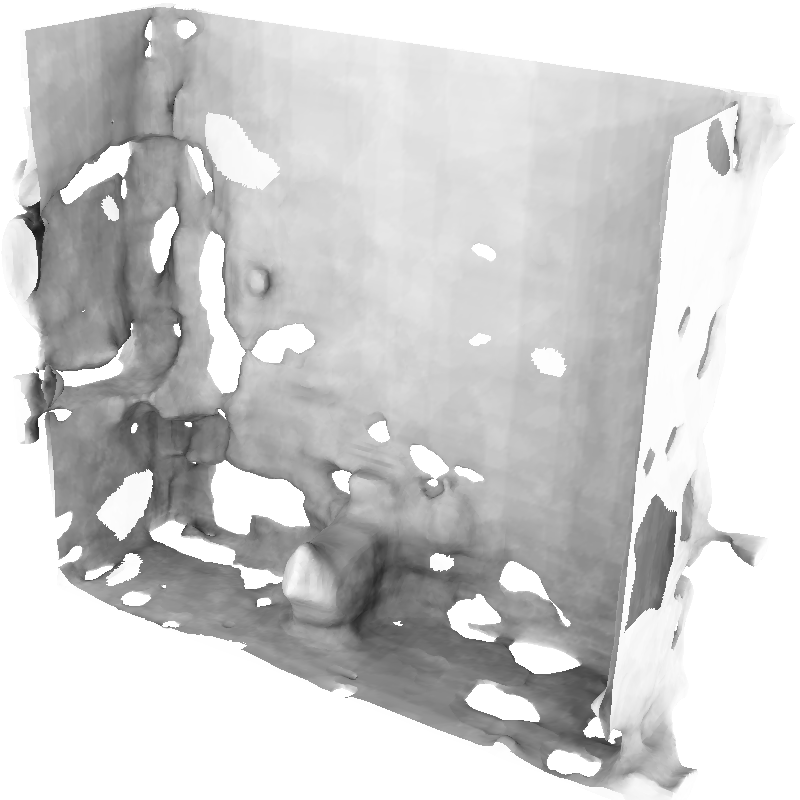}&
    \includegraphics[width=0.20\linewidth]{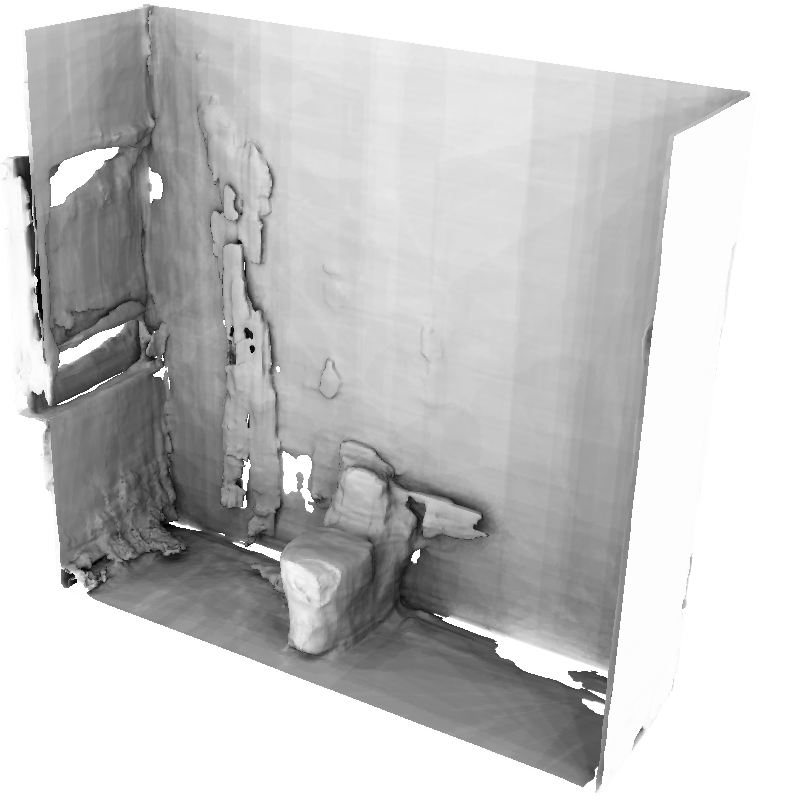} 
    \\
    \rotatebox{90}{\hspace{12mm} 100 pts/m$^2$} &
    \includegraphics[width=0.20\linewidth]{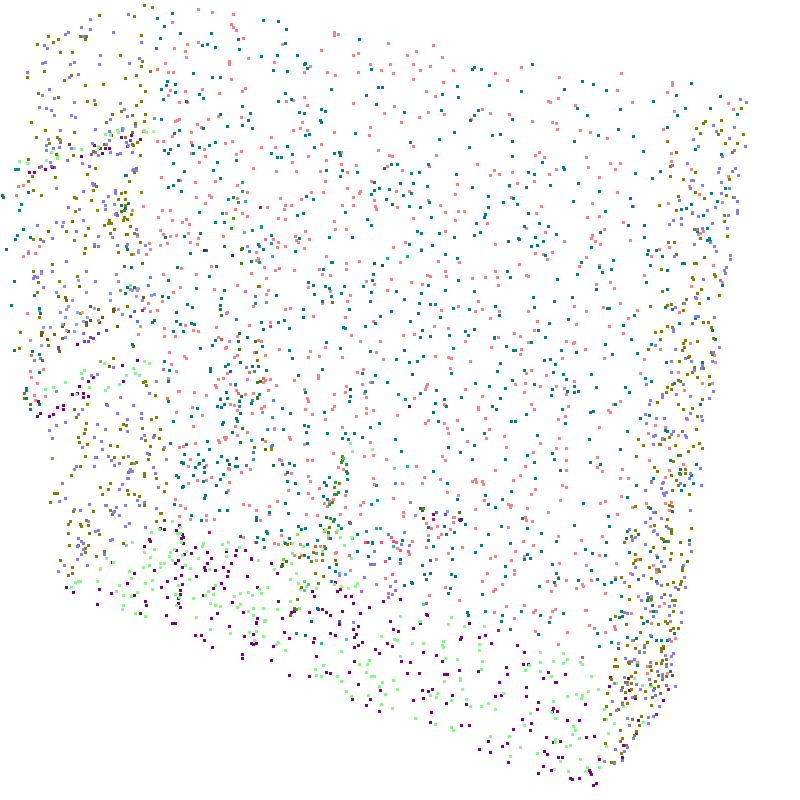}&
    \includegraphics[width=0.20\linewidth]{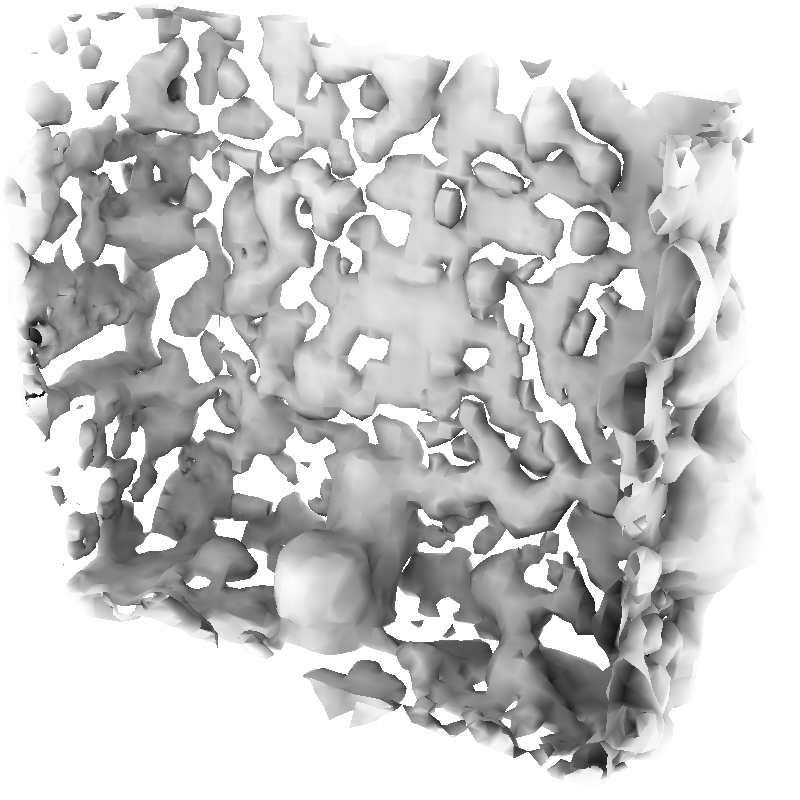}&
    \includegraphics[width=0.20\linewidth]{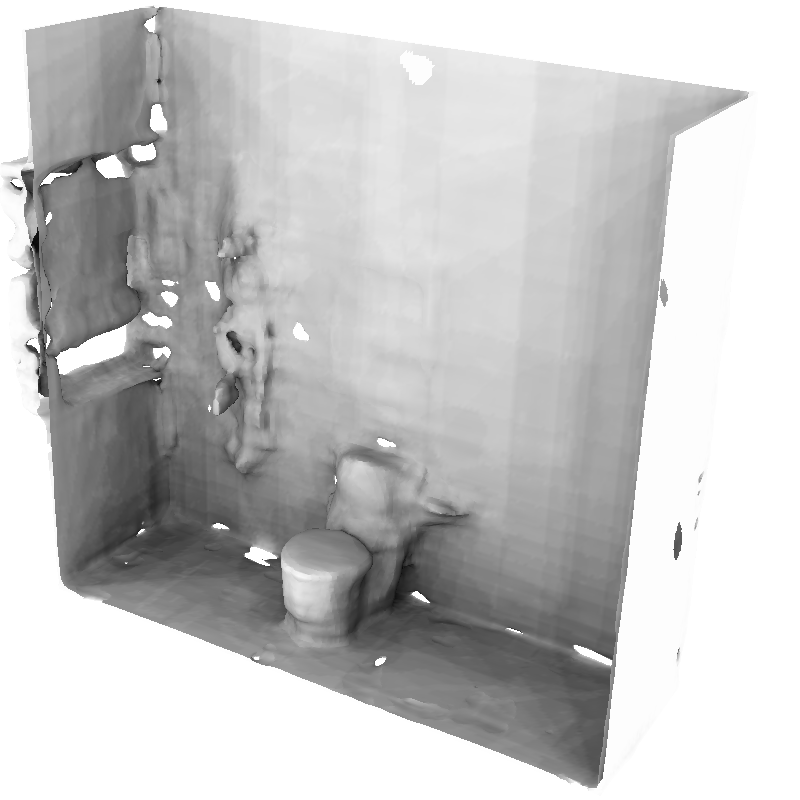}&
    \includegraphics[width=0.20\linewidth]{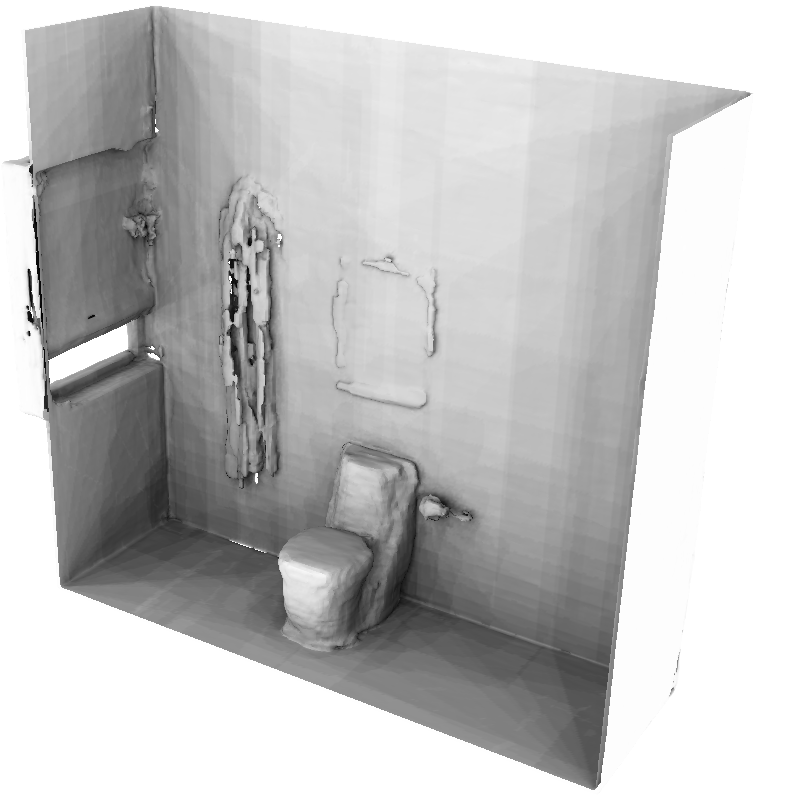} 
    \\
    \rotatebox{90}{\hspace{12mm} 500 pts/m$^2$} &
    \includegraphics[width=0.20\linewidth]{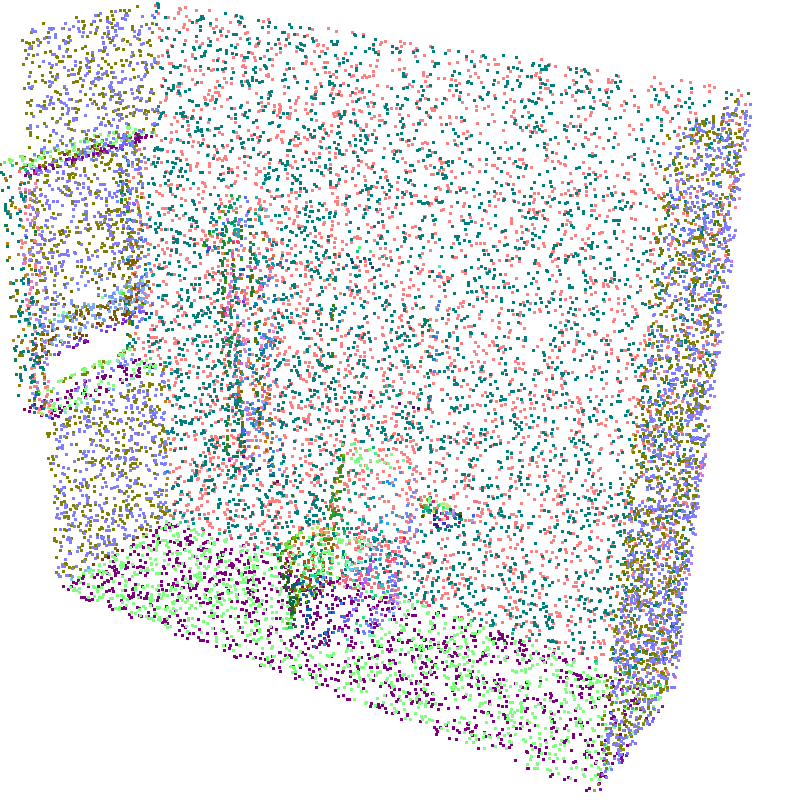}&
    \includegraphics[width=0.20\linewidth]{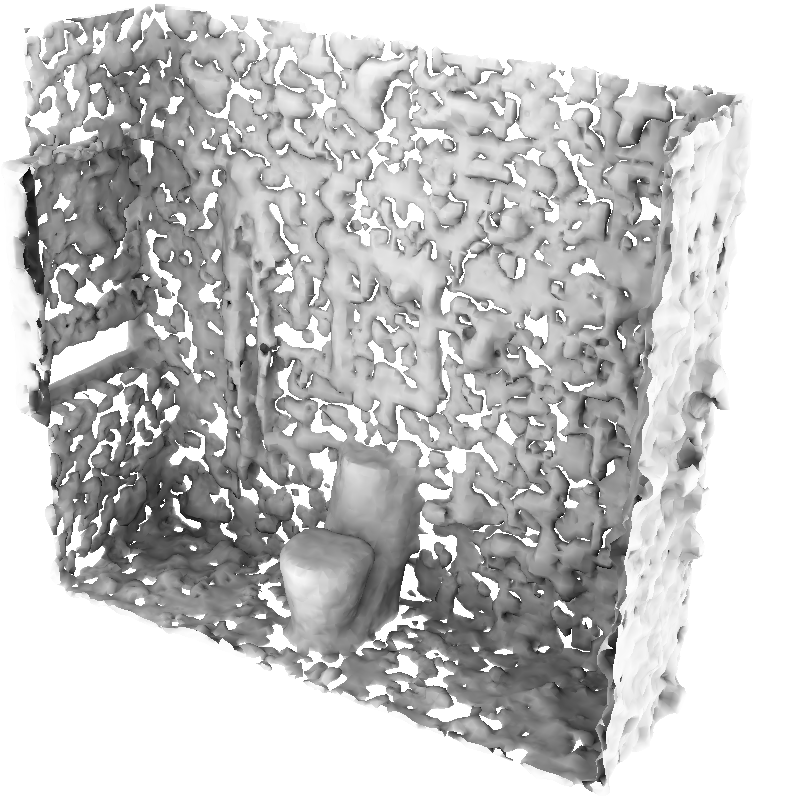}&
    \includegraphics[width=0.20\linewidth]{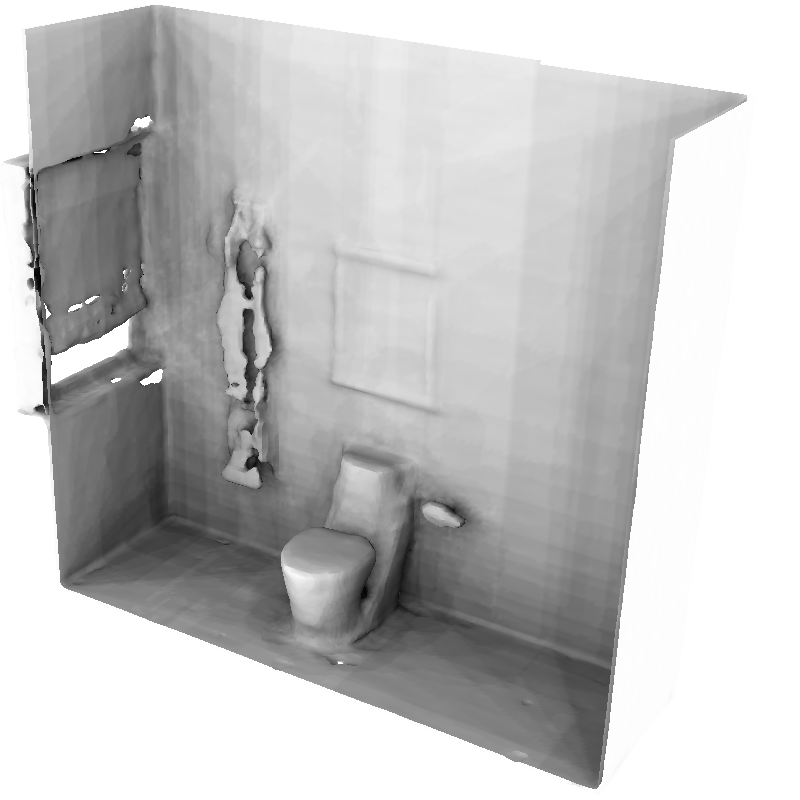}&
    \includegraphics[width=0.20\linewidth]{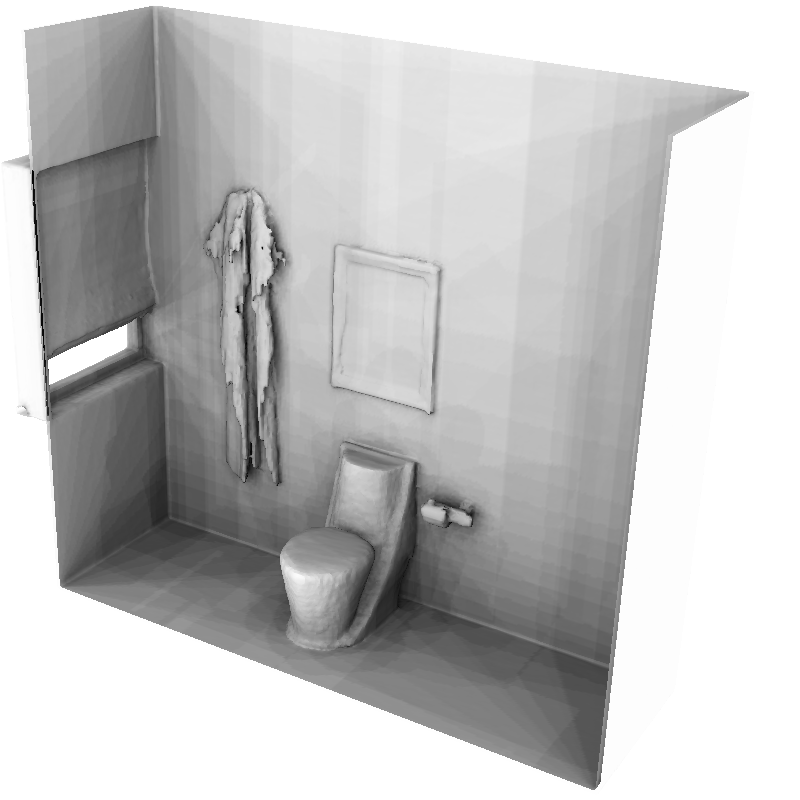} 
    \\
    \rotatebox{90}{\hspace{12mm} 1000 pts/m$^2$} &
    \includegraphics[width=0.20\linewidth]{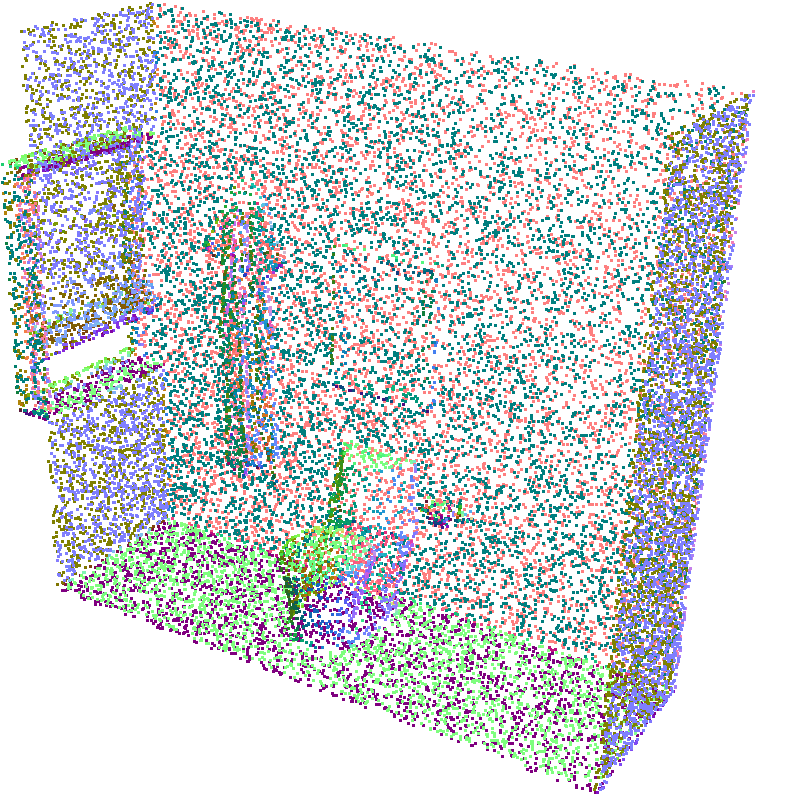}&
    \includegraphics[width=0.20\linewidth]{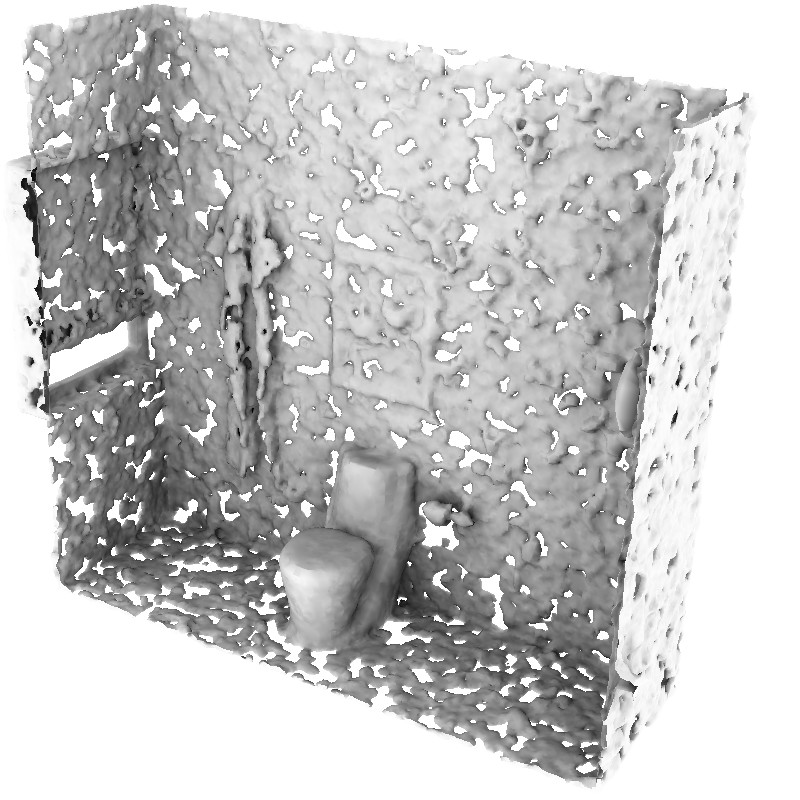}&
    \includegraphics[width=0.20\linewidth]{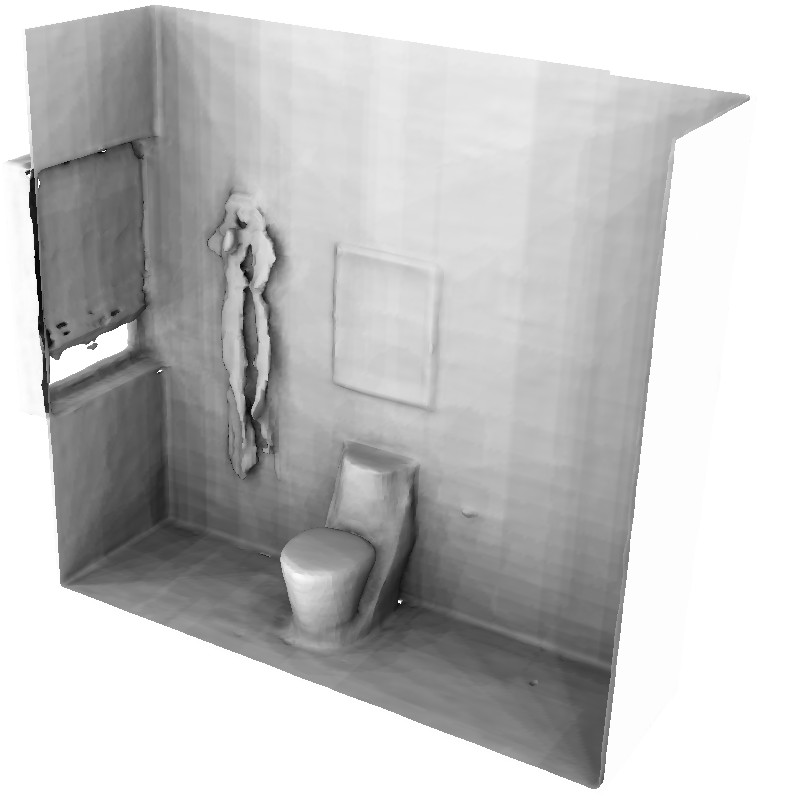}&
    \includegraphics[width=0.20\linewidth]{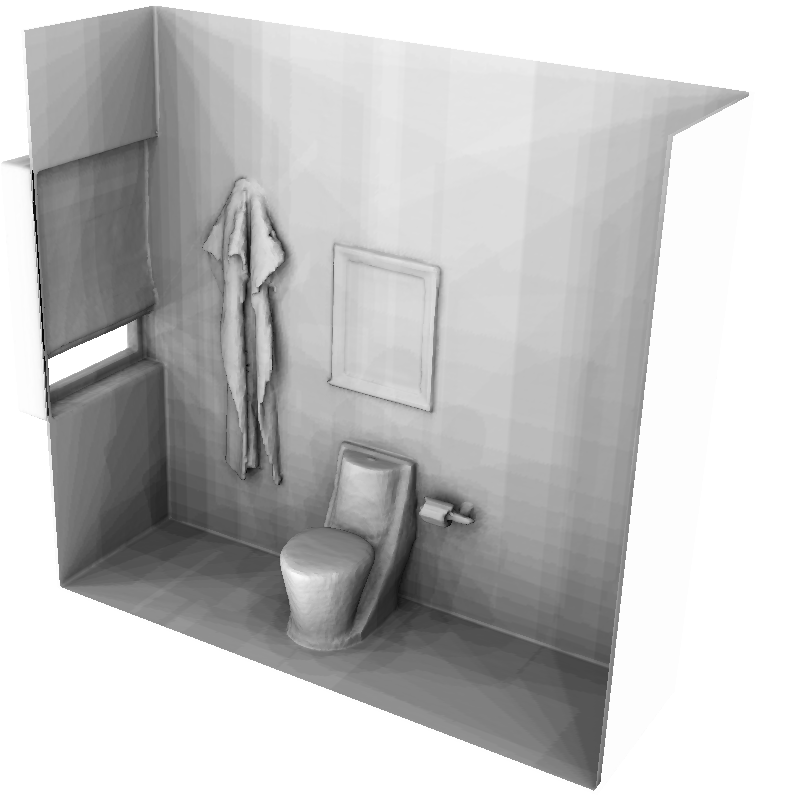} \\
    \midrule
    &&&& Ground truth \\
    &&&& 
    \includegraphics[width=0.20\linewidth]{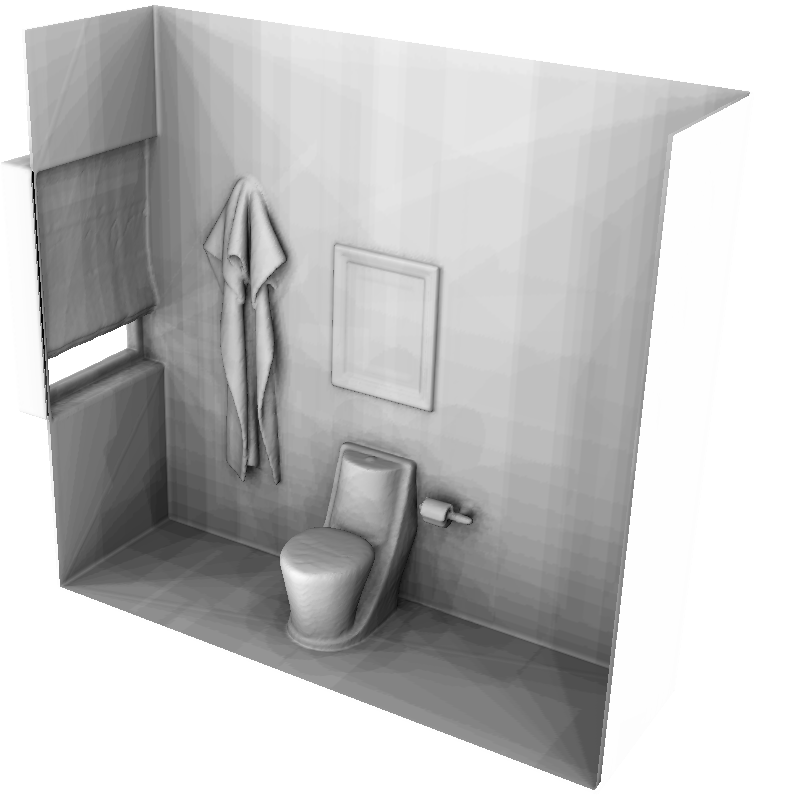}
    \end{tabular}
\end{figure*}

\begin{figure*}[p]
    \centering
    \caption{\textbf{Synthetic Rooms} reconstructions using ConvONet and {\OURS} (ours), from 10k points with noise.}
    \label{fig:syntheticrooms}
    \begin{tabular}{ccc|ccc}
    Input & ConvONet & {\OURS} & Input & ConvONet & {\OURS}
    \\
    \midrule
    \multicolumn{3}{l|}{Room 04 - scene 801} & \multicolumn{3}{l}{Room 04 - scene 802} \\
    \includegraphics[width=0.15\linewidth]{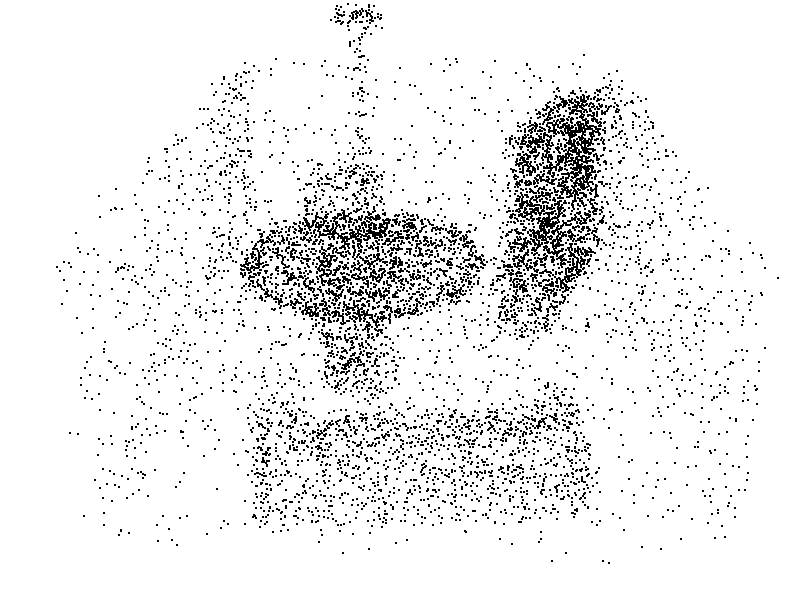}&
    \includegraphics[width=0.15\linewidth]{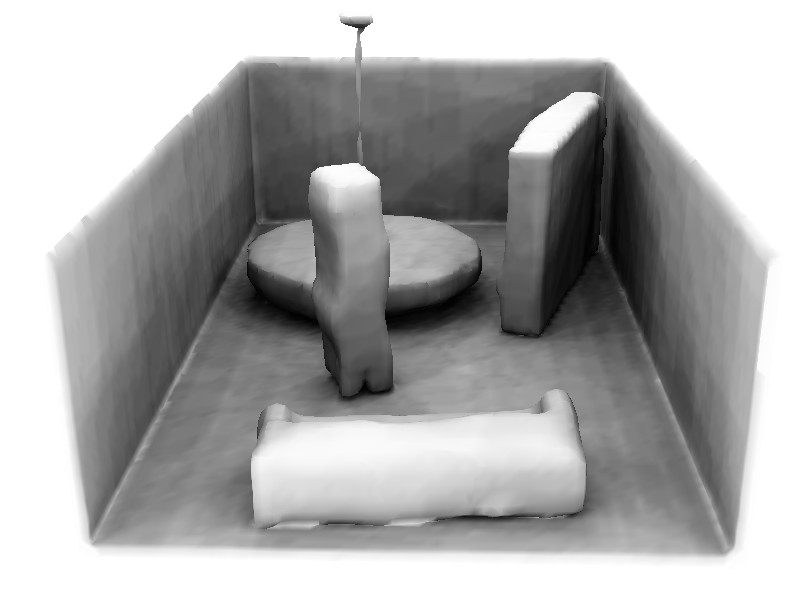}&
    \includegraphics[width=0.15\linewidth]{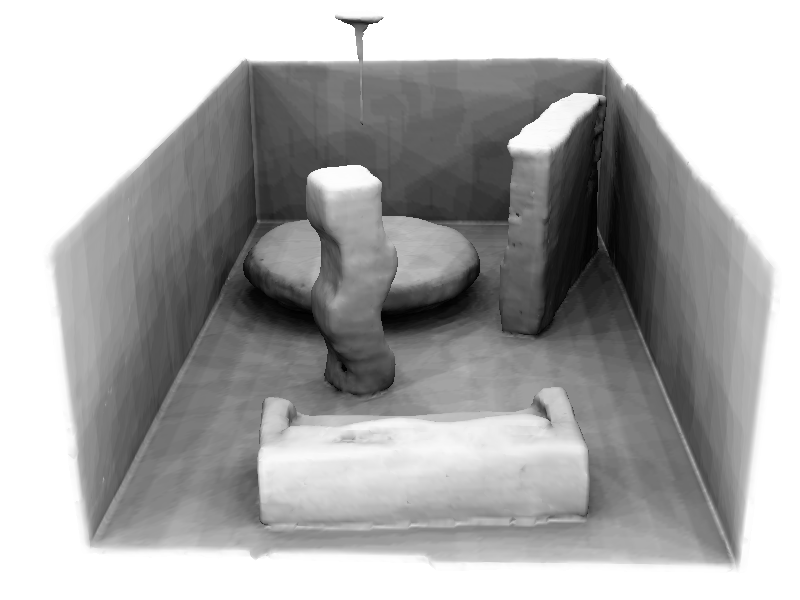}&
    \includegraphics[width=0.15\linewidth]{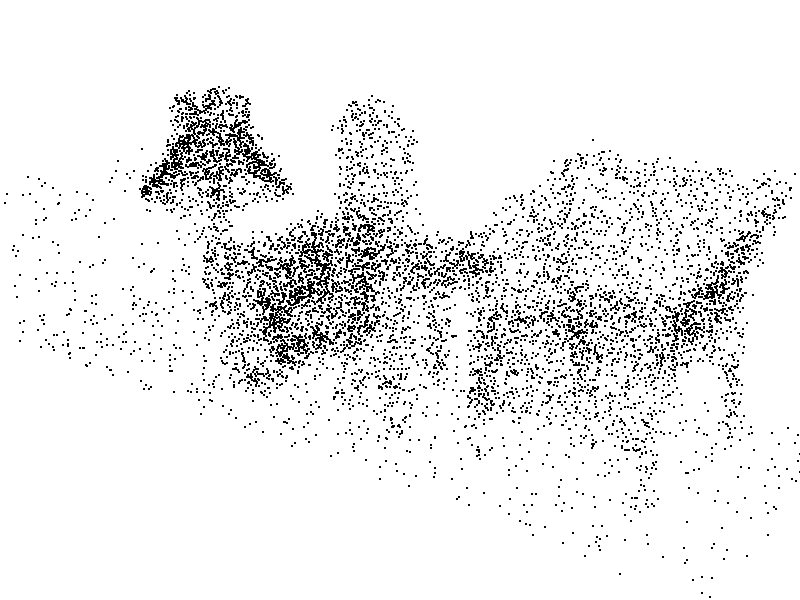}&
    \includegraphics[width=0.15\linewidth]{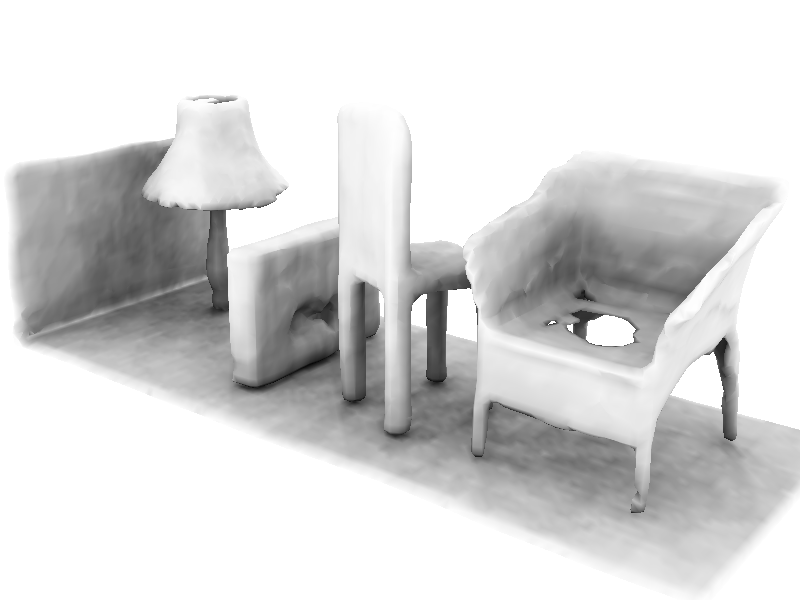}&
    \includegraphics[width=0.15\linewidth]{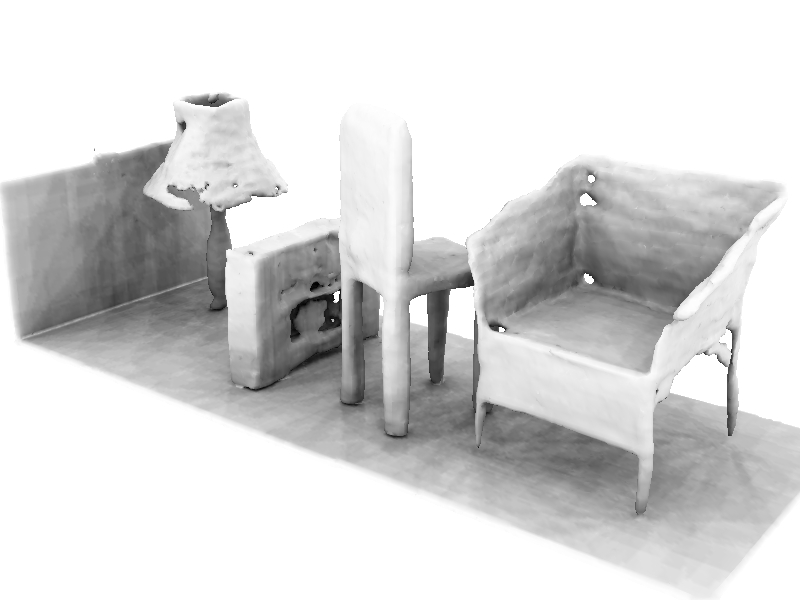}\\
    \midrule
    \multicolumn{3}{l|}{Room 05 - scene 801} & \multicolumn{3}{l}{Room 05 - scene 802} \\
    \includegraphics[width=0.15\linewidth]{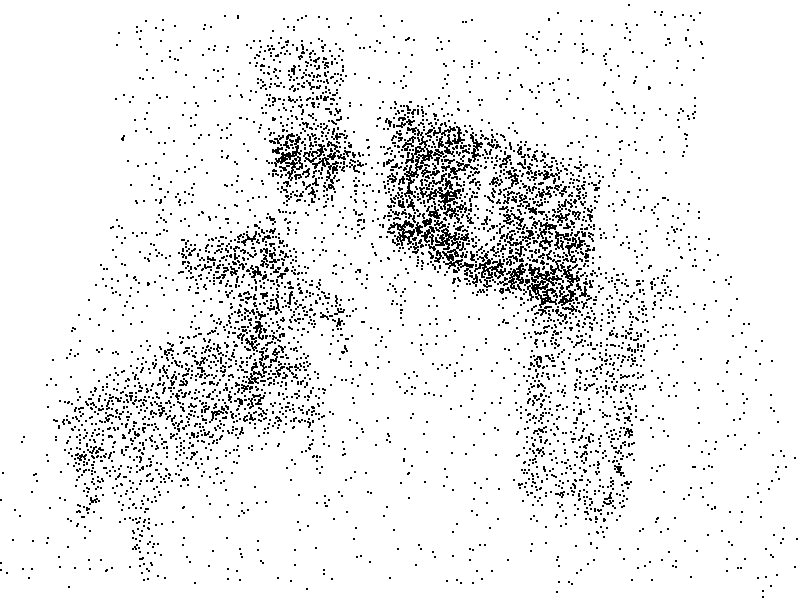}&
    \includegraphics[width=0.15\linewidth]{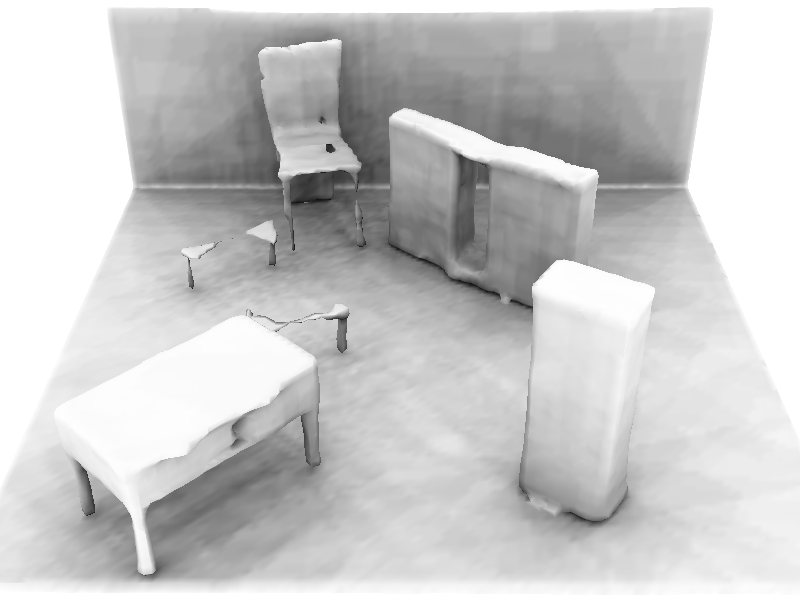}&
    \includegraphics[width=0.15\linewidth]{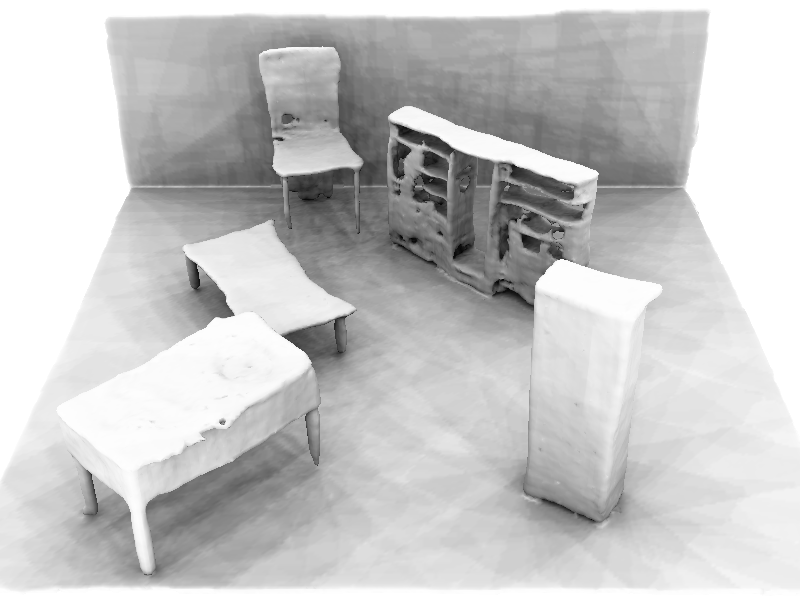}&
    \includegraphics[width=0.15\linewidth]{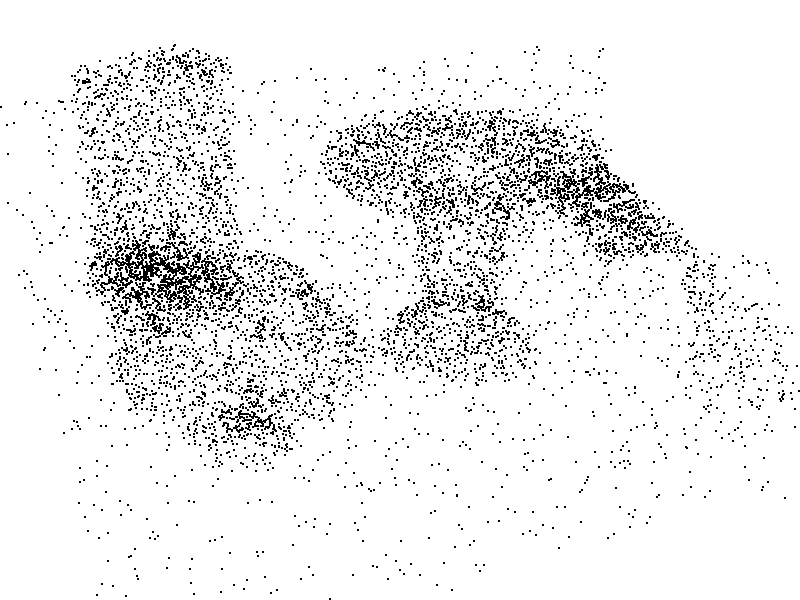}&
    \includegraphics[width=0.15\linewidth]{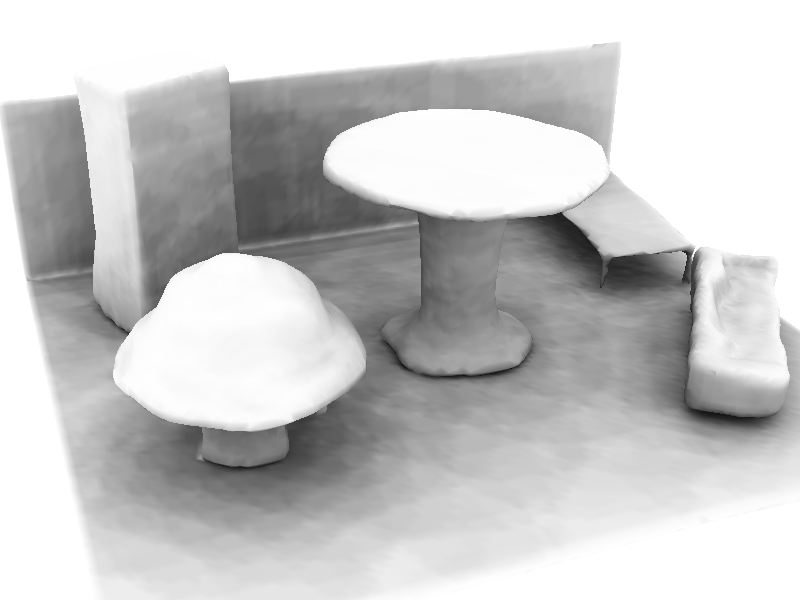}&
    \includegraphics[width=0.15\linewidth]{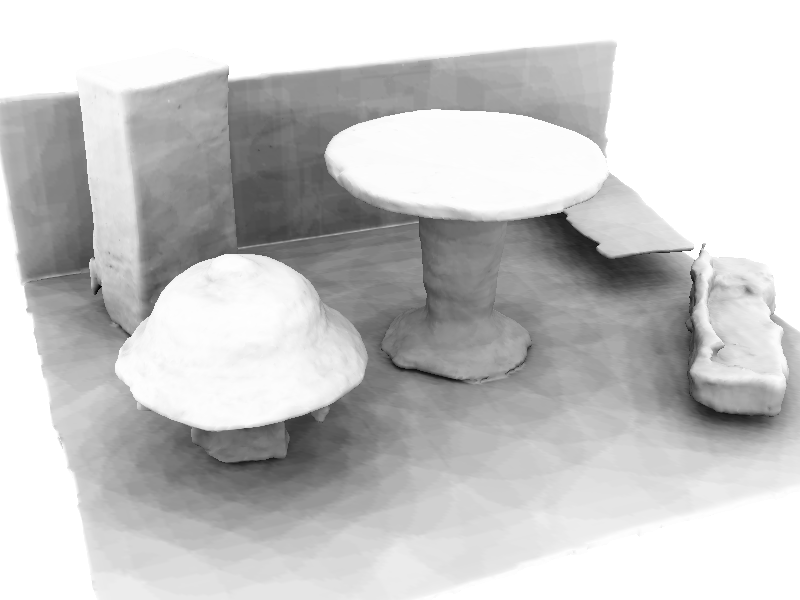}\\
    \midrule
    \multicolumn{3}{l|}{Room 06 - scene 801} & \multicolumn{3}{l}{Room 06 - scene 802} \\
    \includegraphics[width=0.15\linewidth]{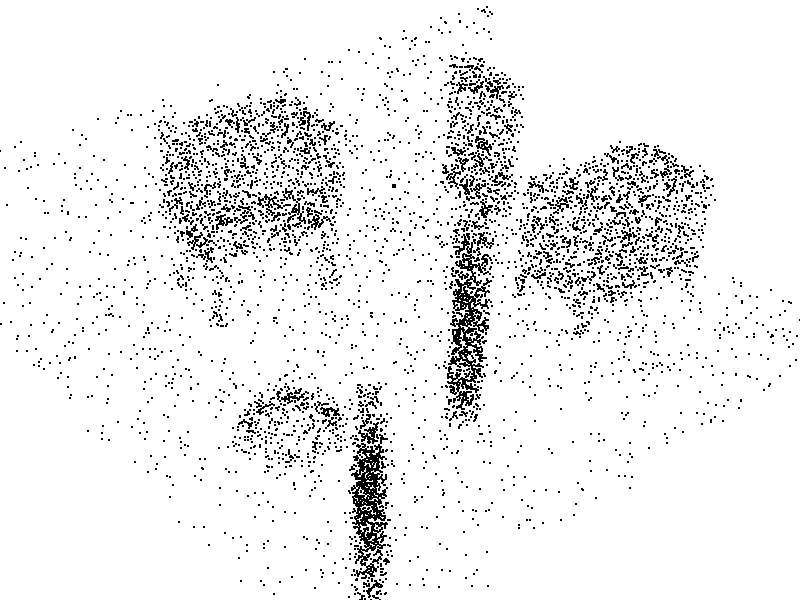}&
    \includegraphics[width=0.15\linewidth]{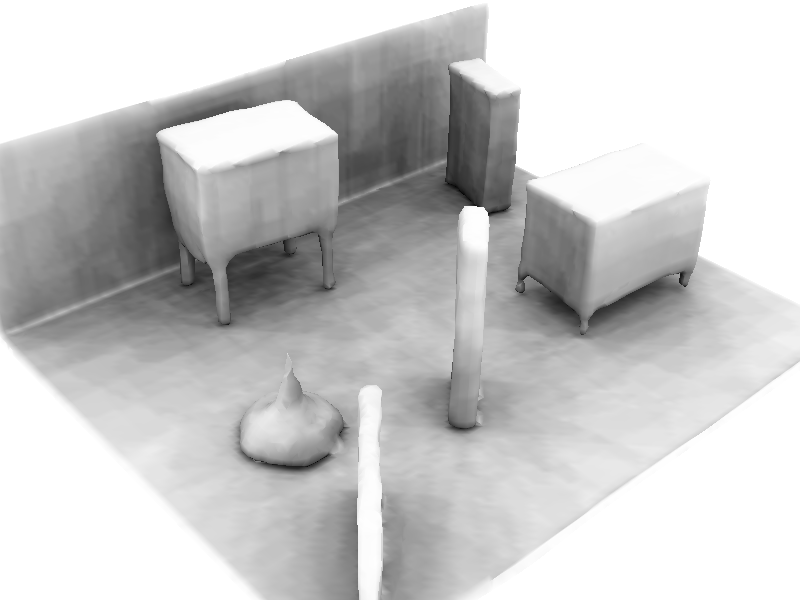}&
    \includegraphics[width=0.15\linewidth]{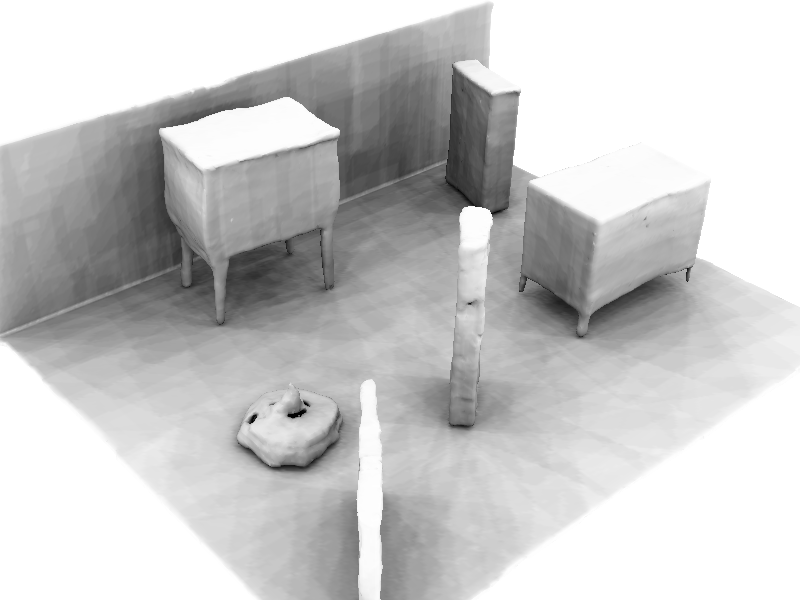}&
    \includegraphics[width=0.15\linewidth]{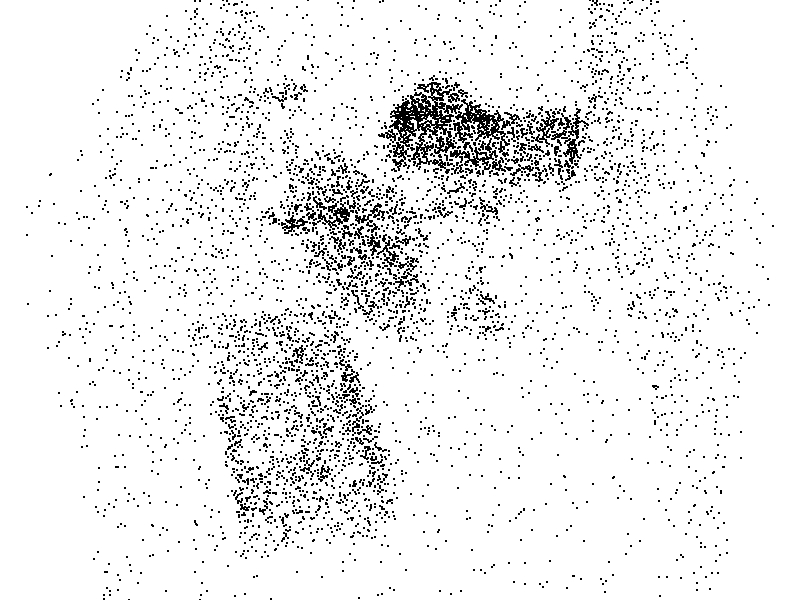}&
    \includegraphics[width=0.15\linewidth]{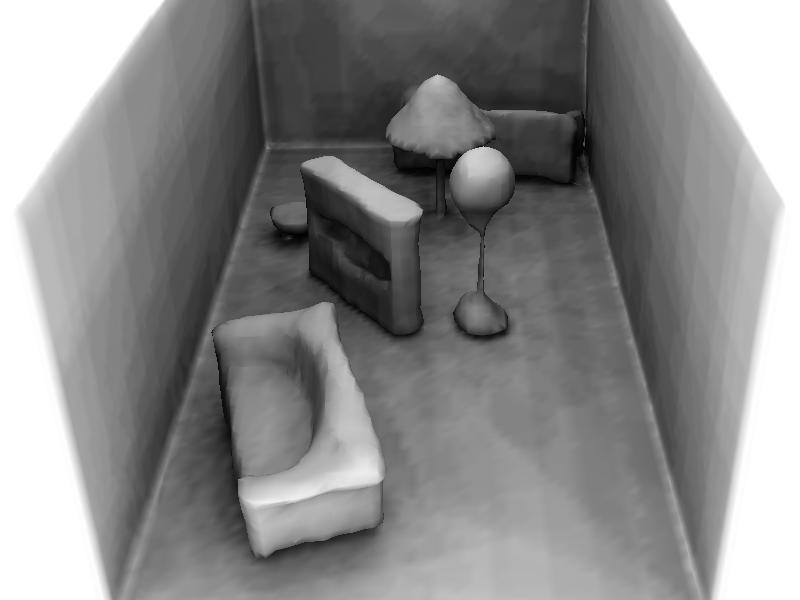}&
    \includegraphics[width=0.15\linewidth]{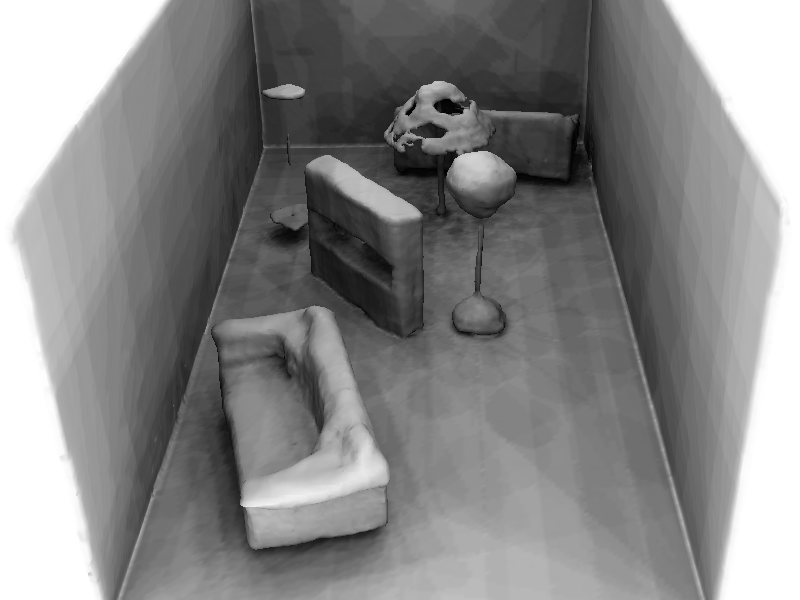}\\
    \midrule
    \multicolumn{3}{l|}{Room 07 - scene 801} & \multicolumn{3}{l}{Room 07 - scene 802} \\
    \includegraphics[width=0.15\linewidth]{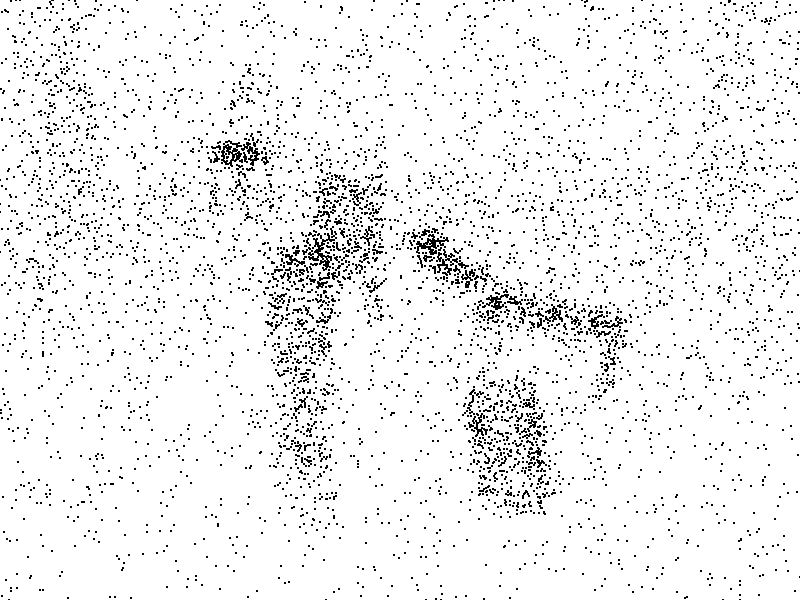}&
    \includegraphics[width=0.15\linewidth]{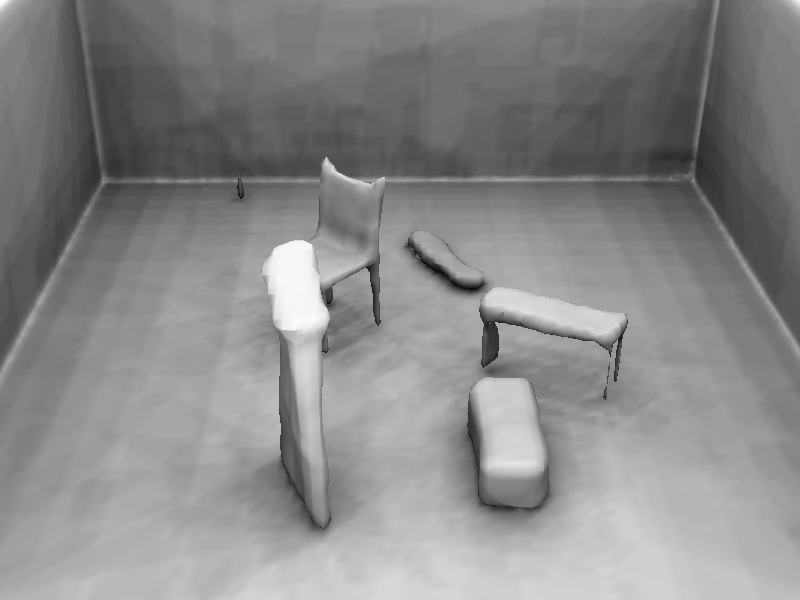}&
    \includegraphics[width=0.15\linewidth]{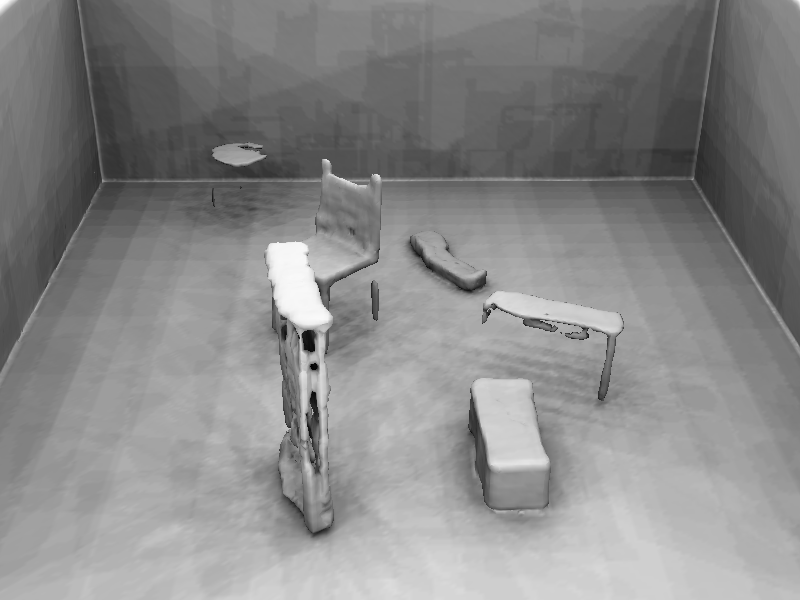}&
    \includegraphics[width=0.15\linewidth]{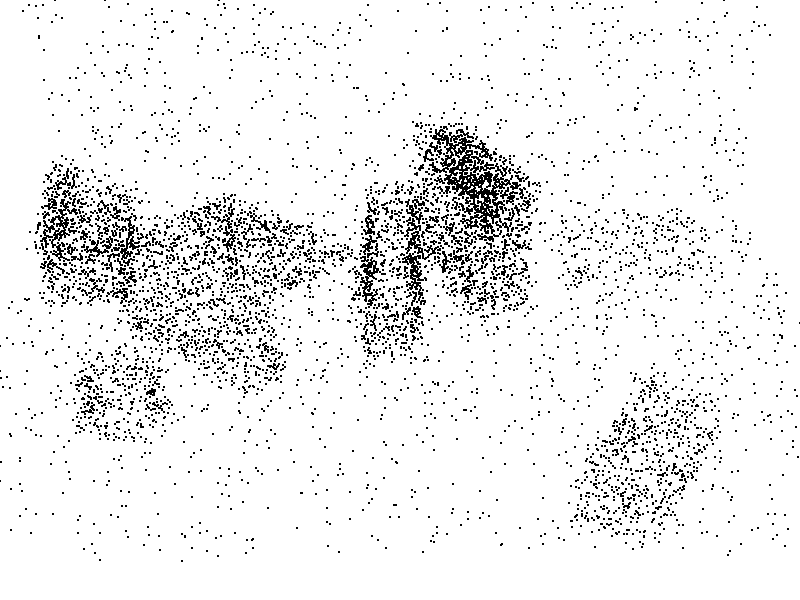}&
    \includegraphics[width=0.15\linewidth]{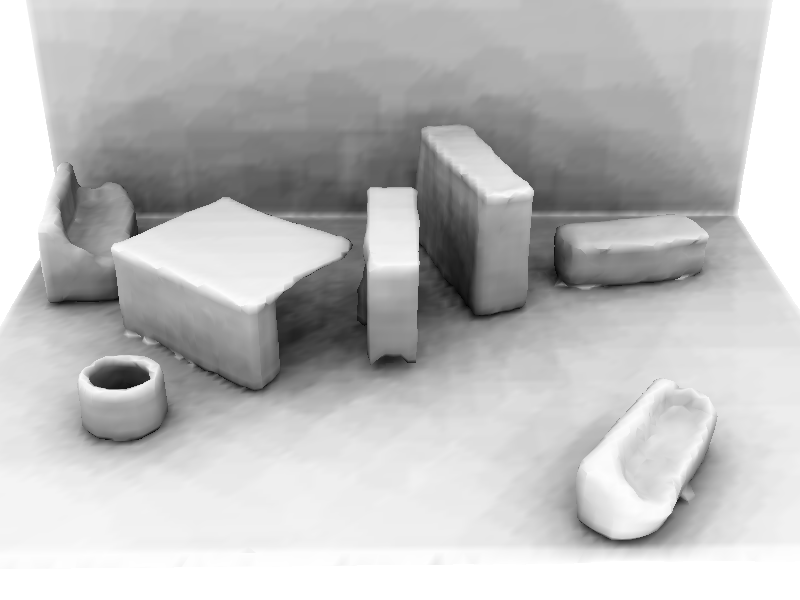}&
    \includegraphics[width=0.15\linewidth]{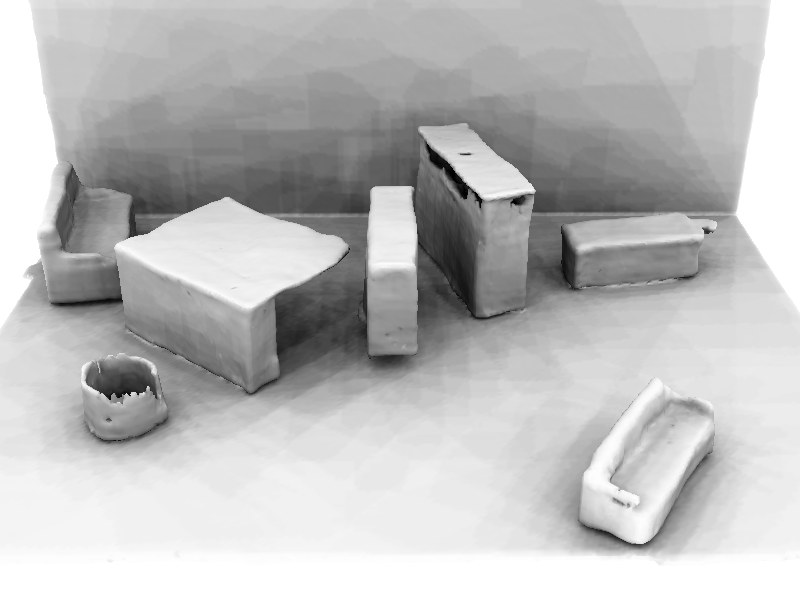}\\
    \midrule
    \multicolumn{3}{l|}{Room 08 - scene 801} & \multicolumn{3}{l}{Room 08 - scene 802} \\
    \includegraphics[width=0.15\linewidth]{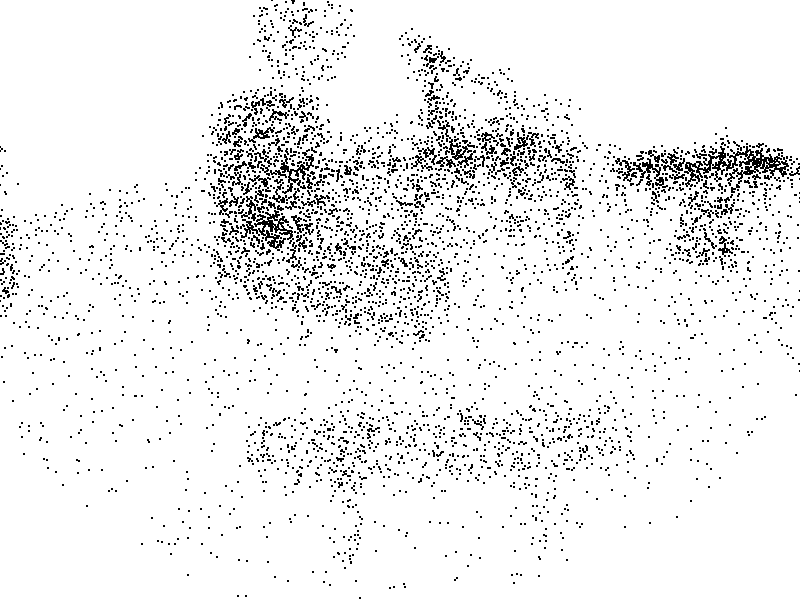}&
    \includegraphics[width=0.15\linewidth]{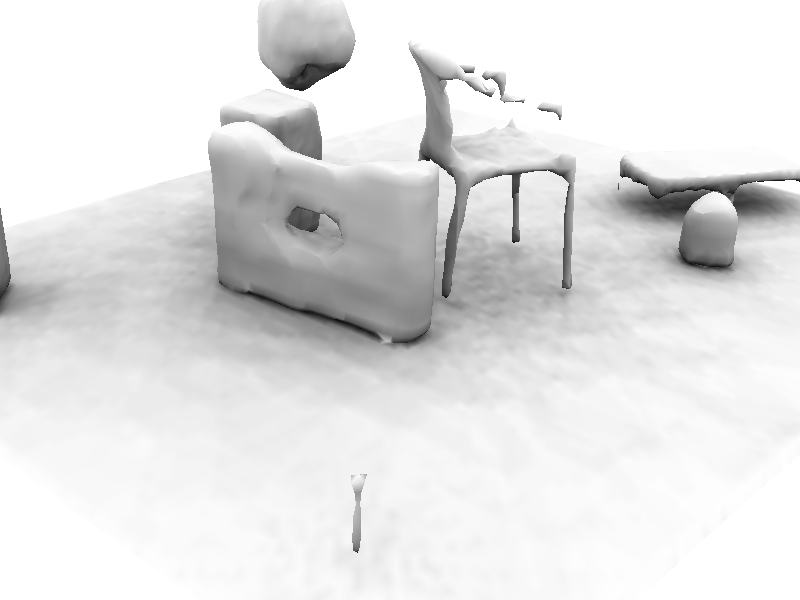}&
    \includegraphics[width=0.15\linewidth]{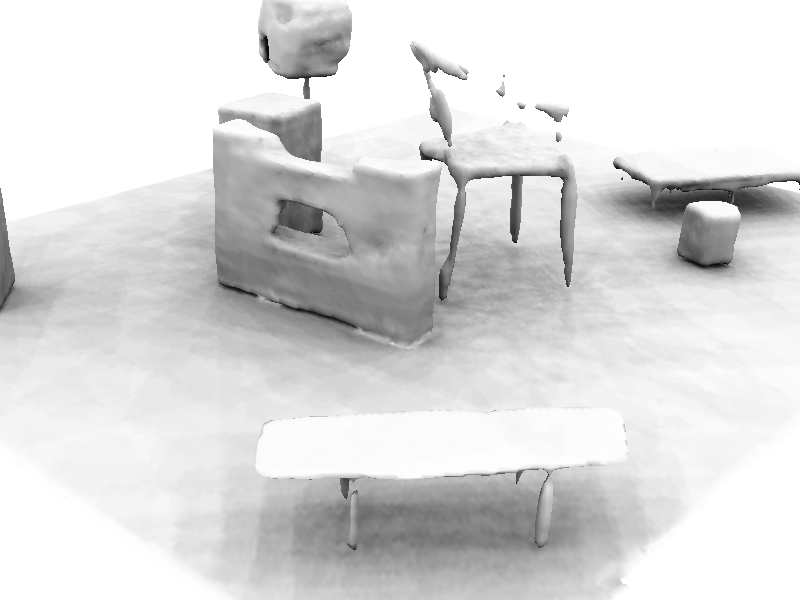}&
    \includegraphics[width=0.15\linewidth]{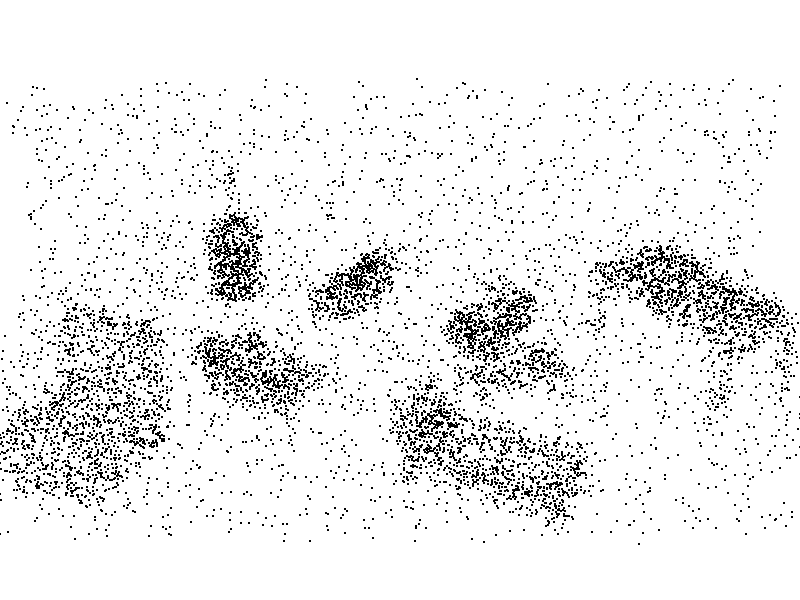}&
    \includegraphics[width=0.15\linewidth]{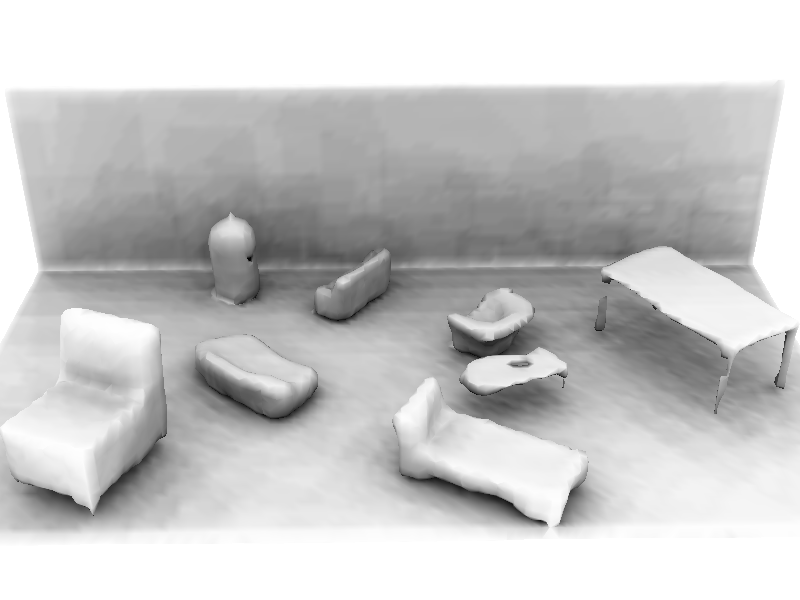}&
    \includegraphics[width=0.15\linewidth]{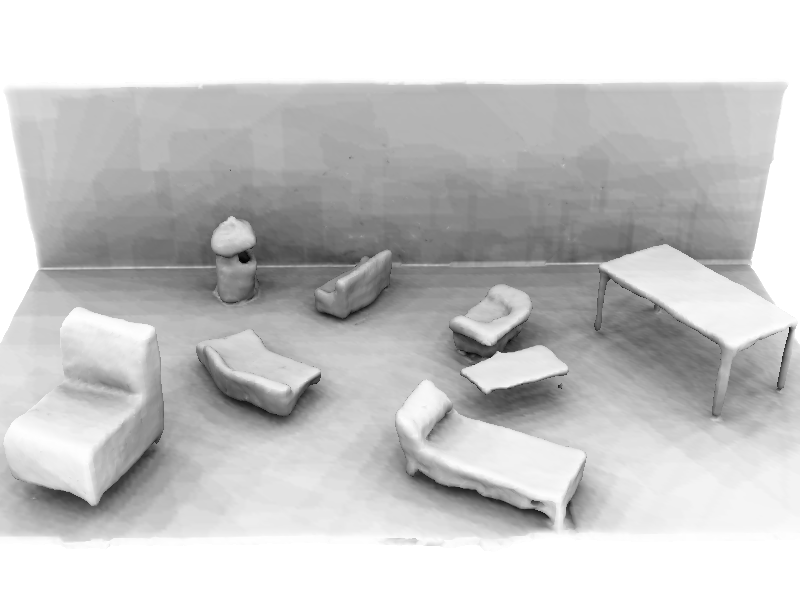}\\
    \end{tabular}
\end{figure*}

\subsection{Metrics}

We use exactly the same evaluation metrics as ConvONet~\cite{Peng2020ECCV}, as specified formally in the supplementary material. However, for our report to be more self-contained, we reformulate here explicitly the metrics that we use.

The surface metrics measure different forms of deviations between two surfaces, i.e., the deviation between the reconstructed surface and the ground-truth surface. In practice, the metrics are approximated by replacing the continuous distances by the distances between points sampled on both surfaces. In particular, the distance of a point~$p$ to a surface~$S$ is approximated by the distance of~$p$ the nearest point~$q$ sampled on surface~$S$. In our experiments, we sample on each surface: 100k points for ShapeNet and Synthetic Rooms; 10k for ABC, Famous and Thingi10k; and 4M points for SceneNet.
As can be seen by the performance of `Oracle' in Table~5 of the paper, which compares the ground-truth against itself via two different samplings, this discretization is a reasonable approximation, although {\OURS} gets close to the error margin when the point cloud is dense and the normals are provided.

\paragraph{Chamfer distance (CD).}

The Chamfer distance between two point clouds $P_1,P_2$ is defined as follows:
\begin{align*}
    \text{Chamfer}(P_1, P_2) =& \> \frac{1}{2\,|P1|} \sum_{p_1 \in P_1} \min_{p_2 \in P_2} d(p_1, p_2) \\ & + \frac{1}{2\,|P2|} \sum_{p_2 \in P_2} \min_{p_1 \in P_1} d(p_1, p_2)
\end{align*}
where $d(p_1, p_2)$ is the distance between points $p_1,p_2$. In the paper, following ONet~\cite{Mescheder2019CVPR} and ConvONet~\cite{Peng2020ECCV}, we use the L1-norm. What we name `CD' in tables is Chamfer$\,\times\,10^2$.

\paragraph{Normal consistency (NC).}

The normal consistency between two point clouds $P_1,P_2$ is defined as follow:
\begin{align*}
    \text{NC}(P_1, P_2) =& \> \frac{1}{2\,|P1|} \sum_{p_1 \in P_1} n_{p_1} . n_{\closest(p_1,P_2)} \\ 
                & + \frac{1}{2\,|P2|} \sum_{p_2 \in P_2} n_{p_2} . n_{\closest(p_2, P_1)}
\end{align*}
where
\begin{align*}
    \closest(p,P) =& \argmin_{p' \in P} d(p, p')
\end{align*}
is the closest point to $p$ in point cloud $P$ and where $n_p$ is the normal at point $p$, given by the orientation of the mesh face on which the point is sampled.

\paragraph{F-Score (FS).}

The F-Score between two point clouds $P_1$ and $P_2$ at a given threshold $t$ is given by:
\begin{align*}
    \text{FS}(t, P_1, P_2) = \frac{2\, \text{Recall} \; \text{Precision} }{\text{Recall}+ \text{Precision}}
\end{align*}
where
\begin{align*}
    \text{Recall}(t, P_1, P_2) = \left| \left\{ p_1 \in P_1, \; \text{s.t.} \min_{p_2 \in P_2} d(p_1, p_2) < t \right\} \right|
    \\
    \text{Precision}(t, P_1, P_2) = \left| \left\{ p_2 \in P_2, \; \text{s.t.} \min_{p_1 \in P_1} d(p_2, p_1) < t \right\} \right|
\end{align*}
In the paper, following ONet~\cite{Mescheder2019CVPR} and ConvONet~\cite{Peng2020ECCV}, we use $t=0.01$.

\paragraph{Intersection over Union (IoU).}

Compared to the previous metrics, which evaluates the quality of the generated surface, the IoU is a volume metric.

Noting TP (resp.\ FP and FN) the number of true positive, i.e., the number of points correctly predicted as full (resp.\ the number of points wrongly predicted as full, and the number of points wrongly predicted as empty), the IoU is defined as follows:
\begin{align*}
    \text{IoU} = \frac{\text{TP}}{\text{TP} + \text{FP} + \text{FN}}
\end{align*}

\subsection{More qualitative results}

\paragraph{{\OURS} vs LIG on ShapeNet (various densities).}
We provide on Figure~\ref{fig:supp:shapenet_lig} more visualizations of ShapeNet reconstructions, comparing LIG to {\OURS} at various densities of input points (with normals). LIG reconstructions were done using the best parameter setting for the method, i.e., with part size 0.20 for 512 and 2048 points, and part size 0.10 for 8192 points. Nevertheless, {\OURS} reconstructs surfaces with more robustness and much sharper details.

\paragraph{{\OURS} vs SPR and LIG on SceneNet (various densities).}
As a complement to Table~5 in the main paper, we provide here on Figure~\ref{fig:scenenet2} the visualization of a reconstruction fragment of a SceneNet scene, also with varying input point densities, comparing SPR, LIG and {\OURS}.  As can be seen, {\OURS} provides a better robustness at low point densities and more details at high point densities.

\paragraph{{\OURS} vs SPR (generalization ability).}
In fact, {\OURS} out-of-the-box adapts well to new shape domains without retraining (Figures~1, 3, 4 and Table~2), especially when given normals (Table~5). 
SPR only works well on high-density point clouds (Figure~4, Tables 2, 4, 5).

\paragraph{{\OURS} vs ConvOnet on Synthetic Rooms.}
As a complement to Table~4 in the main paper, we provide here on Figure~\ref{fig:syntheticrooms} the visualization of reconstructions on the SyntheticRooms dataset (2 first scenes of each data bunch), comparing ConvOnet and {\OURS}. In general, we provide more and sharper details; we are also more robust to thin surfaces, e.g., selves of the bookcase in ``Room 05 - scene 801'' and coffee table in the foreground of ``Room 08 - scene 801''.

\subsection{More quantitative results}

\paragraph{{\OURS} vs PointConv, ONet and ConvOnet on ShapeNet.}
As a complement to Table~3 in the main paper, we provide here in Table~\ref{tab:shapenet_per_class} classwise quantitative results on ShapeNet, comparing {\OURS} to PointConv, ONet and ConvONet (the $3 \times 64^2$ variant, that performs best on ShapeNet).

PointConv is a baseline method which is defined in the ConvONet paper \cite{Peng2020ECCV}. It proceeds as follows: point-wise features are extracted using PointNet++~\cite{Qi2017NIPS}, interpolated using Gaussian kernel regression and feed into the same fully-connected network used in ConvONet~\cite{Peng2020ECCV}. While this baseline uses local information, it does not exploit convolutions. ONet~\cite{Mescheder2019CVPR} is not convolutional either; it operates on shapes as a whole.

As can be seen in the table, {\OURS} largely outperforms the compared methods on all categories, especially on classes featuring complex details such as \emph{lamp}, \emph{rifle}, \emph{vessel} and, to a lesser extent, \emph{airplane}, \emph{car}, \emph{chair} and \emph{loudspeaker}. Yet, the most difficult classes are more or less the same for all methods, including {\OURS}: \emph{lamp} and \emph{car}.

\FloatBarrier

\begin{table*}[h!]
    \centering
    \begin{tabular}{l|cccc|cccc}
                    & \multicolumn{4}{c|}{IoU\,$\uparrow$} & \multicolumn{4}{c}{CD$\downarrow$}\\[1mm] 
        Category    & PointConv & ONet & \!ConvONet\! & \OURS & PointConv & ONet & \!ConvONet\! & \OURS \\
        \midrule
        Airplane    & 0.579 & 0.734 & 0.849 & \bf 0.902 & 1.40 & 0.64 & 0.34 & \bf 0.23	\\
        Bench       & 0.537 & 0.682 & 0.830 & \bf 0.865 & 1.20 & 0.67 & 0.35 & \bf 0.28\\
        Cabinet     & 0.824 & 0.855 & 0.940 & \bf 0.960 & 1.15 & 0.82 & 0.46 & \bf 0.37\\
        Car         & 0.767 & 0.830 & 0.886 & \bf 0.921	& 1.49 & 1.04 & 0.75 & \bf 0.41\\
        Chair       & 0.667 & 0.720 & 0.871 & \bf 0.919	& 1.29 & 0.95 & 0.46 & \bf 0.33\\
        Display     & 0.743 & 0.799 & 0.927 & \bf 0.956	& 1.06 & 0.82 & 0.36 & \bf 0.28\\
        Lamp        & 0.495 & 0.546 & 0.785 & \bf 0.877	& 2.15 & 1.59 & 0.59 & \bf 0.33\\
        Loudspeaker & 0.807 & 0.826 & 0.918 & \bf 0.957	& 1.48 & 1.18 & 0.64 & \bf 0.41\\
        Rifle       & 0.565 & 0.668 & 0.846 & \bf 0.897	& 0.98 & 0.66 & 0.28 & \bf 0.19\\
        Sofa        & 0.811 & 0.865 & 0.936 & \bf 0.963	& 1.04 & 0.73 & 0.42 & \bf 0.30\\
        Table       & 0.654 & 0.739 & 0.888 & \bf 0.924	& 1.13 & 0.76 & 0.38 & \bf 0.31\\
        Telephone   & 0.856 & 0.896 & 0.955 & \bf 0.968	& 0.61 & 0.46 & 0.27 & \bf 0.22\\
        Vessel      & 0.652 & 0.729 & 0.865 & \bf 0.927	& 1.38 & 0.94 & 0.43 & \bf 0.25\\
        \midrule
        Mean        & 0.689 & 0.761 & 0.884 & \bf 0.926	& 1.26 & 0.87 & 0.44 & \bf 0.30\\
    \end{tabular}
    
    \bigskip
    
    \begin{tabular}{l|cccc|cccc}
                    & \multicolumn{4}{c|}{NC\,$\uparrow$} & \multicolumn{4}{c}{FS$\uparrow$}\\[1mm] 
        Category    & PointConv & ONet & \!ConvONet\! & \OURS & PointConv & ONet & \!ConvONet\! & \OURS \\
        \midrule
        Airplane    & 0.819 & 0.886 & 0.931 & \bf 0.944	& 0.562 & 0.829 & 0.965 & \bf 0.994\\
        Bench       & 0.811 & 0.871 & 0.921 & \bf 0.928	& 0.617 & 0.827 & 0.964 & \bf 0.988 \\
        Cabinet     & 0.895 & 0.913 & 0.956 & \bf 0.961 & 0.719 & 0.833 & 0.956 & \bf 0.979 \\
        Car         & 0.845 & 0.874 & 0.893 & \bf 0.894 & 0.577 & 0.747 & 0.849 & \bf 0.946 \\
        Chair       & 0.851 & 0.886 & 0.943 & \bf 0.956 & 0.618 & 0.730 & 0.939 & \bf 0.985 \\
        Display     & 0.910 & 0.926 & 0.968 & \bf 0.975 & 0.679 & 0.795 & 0.971 & \bf 0.994 \\
        Lamp        & 0.779 & 0.809 & 0.900 & \bf 0.929 & 0.453 & 0.581 & 0.892 & \bf 0.975 \\
        Loudspeaker & 0.894 & 0.903 & 0.939 & \bf 0.952 & 0.647 & 0.727 & 0.892 & \bf 0.964 \\
        Rifle       & 0.796 & 0.849 & 0.929 & \bf 0.949 & 0.682 & 0.818 & 0.980 & \bf 0.998 \\
        Sofa        & 0.900 & 0.928 & 0.958 & \bf 0.967 & 0.697 & 0.832 & 0.953 & \bf 0.989 \\
        Table       & 0.878 & 0.917 & 0.959 & \bf 0.966 & 0.694 & 0.824 & 0.967 & \bf 0.991 \\
        Telephone   & 0.961 & 0.970 & 0.983 & \bf 0.985 & 0.880 & 0.930 & 0.989 & \bf 0.998 \\
        Vessel      & 0.817 & 0.857 & 0.919 & \bf 0.940 & 0.550 & 0.734 & 0.931 & \bf 0.989 \\
        \midrule
        Mean        & 0.858 & 0.891 & 0.938 & \bf 0.950 & 0.644 & 0.785 & 0.942 & \bf 0.984
    \end{tabular}
    
    \caption{\textbf{Classwise ShapeNet reconstruction.} All models are trained on 3k noisy points. Results for methods other than {\OURS} are reported from the supplementary material of ConvONet~\cite{Peng2020ECCV}.}
    \label{tab:shapenet_per_class}
\end{table*}

\FloatBarrier

\section{Use of existing assets}

\subsection{Pre-existing code}

The implementation of our approach has several dependencies, that are all free to use for research purposes. The main dependencies of our code are as follows:
\begin{itemize}[itemsep=4pt,topsep=4pt, parsep=0pt]

    \item \textbf{FKAConv}\footnote{\label{foot:fkaconv}\url{https://github.com/valeoai/FKAConv}} \cite{Boulch2020ACCV}, under Apache License v2.0.

    \item \textbf{PyTorch}\footnote{\label{foot:pytorch}\url{https://pytorch.org/}}, under the Apache CLA,

    \item \textbf{PyTorch-Geometric}\footnote{\label{foot:pytorchgeom}\url{https://pytorch-geometric.readthedocs.io/}}, under the MIT License,

\end{itemize}
The code of {\OURS}\footnote{\url{https://github.com/valeoai/POCO}} itself is freely available, under Apache License v2.0.

\subsection{Datasets}\label{sec:assetsdatasets}

For the experiments, we used several datasets that are freely available for research purpose:
\begin{itemize}[itemsep=4pt,topsep=4pt, parsep=0pt]

    \item \textbf{ABC}\footnote{\label{foot:abc}\url{https://deep-geometry.github.io/abc-dataset/}} is under the Onshape Terms of Use\footnote{\url{https://www.onshape.com/en/legal/terms-of-use}}. We used the subset preprocessed and made available by the authors of Points2Surf\footnoteref{foot:points2surf} \cite{Erler2020Points2Surf}.
    
    \item \textbf{Famous} is a set of shapes of various origins, among which the Stanford 3D Scanning Repository\footnote{\url{http://graphics.stanford.edu/data/3Dscanrep/}} \cite{krishnamurthy1996fitting}. This set of shapes is described, preprocessed and made available by the authors of Points2Surf\footnoteref{foot:points2surf} \cite{Erler2020Points2Surf}.

    \item \textbf{MatterPort3D}\footnote{\label{foot:matterport3d}\url{https://niessner.github.io/Matterport/}} \cite{chang2017matterport3d} is under a user license agreement for academic use. We used scenes preprocessed by the authors of SA-ConvONet\footnoteref{foot:saconvonet} \cite{tang2021sign}.

    \item \textbf{Real-World} point clouds used in the paper are described, preprocessed and made available by the authors of Points2Surf\footnoteref{foot:points2surf} \cite{Erler2020Points2Surf}.

    \item \textbf{SceneNet}\footnote{\url{https://robotvault.bitbucket.io/}} \cite{handa2015scenenet, handa2016scenenet, Handa2016Understanding} is under the CC BY-NC 4.0,
    for research purposes only. We made meshes watertight using \textbf{Watertight Manifold}\footnote{\label{foot:watertightmanifold}\url{https://github.com/hjwdzh/Manifold}} \cite{huang2018robust}, that enables code use under mild conditions.

    \item \textbf{ShapeNet}\footnote{\label{foot:shapenet}\url{https://shapenet.org/}} \cite{Chang2015ARXIV} has a licence for non commercial research or educational purposes. We used the version of ShapeNet as preprocessed by the authors of \textbf{ONet}\footnote{\label{foot:onet}\url{https://github.com/autonomousvision/occupancy_networks}}  \cite{Mescheder2019CVPR}, which itself reuses the preprocessing of the authors of \textbf{3D-R2N2}\footnote{\label{foot:3dr2n2}\url{https://github.com/chrischoy/3D-R2N2}} \cite{Choy2016ECCV}.

    \item \textbf{Synthetic Rooms}\footnoteref{foot:convonet} is a dataset created by the authors of ConvONet \cite{Peng2020ECCV} based on ShapeNet models.

    \item \textbf{Thingi10K}\footnote{\label{foot:thingi10k}\url{https://ten-thousand-models.appspot.com}} \cite{zhou2016thingi10k} is a freely available collection of shapes under various licences. We used the subset preprocessed and made available by the authors of Points2Surf\footnoteref{foot:points2surf} \cite{Erler2020Points2Surf}.

\end{itemize}

\subsection{Methods}\label{sec:assetsmethods}

We compared to a number of reconstruction methods, reusing the code made available by their authors:
\begin{itemize}[itemsep=4pt,topsep=4pt, parsep=0pt]

\item \textbf{ConvONet}\footnote{\label{foot:convonet}\url{https://github.com/autonomousvision/convolutional_occupancy_networks}} \cite{Peng2020ECCV} under the MIT License.

\item \textbf{LIG}\footnote{\label{foot:lig}\url{https://github.com/tensorflow/graphics/tree/master/tensorflow_graphics/projects/local_implicit_grid}} \cite{Jiang2020CVPR} probably under Apache License v2,

\item \textbf{Neural Splines}\footnote{\label{foot:neuralsplines}\url{https://github.com/fwilliams/neural-splines}} \cite{Williams2021NeuralSplines}  under the
MIT License,

\item \textbf{Points2Surf}\footnote{\label{foot:points2surf}\url{https://github.com/ErlerPhilipp/points2surf}} \cite{Erler2020Points2Surf} under the MIT License,

\item \textbf{SA-ConvONet}\footnote{\label{foot:saconvonet}\url{https://github.com/tangjiapeng/SA-ConvONet}} \cite{tang2021sign} under the MIT License,

\item \textbf{SPR}\footnote{\label{foot:spr}\url{https://github.com/mkazhdan/PoissonRecon}} \cite{Kazhdan2013SIGGRAPH} under the MIT License.

\end{itemize}
We also compared to \textbf{AtlasNet} \cite{Groueix2018CVPR}, \textbf{DeepSDF} \cite{Park2019CVPR}, \textbf{DP-ConvONet} \cite{Lionar_2021_WACV}, \textbf{ONet} \cite{Mescheder2019CVPR}, but only reusing the numbers mentioned in \cite{Peng2020ECCV, Lionar_2021_WACV}.

\medskip

Here are some methods we would have liked to compare to, but could not in practice:
\begin{itemize}[itemsep=-2pt,topsep=1pt]

\item \textbf{AdaConv}\footnote{\label{foot:adaconv}\url{https://github.com/isl-org/adaptive-surface-reconstruction}} \cite{Ummenhofer2021Adaptive}: The repository provides raw code but no pre-trained model nor instructions or scripts to train or to test, which may lead to misuses and wrong comparisons. 

\item \textbf{NDF}\footnote{\url{https://github.com/jchibane/ndf}} \cite{Chibane2020Neural}: The repository provides code but only a pre-trained model for ShapeNet cars. For scene reconstruction, it does not offer preprocessed data or any data preprocessing procedure to retrain a model, nor instructions to run NDF using a sliding window scheme, as alluded to in the supplementary material.

\end{itemize}
As indicated in Table~\ref{tab:datasets}, some authors also have not made their code or their model available to allow comparisons.

\section{Societal impact}

We believe our 3D reconstruction approach has \textbf{very little potential for malicious uses} (including disinformation, surveillance, invasion of privacy, endangering security), not more, e.g., than image enhancement methods in the 2D data case, and not more than hundreds of previously published 3D reconstruction methods. Besides, we are not bound nor promoting any dataset that would lead to unfairness in any sense. The use of our method has a \textbf{modest environmental impact} as the training time (a few days on a single GPU for a large dataset) and the inference times (minutes, or hours for very large point clouds) are somewhat moderate, and favorably compare to many learning-based approaches.

On the contrary, applications of our method can be found in various domains, with positive societal impacts:

\textbf{Heritage preservation.}
Digitizing cultural objects and monuments allows a form of heritage preservation and enables virtual museums to make works of art and culture more widely accessible.
    
\textbf{Infrastructure and building maintenance.}
Reconstructing models of existing infrastructures and buildings is of high interest for the construction industry. These models are particularly useful to plan and organize maintenance. This is particularly useful in a context of aging infrastructures and building renovation for energy-saving insulation.

\textbf{Augmented and virtual reality.}
Surface and volume reconstruction are useful assets for augmented and virtual reality, whether it is for professional use (e.g., on-site maintenance of equipment) or entertainment (video games, special effects for the film industry), which is however to be consumed in moderation.

\end{document}